\documentclass[3p,times,review]{elsarticle}

\usepackage{amssymb}
\usepackage{graphicx}
\usepackage{adjustbox}
\usepackage{courier}
\usepackage{amsmath, array}
\usepackage[english]{babel}
\usepackage{ctable}
\usepackage{subfig}
\usepackage{multirow}
\usepackage{lscape}
\usepackage{colortbl}
\usepackage{xcolor}
\usepackage{float}
\usepackage[linesnumbered, ruled, vlined]{algorithm2e}
\usepackage{url}
\usepackage{stackengine} 
\usepackage{comment}
\usepackage[pdftex]{hyperref}
\usepackage[flushleft]{threeparttable}

\usepackage{amsthm}
\usepackage[clock]{ifsym}
\usepackage{pifont}

\theoremstyle{plain}

\theoremstyle{definition}
\newtheorem{defn}{Definition}[]

\makeatletter
\def\th@plain{%
  \thm@notefont{}
  \itshape 
}
\def\th@definition{%
  \thm@notefont{}
  \normalfont 
}
\makeatother

\newcolumntype{R}[2]{%
    >{\adjustbox{angle=#1,lap=\width-(#2)}\bgroup}%
    l%
    <{\egroup}%
}
\newcommand*\rot{\multicolumn{1}{R{30}{1.5em}}}

\newcommand{\cmark}{\ding{51}}%
\newcommand{\xmark}{\ding{55}}%

\hypersetup{pdfborder=0 0 0}
\SetAlgoCaptionSeparator{.}
\SetKwRepeat{Do}{do}{while}
\begin{document}

    \begin{frontmatter}

\title{Mining Frequent Patterns in Process Models}

\author[CiTIUS]{David Chapela-Campa\corref{mycorrespondingauthor}}
\cortext[mycorrespondingauthor]{Corresponding author}
\ead{david.chapela@usc.es}

\author[CiTIUS]{Manuel Mucientes}
\ead{manuel.mucientes@usc.es}

\author[CiTIUS]{Manuel Lama}
\ead{manuel.lama@usc.es}

\address[CiTIUS]{Centro Singular de Investigaci\'{o}n en Tecnolox\'{i}as da Informaci\'{o}n (CiTIUS)\\
 Universidade de Santiago de Compostela. Santiago de Compostela, Spain\\}

\begin{abstract}
Process mining has emerged as a way to analyze the behavior of an organization by extracting knowledge from event logs and by offering techniques to discover, monitor and enhance real processes.
In the discovery of process models, retrieving a complex one, i.e., a hardly readable process model, can hinder the extraction of information.
Even in well-structured process models, there is information that cannot be obtained with the current techniques.
In this paper, we present WoMine, an algorithm to retrieve frequent behavioural patterns from the model.
Our approach searches in process models extracting structures with sequences, selections, parallels and loops, which are frequently executed in the logs.
This proposal has been validated with a set of process models, including some from BPI Challenges, and compared with the state of the art techniques.
Experiments have validated that WoMine can find all types of patterns, extracting information that cannot be mined with the state of the art techniques.
\end{abstract}


\end{frontmatter}
    
    \section{Introduction\label{sec:introduction}}

With the explosion of process-related data, the behavioural analysis and study of business processes has become more popular.
Process mining offers techniques to discover, monitor and enhance real processes by extracting knowledge from event logs, allowing to understand \textit{what is really happening in a business process}, and not \textit{what we think is going on}~\cite{vanderAalst2011discovery}.
Nevertheless, there are scenarios ---highly complex process models--- where process discovery techniques are not able to provide enough intelligible information to make the process model understandable to users.

There are four quality dimensions to measure \textit{how good} a process model is: \textit{fitness replay}, \textit{precision}, \textit{generalization} and \textit{simplicity}.
Regarding the latter, discovering a complex process model, i.e., a hardly readable process model, can totally hinder its quality~\cite{de2015log} making difficult the retrieval of behavioural information.
Different techniques have been proposed to tackle this problem: the simplification of already mined models~\cite{de2015log,fahland2011simplifying}, the search of simpler structures in the logs~\cite{leemans2014discovery,mannila1997discovery,tax2016mining}, or the clusterization of the log into smaller and more homogeneous subsets of traces to discover different models within the same process~\cite{greco2004mining,greco2006discovering,song2008trace}.
Although these techniques improve the understandability of the process models, for real processes the model structure remains complex, being difficult to understand by users.

As an alternative to these techniques, there exist some proposals whose aim is to extract structures ---or sub\-pro\-cesses--- within the model which are relevant to describe the process model.
In these approaches, the relevance of a structure is measured as: \textit{i)} the total number of executions~\cite{leemans2014discovery,tax2016mining}, e.g., a structure executed 1,000 times; or \textit{ii)} its high repetition in the traces of the log~\cite{bui2012framework,greco2006mining}, e.g., a subprocess which appears most of the times the process is executed.
In this paper, we will focus in the search of frequent behavioural patterns of the second type.
The extraction of these frequent structures is useful in both highly complex and well-structured models.
In complex models, it allows to abstract from all the behaviour and focus on relevant structures.
Additionally, the application of these techniques in well-structured process models retrieves frequent subprocesses which can be, for instance, the objective of optimizations due to its frequent execution within the process.
The knowledge obtained by extracting these frequent patterns is valuable in many fields.
For instance, in e-learning, the interactions of the students with the learning management system can be registered in order to reconstruct their behaviour during the subject~\cite{barreiros2014softlearn}, i.e., to reconstruct the learning path followed by the students.
The extraction of frequent patterns from these learning paths ---processes--- can help teachers to improve the learning design of the subject, enabling its adaptation to the students behaviour.
It can also reveal behavioral patterns that should not be happening.
In addition, for other business processes like, for example, call centers where the objective is to retain the customers, the discovery of frequent behaviours can be decision-making.
In this domain, process models tend to contain numerous choices and loops, where frequent structures can show a possible behavior which leads to retain customers.
This knowledge can be used to plan new strategies in order to reduce the number of clients who drop out, by exploiting the paths that lead to retain customers, or avoiding those which end in dropping out.

In this paper we present WoMine, an algorithm to mine frequent patterns from a process model, measuring their frequency in the instances of the log.
The main novelty of WoMine, which is based on the \textit{$w$-find} algorithm~\cite{greco2006mining}, is that it can detect frequent patterns with all type of structures ---even n-length cycles, very common structures in real processes.
It can also ensure which traces are compliant with the frequent pattern in a percentage over a threshold.
Furthermore, WoMine is robust w.r.t. the quality of the mined models with which it works, i.e., its results do not depend highly on the fitness replay and precision of the mined models.
The algorithm has been tested with 20 synthetic process models ranging from 20 to 30 unique tasks, and containing loops, parallelisms, selections, etc.
Experiments have been also run with 12 real complex logs of the Business Process Intelligence Challenges.

The remainder of this paper is structured as follows. 
Section~\ref{sec:state-of-art} introduces the algorithms related with the purpose of this paper.
The required background of the paper is introduced in Section~\ref{sec:preliminaries}.
Section~\ref{sec:algorithm} presents the main structure of WoMine, followed by detailed explanations in Sections~\ref{sec:measure-frequency} and~\ref{sec:post-simplification}.
Finally, Section~\ref{sec:experiments} describes an evaluation of the approach, and Section~\ref{sec:conclusions} summarizes the conclusions of the paper.

    \section{Related Work\label{sec:state-of-art}}

A simple and popular technique to detect frequent structures in a process model is the use of \textit{heat maps}, which can be found in applications like DISCO~\cite{gunther2012disco}. 
It provides a simple technique which can retrieve the frequent structures of a process model considering the individual frequency of each arc.
Other techniques check the frequency of each pattern taking into account all the structure, and not the individual frequency.
These approaches, under the frequent pattern mining field~\cite{han2007frequent}, can build frequent patterns based just on the logs, searching in them for frequent sequences of tasks~\cite{han2000freespan,han2001prefixspan,zaki2001spade}.
Improving this search, episode mining techniques focus their search in frequent, and more complex structures such as parallels~\cite{leemans2014discovery,mannila1997discovery}.
With a different approach, the $w$-find algorithm~\cite{greco2006mining} uses the process model to build the patterns, checking their frequency in the logs.
Extending these mining techniques, the local process mining approach of Niek Tax et al.~\cite{tax2016mining} discovers frequent patterns from the logs providing support to loops.
Finally, in~\cite{bui2012framework} tree structured patterns in the XML structure of the XES\footnote{XES is an XML-based standard for event logs. Its purpose is to provide a generally-acknowledged format for the interchange of event log data between tools and application domains (\url{http://www.xes-standard.org/}).} logs are searched.
Nevertheless, as we will show in this section, all these techniques fail to measure the frequency of a pattern in some cases, and specially when the model presents loops or optional tasks\footnote{In this paper we will refer as optional tasks to the tasks of a selection (choice) where one of the branches has no tasks, leaving the other as optional.}.


The heat maps approach performs a highlight of the arcs and tasks of a process model depending on their individual frequency, i.e., the number of executions.
To obtain the frequent patterns of a model using heat maps, the arcs which frequency exceeds a defined threshold are retrieved.
The problem is that the individual frequency does not consider the causality between tasks.
A highlighted structure can have all its arcs individually frequent, i.e., each arc is executed individually in a percentage of traces considered frequent, but it does not have to be necessarily frequent, i.e., the highlighted structure is not executed completely in a percentage of traces considered frequent.

\begin{figure}[t]
    \centering
    \subfloat[Heat Maps example: C-net with the arcs highlighted depending on their absolute frequency.]{
        \includegraphics[width=0.46\textwidth]{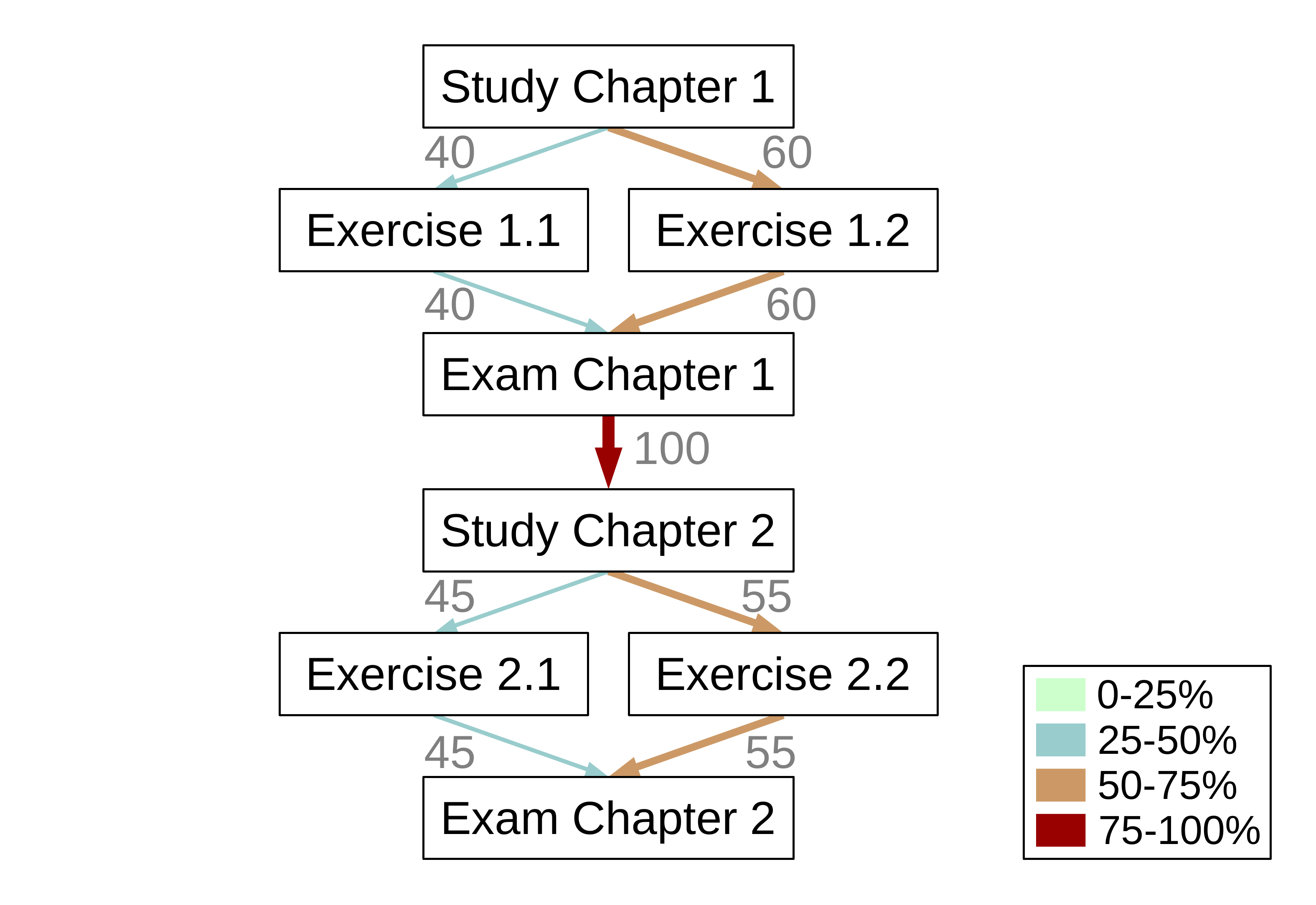}
        \label{fig:motivational-example-heat}
    }
    \hfill
    \subfloat[WoMine example: Model with a frequent pattern (40\%) highlighted.]{
        \includegraphics[width=0.46\textwidth]{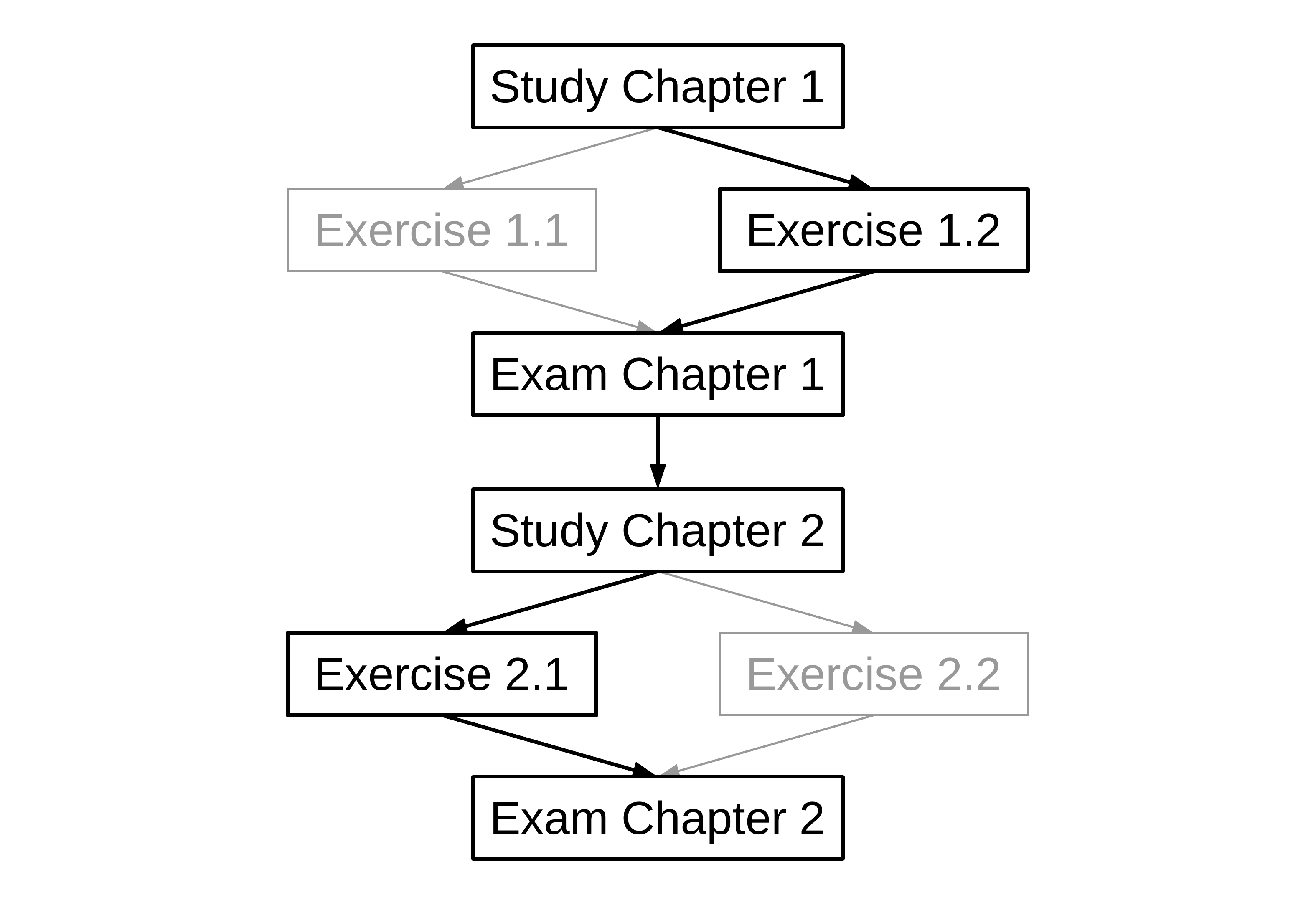}
        \label{fig:motivational-example-pattern}
    }
    \caption{Process model of a simple process in the education domain.}
    \label{fig:motivational-example}
\end{figure}

As an example, the process model in Fig.~\ref{fig:motivational-example} is provided.
If we prune on 50\% ---retrieve the arcs with a frequency over 50--- the model in Fig.~\ref{fig:motivational-example-heat}, the path (\textit{'Study Chapter 1'}, \textit{'Exercise 1.2'}, \textit{'Exam Chapter 1'}, \textit{'Study Chapter 2'}, \textit{'Exercise 2.2'}, \textit{'Exam Chapter 2'}), is obtained.
Conversely, if WoMine searches with a threshold of 40\%, this behavioural structure is not among the results ---because the individual arcs are frequent, but the sequence is not, i.e., the students solving \textit{'Exercise 1.2'} and those doing \textit{'Exercise 2.2'} are not the same.
Instead, with WoMine, the structure of Fig.~\ref{fig:motivational-example-pattern} is obtained, providing the information that in 40 traces of 100, the students select exercises 1.2 and 2.1.
Besides, the 88.88\% ---40 out of 45--- of the students who choose the exercise 2.1 came from exercise 1.2.
This behaviour can hint a predilection in the students who solve the exercise 1.2 to choose the exercise 2.1.


As can be seen, heat maps cannot find frequent structures to identify real common behaviour in processes.
There are approaches that retrieve frequent patterns measuring the frequency of the whole structure~\cite{agrawal1995mining,bui2012framework,han2001prefixspan,leemans2014discovery,mannila1997discovery}.
Some of these techniques are based on sequential pattern mining (SPM), and search subsequences of tasks with a high frequency in large sequences~\cite{agrawal1995mining,han2001prefixspan}.
One of the first approaches was proposed by Agrawal et al.~\cite{agrawal1995mining}, preceded by techniques to retrieve association rules between itemsets~\cite{agrawal1993mining}.
Most of the designed sequential pattern mining techniques present an expensive candidate generation and testing, which induces a long runtime in complex cases.
To compare the features of SPM against other discussed in this paper, we will use PrefixSpan~\cite{han2001prefixspan}, which performs a pattern-growth mining with a projection to a database based in frequent prefixes, instead of considering all the possible occurrences of frequent subsequences.
The main drawback of the sequential pattern mining based approaches is the simplicity of the patterns mined ---sequences of tasks.
Structures like concurrences or selections are treated as different sequences depending on the order of the tasks.
Also, in a retrieved frequent sequence, the execution of the $i$-th task is not ensured to be caused by the $i-1$-th task in all the occurrences in the log ---SPM only checks if the tasks of the sequence appear in the trace in the same order.


The episode mining based approaches appear to improve the results of SPM.
The first reference to episode mining is done by Mannila et al.~\cite{mannila1997discovery}.
An episode is a collection of tasks occurring close to each other.
Their algorithm uses \textit{windows} with a predefined width to extract frequent episodes, being able to detect episodes with sequences and concurrency.
In this approach, an episode is frequent when it appears in many different windows.
Based on this, Leemans et al. have designed an algorithm~\cite{leemans2014discovery} to extract association rules from frequent episodes measuring their frequency with the instances of the process, instead of using windows in the task sequences.
For instance, an episode with a frequency of 50\% has been executed in a half of the recorded traces of the process model.
One of the drawbacks of episode mining based techniques, which this paper tries to tackle, is that its search is not based in the model and, thus, it does not take advantage of the relation among tasks presented in it.
For the same frequent behaviour, the algorithm extracts various patterns with the same tasks, but with different relations among them, making difficult the extraction of information.

\begin{figure}[t]

    \begin{minipage}[c]{.45\textwidth}
    
        \subfloat[Petri net with two parallel branches, one with a loop and the other one with an optional task.\label{fig:why-topological-wrong-workflow}]{
            \includegraphics[width=0.9\textwidth]{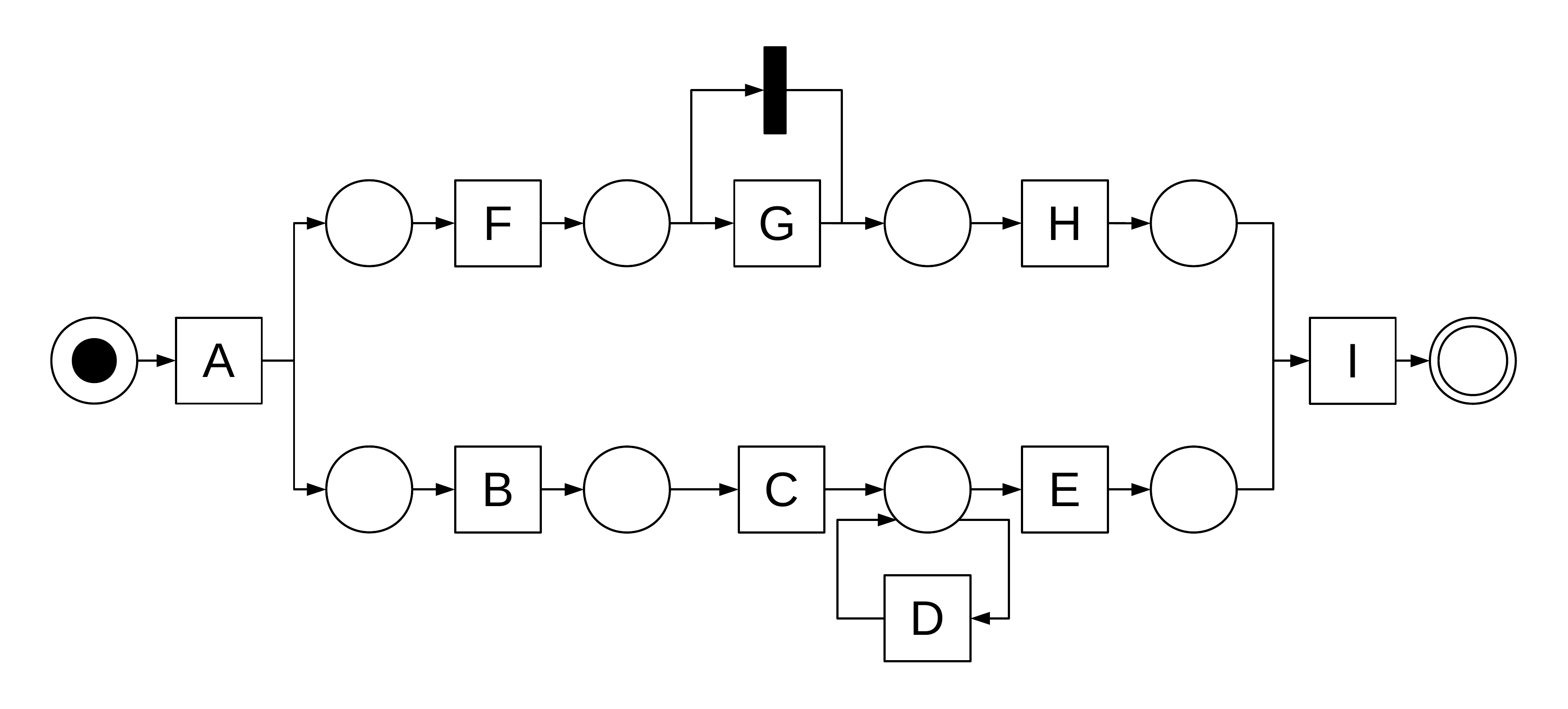}
        }
    
    \end{minipage}   
    \begin{minipage}[c]{.55\textwidth}
        \begin{tabular}{cc}
        
        \subfloat[\label{fig:why-topological-wrong-pattern-1}]{
            \centering
            \includegraphics[width=0.45\textwidth]{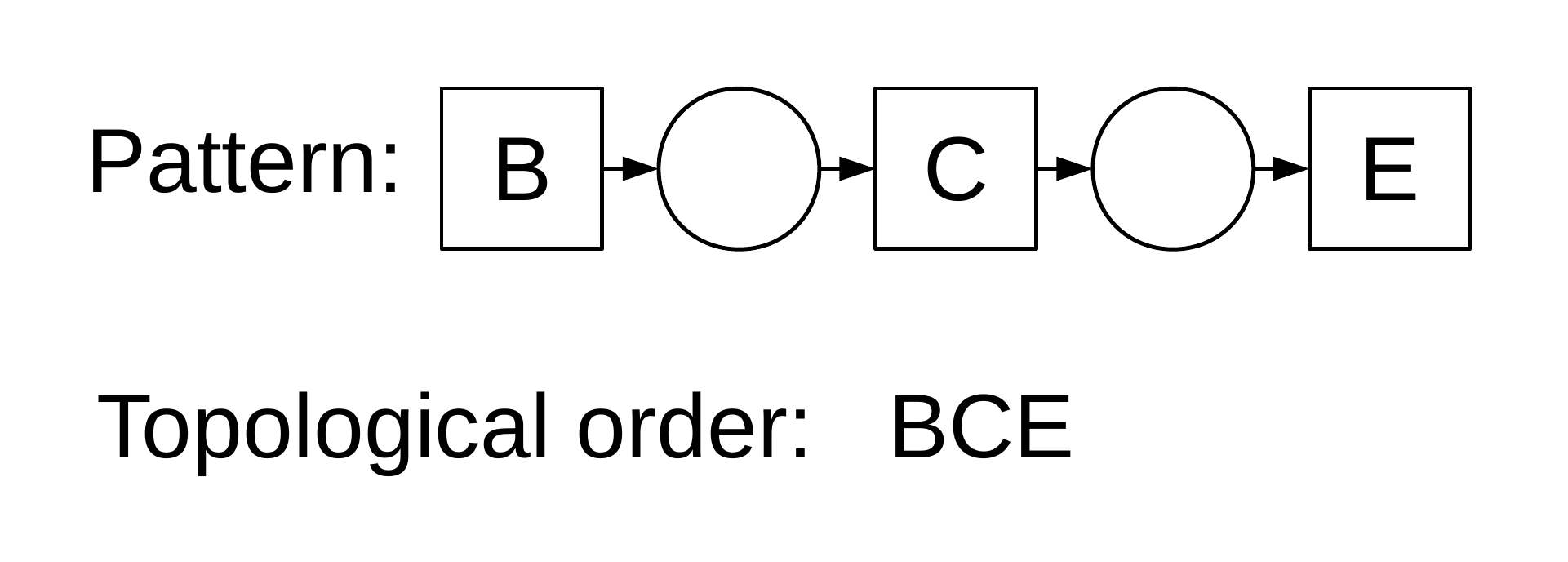}
        }
        
        &
        
        \subfloat[\label{fig:why-topological-wrong-table-1}]{
            {\scriptsize
            \centering
            \begin{tabular}[b]{|l|c|c|}
                \hline
                \multicolumn{1}{|c|}{\textbf{Trace}}    & \textbf{det}         & \textbf{exec}        \\ \hline
                AF\textbf{BCE}HI                        & \cmark               & \cmark               \\ \hline
                A\textbf{BC}F\textbf{E}HI               & \cmark               & \cmark               \\ \hline
                AFGH\textbf{BC}D\textbf{E}I            & \cmark               & \xmark               \\ \hline
            \end{tabular}
            }
        }
        \\
        
        \subfloat[\label{fig:why-topological-wrong-pattern-2}]{
            \centering
            \includegraphics[width=0.45\textwidth]{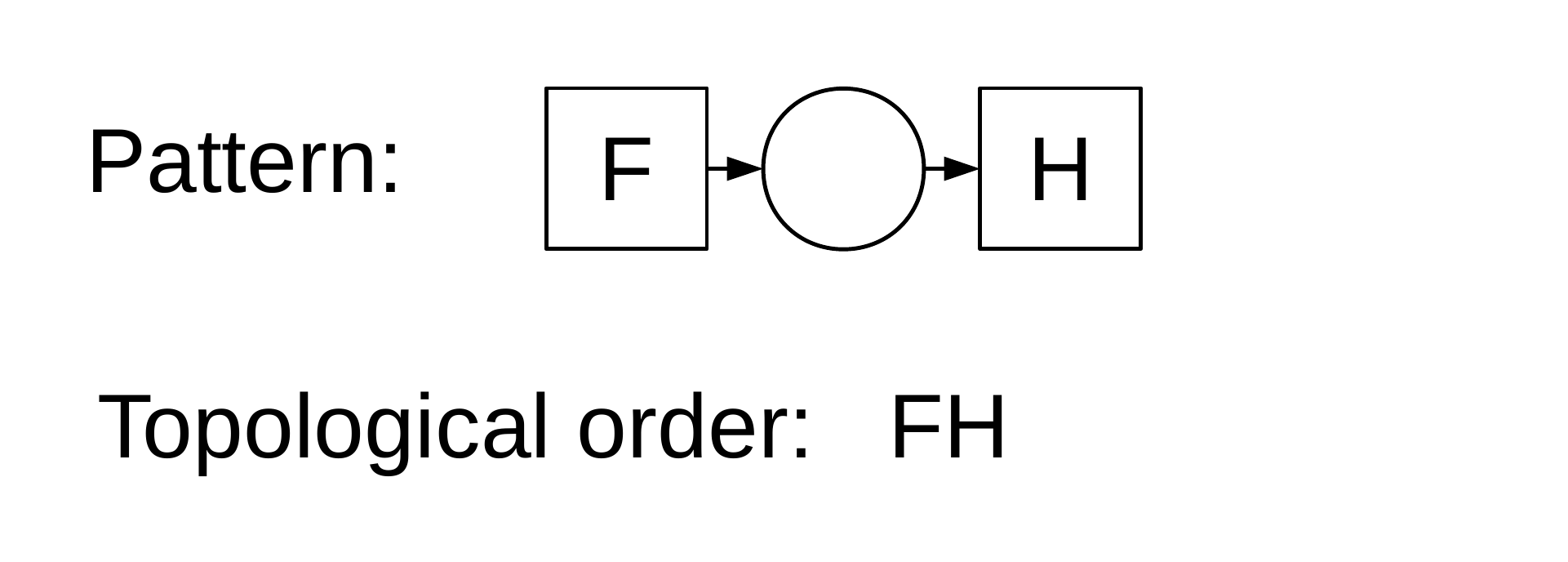}
        }
        
        & 
                          
       \subfloat[\label{fig:why-topological-wrong-table-2}]{
            {\scriptsize
            \centering
            \begin{tabular}[b]{|l|c|c|}
                \hline
                \multicolumn{1}{|c|}{\textbf{Trace}}    & \textbf{det}         & \textbf{exec}        \\ \hline
                AB\textbf{FH}CEI                        & \cmark               & \cmark               \\ \hline
                A\textbf{F}BC\textbf{H}EI               & \cmark               & \cmark               \\ \hline
                A\textbf{F}GB\textbf{H}CEI              & \cmark               & \xmark               \\ \hline
            \end{tabular}
            }
        }

        \end{tabular}
    \end{minipage}

    \caption{Model and examples to show the problems with the use of the topological order of a pattern to measure its frequency.}
    \label{fig:why-topological-wrong}
\end{figure}


To take advantage of the knowledge generated by the discovery algorithm, the frequent structures can be built based on the process model.
Greco et al. have developed an algorithm which mines frequent patterns in process models, $w$-find~\cite{greco2006mining} ---the algorithm on which WoMine is based.
This approach uses the model to build patterns compliant to the model, reducing the search space and measuring their frequency with the set of instances of the log.
Thus, using an a priori approach which grows the frequent patterns, this algorithm retrieves structures from the model, which are executed in a high percentage of the instances.
The drawback of this approach is the simplicity of the mined structures.
The patterns cannot contain selections and, thus, it will never retrieve patterns with loops.


Previous techniques ---SPM, episode mining and $w$-find--- measure the frequency of the structures checking for the topological order of the structure in the traces.
But this method is not able to detect correctly the execution of a pattern when loops or optional tasks appear.
For a better understanding the example shown in Fig.~\ref{fig:why-topological-wrong} is presented.
The pattern shown in Fig.~\ref{fig:why-topological-wrong-pattern-1} represents the bottom branch of the model in Fig.~\ref{fig:why-topological-wrong-workflow} without the loop.
In Fig.~\ref{fig:why-topological-wrong-table-1} three examples of traces are presented.
\textit{Exec} stands for the real execution of the pattern in the trace, while \textit{det} is positive when the topological order appears in the trace.
A foreign task in the middle of the topological order does not invalid the detection because it can be from the execution of the other parallel branch ---trace 2---.
This pattern is correctly executed and detected in the two first traces.
However, when the loop is executed ---third trace--- disrupting the execution of the pattern, its topological order still appears in the trace and, thus, the pattern is detected incorrectly.
The same problem occurs with the pattern shown in Fig.~\ref{fig:why-topological-wrong-pattern-2}, which is disrupted in the third trace with the execution of \texttt{G} in the middle.


Another approach, called local process mining, is presented in~\cite{tax2016mining}. In this approach a discovery of simple process models representing frequent behaviour is performed, instead of a complex process model representing all the behaviour.
In local process mining, tree models are built by performing an iterative growing process, starting with single tasks, and adding different relations with other tasks.
The evaluation of the frequency is done with an alignment-based method which, starting with an initial marking, considers that the model is executed when the final marking is reached.
This leads to one of the drawbacks of this technique; when a model contains loops or selections, the evaluation counts the model as executed even if the loop, or a choice of the selection, has not been executed in that trace.


Finally, the approach presented by Bui et al.~\cite{bui2012framework} uses logs in a XES format, and performs a search of tree structures in the XML structure of the log.
This approach builds a tree with the characteristics of each trace, and uses tree mining techniques to search frequent structures.
The information retrieved is a frequent subset of tasks and common characteristics of the XES structure.
A drawback of this approach, as with SPM algorithms, is that the retrieved patterns can only ensure the order of the tasks, but not the relation between them.


In summary, the extraction of frequent patterns could be done highlighting the frequent elements ---heat maps--- of the process model and pruning them, but without ensuring the real frequency of the results.
Sequential pattern mining could be used to retrieve real frequent patterns but the structures are limited to sequences, and it only ensures the precedence between the tasks.
The approaches based on episode mining~\cite{leemans2014discovery,mannila1997discovery} and frequent pattern mining~\cite{greco2006mining} retrieves structures that are still simple for real processes, and the measure of the frequency presents problems when the process model has loops or optional tasks.
Finally, local process mining provides similar results, with the addition of loops and choices to the extracted models.
Other types of search techniques have also been applied like mining tree structures but, as sequential pattern mining, the tasks of the retrieved patterns have only sequences.

\begin{table}[b!]
\centering
\begin{tabular}{llcccccc}

                                                                              &   & \rot{Mine from model} & \rot{Expl. proc. instances} & \rot{Mine sequences} & \rot{Mine parallels} & \rot{Mine choices} & \rot{Mine loops} \\ \hline
\multicolumn{1}{|l|}{Sequential Pattern Mining - Based Algorithms~\cite{han2001prefixspan}}   & \multicolumn{1}{l|}{} & \multicolumn{1}{c|}{-}   & \multicolumn{1}{c|}{-}   & \multicolumn{1}{c|}{+}   & \multicolumn{1}{c|}{-}   & \multicolumn{1}{c|}{-}   & \multicolumn{1}{c|}{-}   \\ \hline
\multicolumn{1}{|l|}{Mannila's Episode mining~\cite{mannila1997discovery}}   & \multicolumn{1}{l|}{} & \multicolumn{1}{c|}{-}   & \multicolumn{1}{c|}{-}   & \multicolumn{1}{c|}{+}   & \multicolumn{1}{c|}{+}   & \multicolumn{1}{c|}{-}   & \multicolumn{1}{c|}{-}   \\ \hline
\multicolumn{1}{|l|}{Bui's Tree Mining~\cite{bui2012framework}}  & \multicolumn{1}{l|}{} & \multicolumn{1}{c|}{-}   & \multicolumn{1}{c|}{+}   & \multicolumn{1}{c|}{+}   & \multicolumn{1}{c|}{-}   & \multicolumn{1}{c|}{-}   & \multicolumn{1}{c|}{-}   \\ \hline
\multicolumn{1}{|l|}{Leemans' Episode discovery~\cite{leemans2014discovery}}  & \multicolumn{1}{l|}{} & \multicolumn{1}{c|}{-}   & \multicolumn{1}{c|}{+}   & \multicolumn{1}{c|}{+}   & \multicolumn{1}{c|}{+}   & \multicolumn{1}{c|}{-}   & \multicolumn{1}{c|}{-}   \\ \hline
\multicolumn{1}{|l|}{$w$-find~\cite{greco2006mining}} & \multicolumn{1}{l|}{} & \multicolumn{1}{c|}{+}   & \multicolumn{1}{c|}{+}   & \multicolumn{1}{c|}{+}   & \multicolumn{1}{c|}{+}   & \multicolumn{1}{c|}{-}   & \multicolumn{1}{c|}{-}   \\ \hline
\multicolumn{1}{|l|}{Tax's Local Process Model~\cite{tax2016mining}} & \multicolumn{1}{l|}{} & \multicolumn{1}{c|}{-}   & \multicolumn{1}{c|}{+}   & \multicolumn{1}{c|}{+}   & \multicolumn{1}{c|}{+}   & \multicolumn{1}{c|}{$\pm$}   & \multicolumn{1}{c|}{$\pm$}   \\ \hline
\multicolumn{1}{|l|}{\textbf{WoMine (this publication)}}   & \multicolumn{1}{l|}{} & \multicolumn{1}{c|}{\textbf{+}}   & \multicolumn{1}{c|}{\textbf{+}}   & \multicolumn{1}{c|}{\textbf{+}}   & \multicolumn{1}{c|}{\textbf{+}}   & \multicolumn{1}{c|}{\textbf{+}}   & \multicolumn{1}{c|}{\textbf{+}}   \\ \hline

\end{tabular}
\caption{Feature comparison of discussed algorithms. \textit{'Mine from model'} is marked if the algorithm uses the process model to retrieve the patterns, basing the search on the relations in the model. \textit{'Expl. proc. instances'} indicates if the algorithm uses the traces of the log to measure the frequency of the patterns being a 100\% the apparition in all traces. \textit{'Mine sequences'}, \textit{'parallels'}, \textit{'choices'} and \textit{'loops'} indicate if the algorithm retrieves frequent patterns with sequences, parallels, choices or loops, respectively. Finally \textit{'+'} stands for a complete support to the feature, \textit{'-'} stands for a non support and \textit{'$\pm$'} stands for a partial support to the feature for the purpose of this paper.}
\label{tab:feature-comparison}
\end{table}


As far as we know, the $w$-find is the only algorithm that searches substructures in the process model, checking the frequency in the traces of the log.
The algorithm presented in this paper, WoMine, realizes an a priori search based on the $w$-find search, being able to get frequent subprocesses in models with loops and optional tasks, ensuring the frequency of each pattern and retrieving structures with sequences, parallels, selections and loops.
A comparison between these algorithms is presented in Table~\ref{tab:feature-comparison}.

    \section{Preliminaries\label{sec:preliminaries}}

In this paper, we will represent the examples with place/transition Petri nets~\cite{desel1998place} due to its higher comprehensibility, and the easiness to explain the behaviour of the execution.
Nevertheless, our algorithm represents the process with a Causal Matrix (Def.~\ref{dfn:causalM1}).

\begin{figure}[H]
    \begin{minipage}{0.49\linewidth}
    \centering
        \subfloat[Petri net, with XOR and AND structures.]{
            \includegraphics[width=0.95\textwidth]{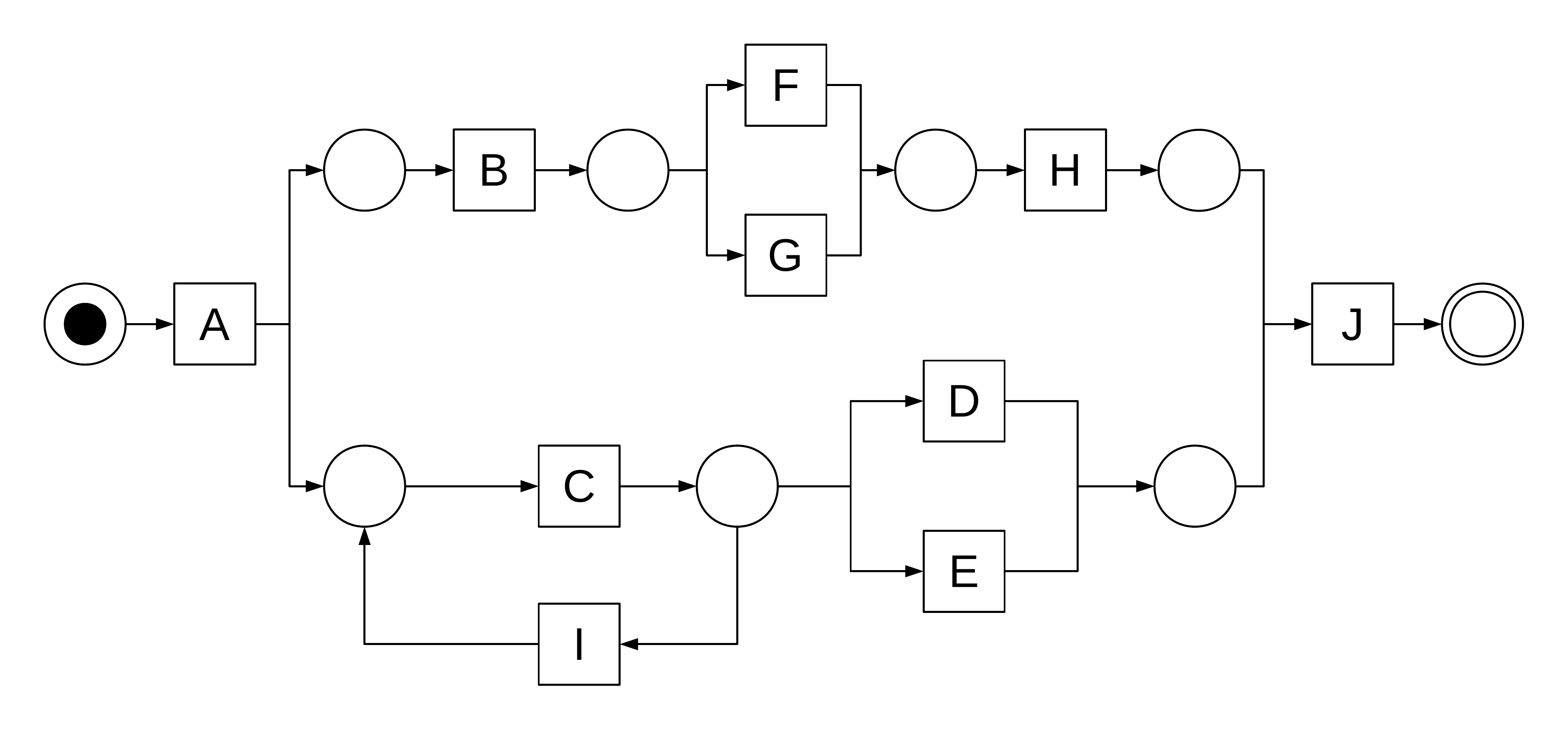}
            \label{fig:petri-net-example}
        }
        \hfill
        \subfloat[Trace of the process model in Fig.~\ref{fig:petri-net-example}.\label{tab:example-trace}]{
            \begin{tabular}{p{1.0\linewidth}}
                \centering
                A C B I C I C F H E J
            \end{tabular}
        }
    \end{minipage}
    \begin{minipage}{0.49\linewidth}
    \centering
        \subfloat[Causal matrix of the process model in Fig.~\ref{fig:petri-net-example}.]{
            \begin{tabular}[b]{|l|l|l|}
            \hline
            \multicolumn{1}{|c|}{\textbf{Task}} & \multicolumn{1}{c|}{\textbf{I(Task)}} & \multicolumn{1}{c|}{\textbf{O(Task)}} \\ \hline
            \texttt{A}                                   & \texttt{\{\}}                              & \texttt{\{\{B,C\}\}}            \\ \hline
            \texttt{B}                                   & \texttt{\{\{A\}\}}                         & \texttt{\{\{F\},\{G\}\}}        \\ \hline
            \texttt{C}                                   & \texttt{\{\{A\},\{I\}\}}                   & \texttt{\{\{D\},\{E\},\{I\}\}}  \\ \hline
            \texttt{D}                                   & \texttt{\{\{C\}\}}                         & \texttt{\{\{J\}\}}              \\ \hline
            \texttt{E}                                   & \texttt{\{\{C\}\}}                         & \texttt{\{\{J\}\}}              \\ \hline
            \texttt{F}                                   & \texttt{\{\{B\}\}}                         & \texttt{\{\{H\}\}}              \\ \hline
            \texttt{G}                                   & \texttt{\{\{B\}\}}                         & \texttt{\{\{H\}\}}              \\ \hline
            \texttt{H}                                   & \texttt{\{\{F\},\{G\}\}}                   & \texttt{\{\{J\}\}}              \\ \hline
            \texttt{I}                                   & \texttt{\{\{C\}\}}                         & \texttt{\{\{C\}\}}              \\ \hline
            \texttt{J}                                   & \texttt{\{\{H,D\},\{H,E\}\}}               & \texttt{\{\}}                   \\ \hline
        \end{tabular}
            \label{tab:causal-matrix-example}
        }
    \end{minipage}
    \caption{Example to show the internal representation of the process models in WoMine. 
    The inputs of task \texttt{J} are composed by two paths or choices.
    One is the tuple \texttt{H} and \texttt{D}, and the other one is the tuple \texttt{H} and \texttt{E}.
    As can be seen, each subset in the set of inputs ($I($\texttt{J}$)$) corresponds to a possible path in the inputs of \texttt{J}.}
    \label{fig:petri-net-and-causal-matrix}
\end{figure}

\begin{quotation}
    \begin{defn}[Causal matrix\label{dfn:causalM1}]
    A Causal matrix is a tuple $(T,I,O)$ where:
    
    T is a finite set of tasks.
    
    I : $T\rightarrow \mathbb{P}(\mathbb{P}(T))$ is the input condition function, where $\mathbb{P}(X)$ denotes the powerset of some set $X$.
    
    O : $T\rightarrow \mathbb{P}(\mathbb{P}(T))$ is the output condition function.
    
    For a task or activity $\alpha\in T$, $I(\alpha)$ denotes a set of sets of tasks representing the inputs of $\alpha$. Each set $\varPhi\in I(\alpha)$ corresponds to a choice in the inputs of $\alpha$ ---$\lvert I(\alpha)\rvert > 1$ represents a selection, and $\lvert \varPhi\rvert > 1$ denotes a parallel input path.
    In the same way, $O(\alpha)$ denotes the set of sets of tasks representing the outputs of $\alpha$ ---$\lvert \varTheta\rvert > 1\colon \varTheta\in O(\alpha)$ denotes a parallel output path.
    Fig.~\ref{fig:petri-net-and-causal-matrix} shows an example.
    \end{defn}
\end{quotation}

\begin{quotation}
    \begin{defn}[Trace\label{def:trace}]
    Let $T$ be the set of tasks of a process model, and $\varepsilon$ an event ---the execution of a task $\alpha\in T$.
    A trace is a list (sequence) $\langle \tau = \varepsilon_{1}, ..., \varepsilon_{n}\rangle$ of events $\varepsilon_{i}$ occurring at a time index $i$ relative to the other events in $\tau$.
    Each trace corresponds to an execution of the process, i.e., a process instance.
    As an example, Fig~\ref{tab:example-trace} shows a trace that corresponds to an execution of the process model of Fig.~\ref{fig:petri-net-example}.
    \end{defn}
\end{quotation}

\begin{quotation}
    \begin{defn}[Log\label{def:log}]
    An event log $L = [\tau_{1}, ..., \tau_{m}]$ is a multiset of traces $\tau_{i}$.
    In this simple definition, the events only specify the name of the task, but usually, the event logs store more information as timestamps, resources, etc.
    \end{defn}
\end{quotation}

\begin{quotation}
    \begin{defn}[Pattern\label{def:pattern}]
    Let $C = (T, I, O)$ be a causal matrix representing a process model $W$.
    A connected subgraph $P = (T', I', O')$ is a pattern of $W$ if and only if:
    
    $T' \subseteq T$
    
    $\lvert I'\rvert = \lvert O'\rvert = \lvert T'\rvert$
    
    $\forall \alpha \in T' \to I'(\alpha) \subseteq I(\alpha), O'(\alpha) \subseteq O(\alpha)$ 
    
    \end{defn}
\end{quotation}

\begin{figure}[b]
    \centering
    \begin{minipage}[b]{0.31\linewidth}
        \centering
        \subfloat[Valid pattern with a selection and a parallel.]{
            \includegraphics[width=0.9\textwidth]{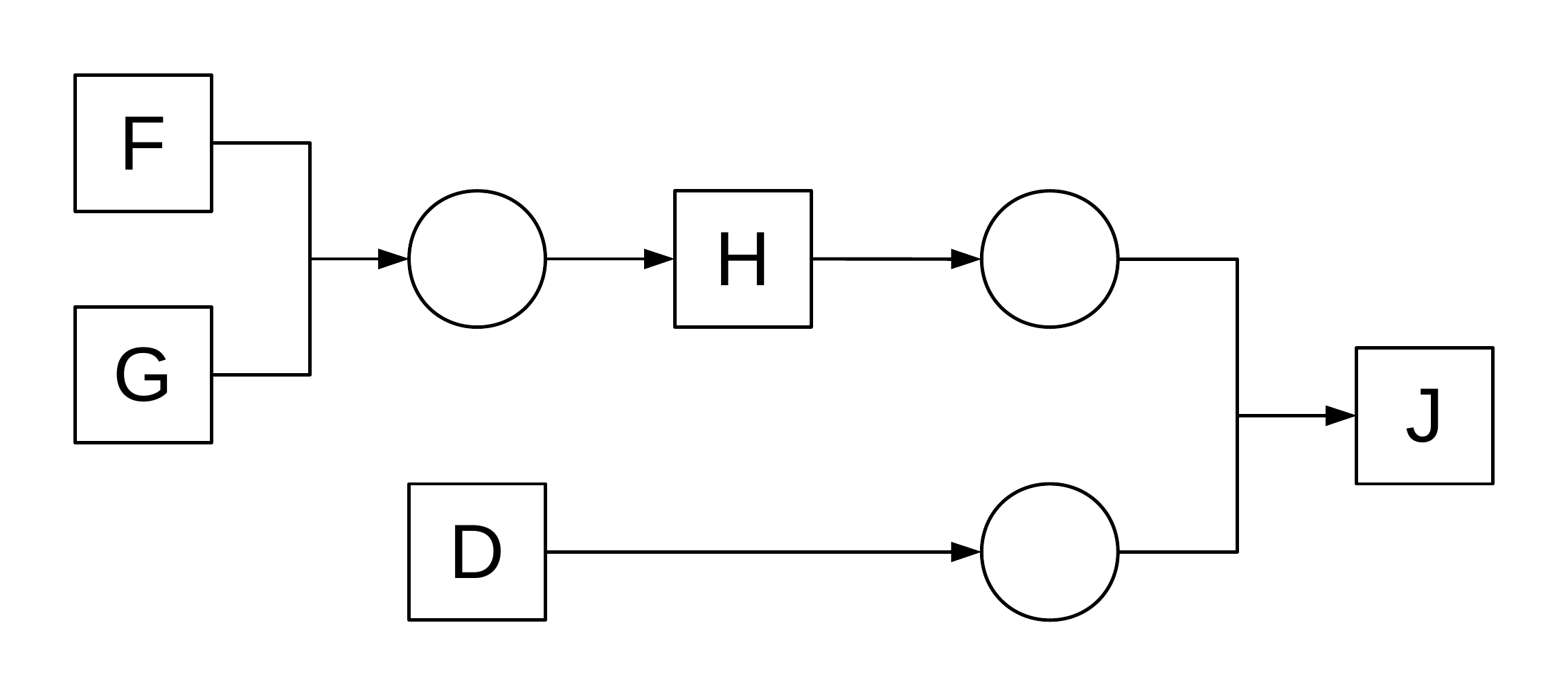}
            \label{fig:pattern-example-1}
        }
    \end{minipage}
    \hfill
    \begin{minipage}[b]{0.31\linewidth}
        \centering
        \subfloat[Valid pattern with a parallel, a selection and a loop.]{
            \includegraphics[width=0.9\textwidth]{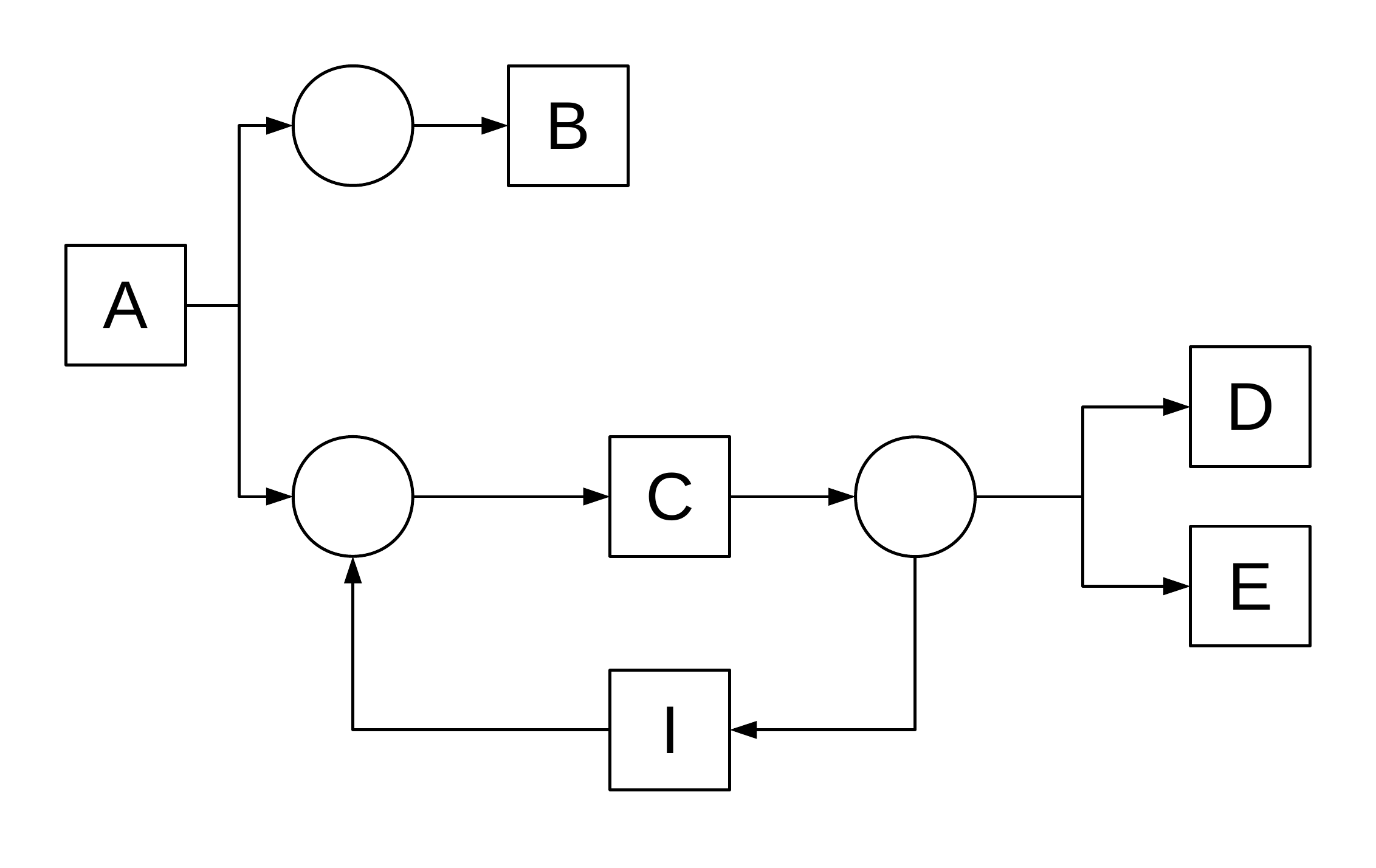}
            \label{fig:pattern-example-2}
        }
    \end{minipage}
    \hfill
    \begin{minipage}[b]{0.31\linewidth}
        \centering
        \subfloat[Invalid pattern. Task \texttt{J} has some incomplete input combinations.]{
            \includegraphics[width=0.9\textwidth]{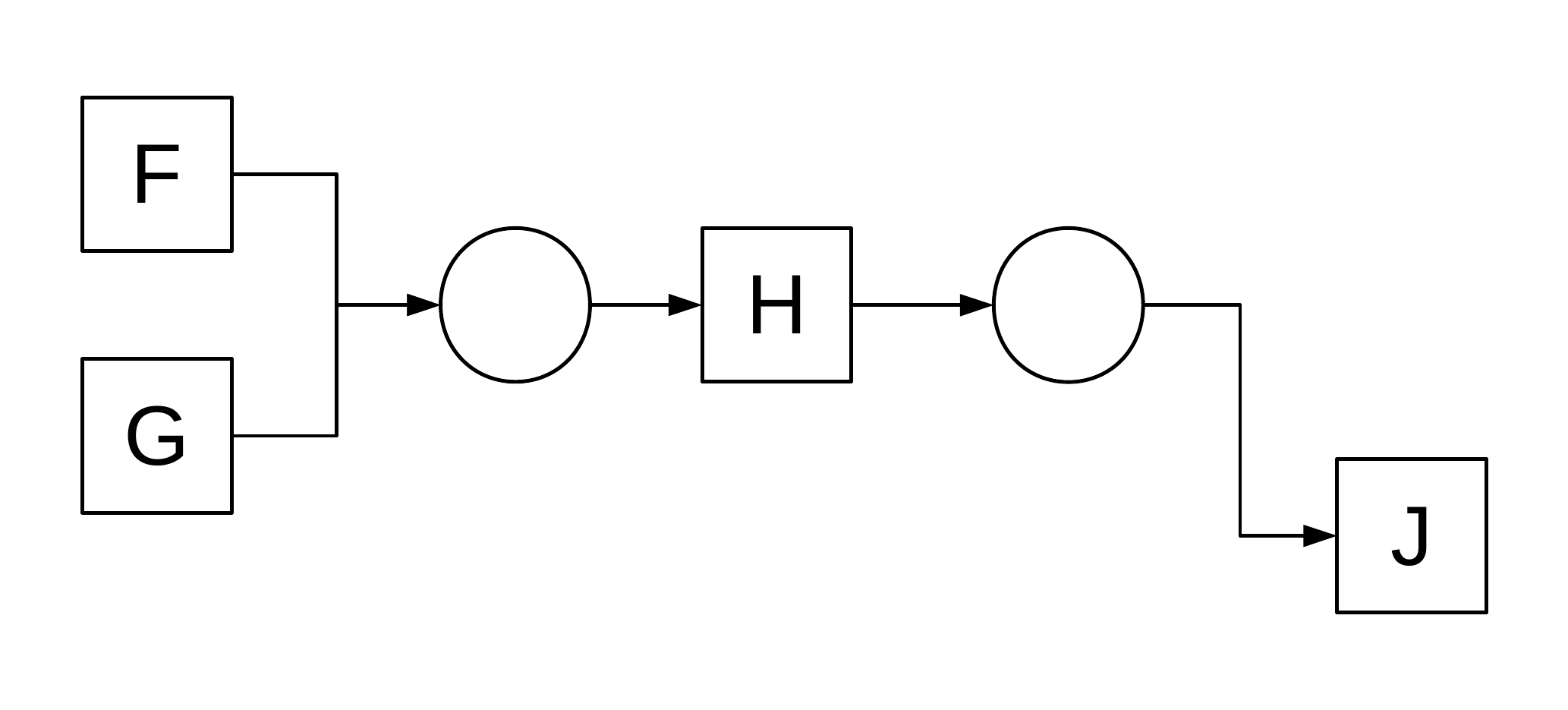}
            \label{fig:pattern-example-wrong}
        }
    \end{minipage}
    
    \caption{Examples of valid and invalid patterns of the process model shown in Fig.~\ref{fig:petri-net-and-causal-matrix}.
    All tasks have as connections a subset of their connections in the causal matrix of Table~\ref{tab:causal-matrix-example}.
    For instance, $I'(J)$ is equal to \texttt{\{\{H,D\}\}}, which is a subset of $I(J)$, \texttt{\{\{H,D\}, \{H,E\}\}}.
    This makes the structure a valid pattern.
    Meanwhile, the structure shown in Fig.~\ref{fig:pattern-example-wrong} is a wrong pattern, because task \texttt{J} has some incomplete input combinations ---\texttt{\{\{H\}\}} $\not\subseteq$ \texttt{\{\{H,D\}, \{H,E\}\}}.}
    \label{fig:pattern-and-wrong-pattern-example}
\end{figure}

A \textit{pattern} (Def.~\ref{def:pattern}) is a subgraph of the process model that represents the behaviour of a part of the process.
For each task $\alpha$ in the pattern, its inputs, $I'(\alpha)$, must be a subset of $I(\alpha)$ in the model it belongs to; and the outputs, $O'(\alpha)$, must be also a subset of $O(\alpha)$ in the model.
This ensures that a pattern has not a partial parallel connection.
Fig.~\ref{fig:pattern-and-wrong-pattern-example} shows some examples of valid and invalid patterns.

\begin{quotation}
    \begin{defn}[Simple Pattern\label{def:simple-pattern}]
    A pattern $P=(T',I',O')$ is a Simple Pattern when:
    
    $\forall \alpha \in T' \to [\exists! \varPhi \in I'(\alpha) \colon \varPhi \not\subseteq R^{+}(\alpha)] \vee [\forall \varPhi \in I'(\alpha) \colon \varPhi \subseteq R^{+}(\alpha)]$
    
    $\forall \alpha \in T' \to [\exists! \varTheta \in O'(\alpha) \colon \varTheta \not\subseteq R^{-}(\alpha)] \vee [\forall \varTheta \in O'(\alpha) \colon \varTheta \subseteq R^{-}(\alpha)]$
    
    Being $R^{+}(\alpha)$ the set of successors of a task $\alpha$, i.e., the set of tasks reachable from $\alpha$; and $R^{-}(\alpha)$ the set of predecessors of a task $\alpha$, i.e., the set of tasks that can reach $\alpha$.
    \end{defn}
\end{quotation}

\begin{figure}[t]

    \centering
    \begin{minipage}[b]{0.31\linewidth}
        \centering
        \subfloat[Valid simple pattern. The pattern is executed in the instance \texttt{[C D H J]}.]{
            \includegraphics[width=0.85\textwidth]{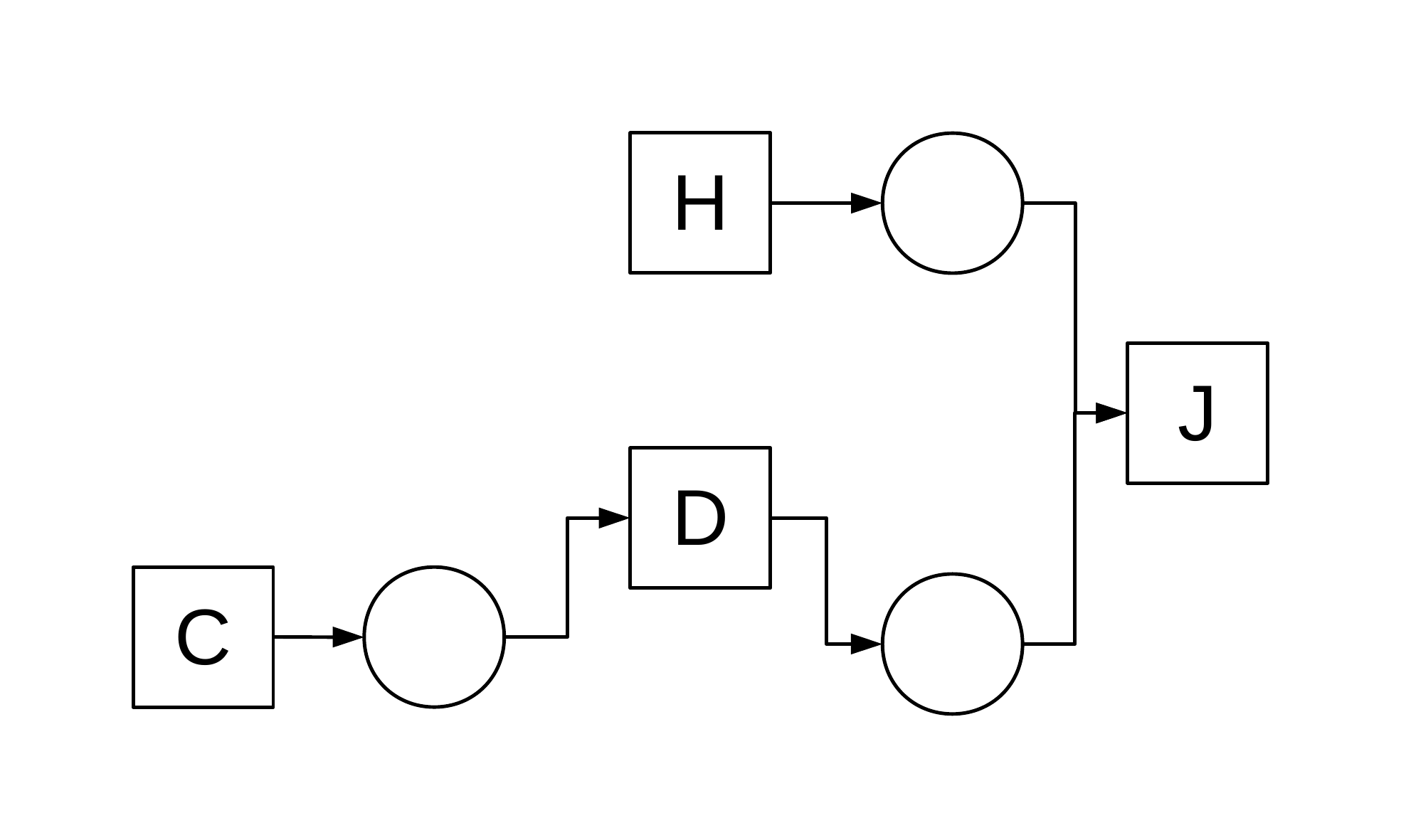}
            \label{fig:simple-pattern-example}
        }
    \end{minipage}
    \hfill
    \begin{minipage}[b]{0.31\linewidth}
        \centering
        \subfloat[Valid simple pattern with a loop. In the output paths of \texttt{C} (\texttt{\{D\}} and \texttt{\{I\}}), the only non predecessor is the path of \texttt{\{D\}}. The same happens in the inputs. The pattern is executed in the instance \texttt{[A B C I C D]}.]{
            \includegraphics[width=0.85\textwidth]{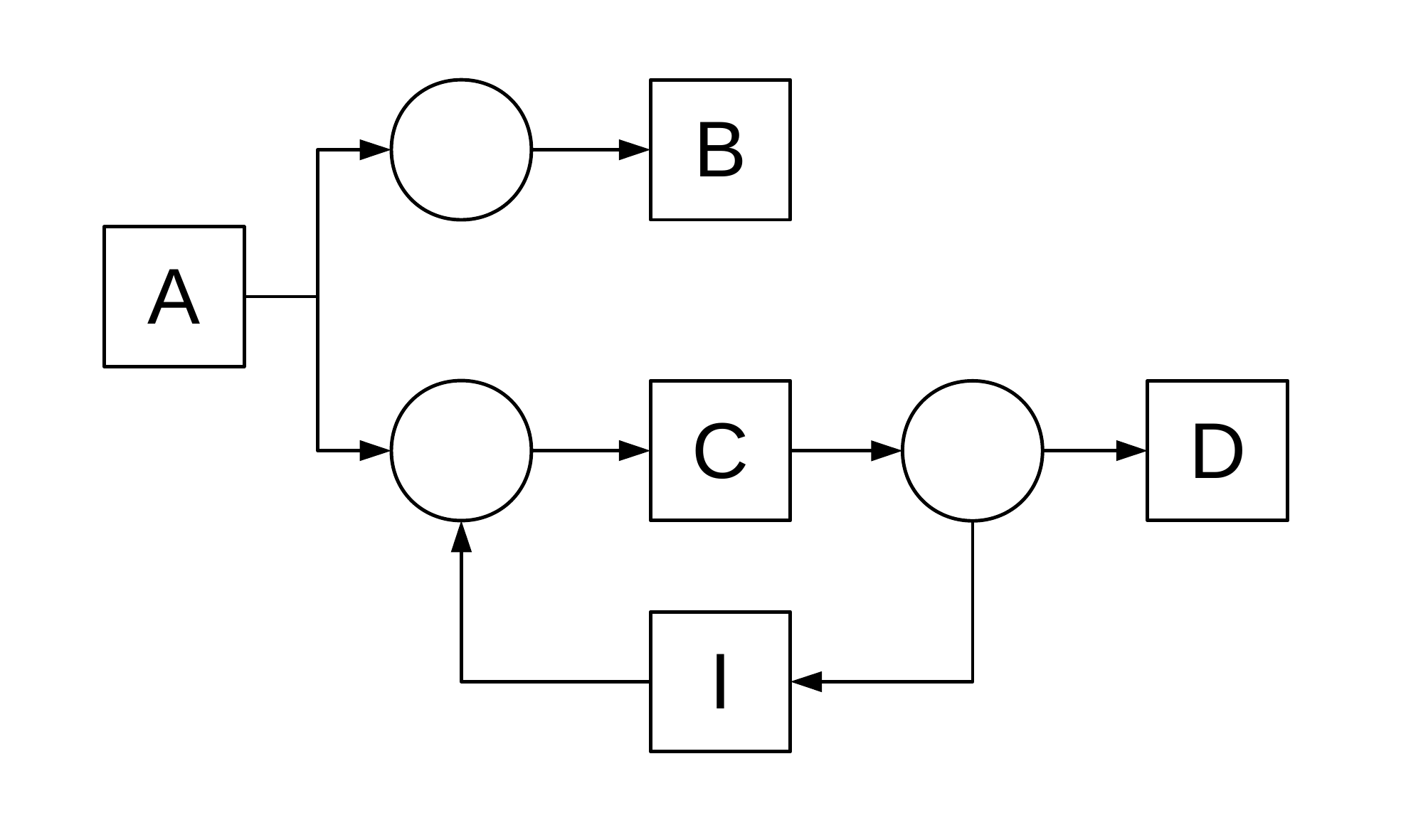}
            \label{fig:simple-pattern-example-loop}
        }
    \end{minipage}
    \hfill
    \begin{minipage}[b]{0.31\linewidth}
        \centering
        \subfloat[Invalid simple pattern. Tasks \texttt{D} and \texttt{E} cannot be in an instance of the pattern at the same time.]{
            \includegraphics[width=0.85\textwidth]{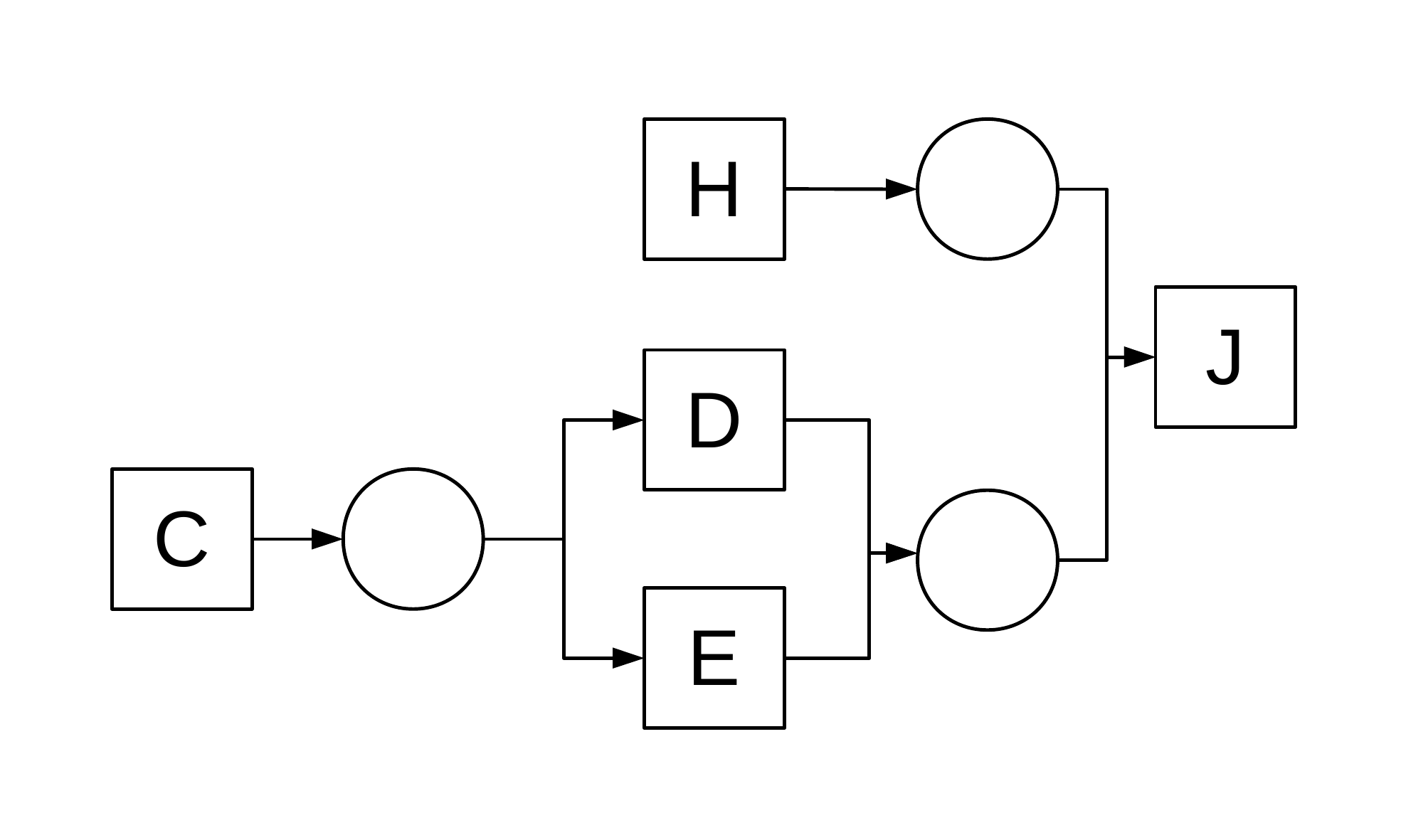}
            \label{fig:simple-pattern-example-wrong}
        }
    \end{minipage}
    
    \caption{Examples of valid and invalid simple patterns of the process model shown in Fig.~\ref{fig:petri-net-and-causal-matrix}.}
    \label{fig:simple-pattern-examples}
\end{figure}

The Simple Patterns (Def.~\ref{def:simple-pattern}) are those patterns which behaviour can be executed, entirely, in one instance.
If a task has a selection, it must be able to execute each path in the same instance.
For this, the inputs of each task $\alpha$ must have all tasks reachable from $\alpha$ except, at most, the tasks of one path.
The outputs present the same constraint, but in this case they must reach $\alpha$, not be reachable by $\alpha$.
Fig.~\ref{fig:simple-pattern-examples} shows two valid simple patterns and an invalid one.

\begin{quotation}
    \begin{defn}[Minimal Pattern, \textit{M}-pattern\label{def:minimal-pattern}]
    Each task of the process model belongs to, at least, one Minimal Pattern.
    The \textit{M}-pattern of a task $\alpha$ corresponds to the closure of $\alpha$, i.e., the structure that is going to be executed when $\alpha$ is executed.
    An exception is made with parallel structures: if $\alpha$ has a parallel in the inputs or outputs, there must be an \textit{M}-pattern with each parallel path.
    Given a causal matrix $C = (T, I, O)$ representing a process model $W$ and a task $\alpha_{1}\in T$, a simple pattern $P = (T', I', O')$ is a Minimal Pattern of $\alpha_{1}$ if all the following rules are satisfied:
    
    \vspace{4pt}
    
    {\small
    \begin{minipage}{0.47\linewidth}
    $
    \lvert I(\alpha_{1}) \rvert > 1, \forall \varPhi \in I(\alpha_{1}) \colon \lvert \varPhi \rvert = 1
        \to
    I'(\alpha_{1}) = \emptyset
    $
    
    $
    \exists \varPhi \in I(\alpha_{1}) \colon \lvert \varPhi \rvert > 1
        \to
    \lvert I'(\alpha_{1}) \rvert = \{\varPhi\}
    $
    
    $
    \forall \alpha \in T' \colon \lvert I(\alpha) \rvert = 1 
        \to 
    I'(\alpha) = I(\alpha)
    $
    
    $
    \forall \alpha \in T', \alpha \not\eq \alpha_{1}, \alpha \not\in R^{+}(\alpha_{1}), \lvert I(\alpha) \rvert > 1
        \to
    I'(\alpha) = \emptyset
    $
    \end{minipage}
    \hfill
    \begin{minipage}{0.47\linewidth}
    $
    \lvert O(\alpha_{1}) \rvert > 1, \forall \varTheta \in O(\alpha_{1}) \colon \lvert \varTheta \rvert = 1
        \to
    O'(\alpha_{1}) = \emptyset
    $
    
    $
    \exists \varTheta \in O(\alpha_{1}) \colon \lvert \varTheta \rvert > 1
        \to
    \lvert O'(\alpha_{1}) \rvert = \{\varTheta\}
    $
    
    $
    \forall \alpha \in T' \colon \lvert O(\alpha) \rvert = 1 
        \to 
    O'(\alpha) = O(\alpha)
    $
    
    $
    \forall \alpha \in T', \alpha \not\eq \alpha_{1}, \alpha \not\in R^{-}(\alpha_{1}), \lvert O(\alpha) \rvert > 1
        \to
    O'(\alpha) = \emptyset
    $
    \end{minipage}
   }
   
    \end{defn}
\end{quotation}

\begin{figure}[t]

    \begin{minipage}[c]{.59\textwidth}
    
        \centering
        
        \subfloat[Petri net of the process model.]{
            \includegraphics[width=0.95\textwidth]{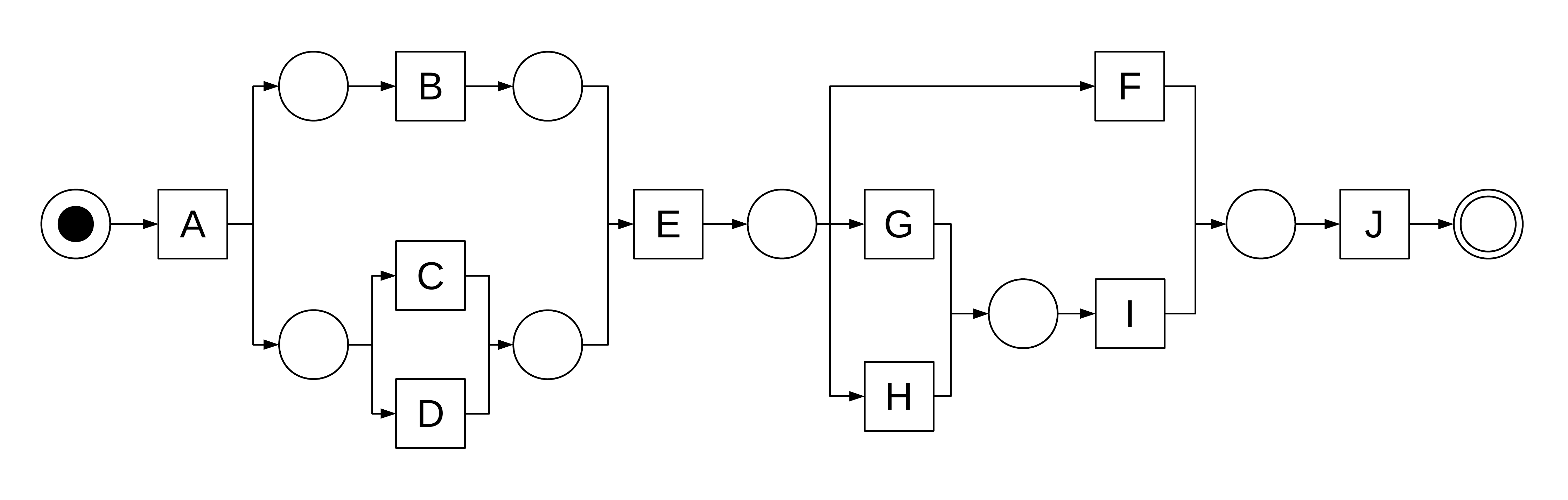}
            \label{fig:example-minimal-pattern-net}
        }
        
    \end{minipage}   
    \begin{minipage}[c]{.4\textwidth}
    
        \centering
    
        \begin{minipage}[c]{.48\textwidth}
        
            \subfloat[\textit{M}-pattern of \texttt{F}.]{
                \includegraphics[width=0.95\textwidth]{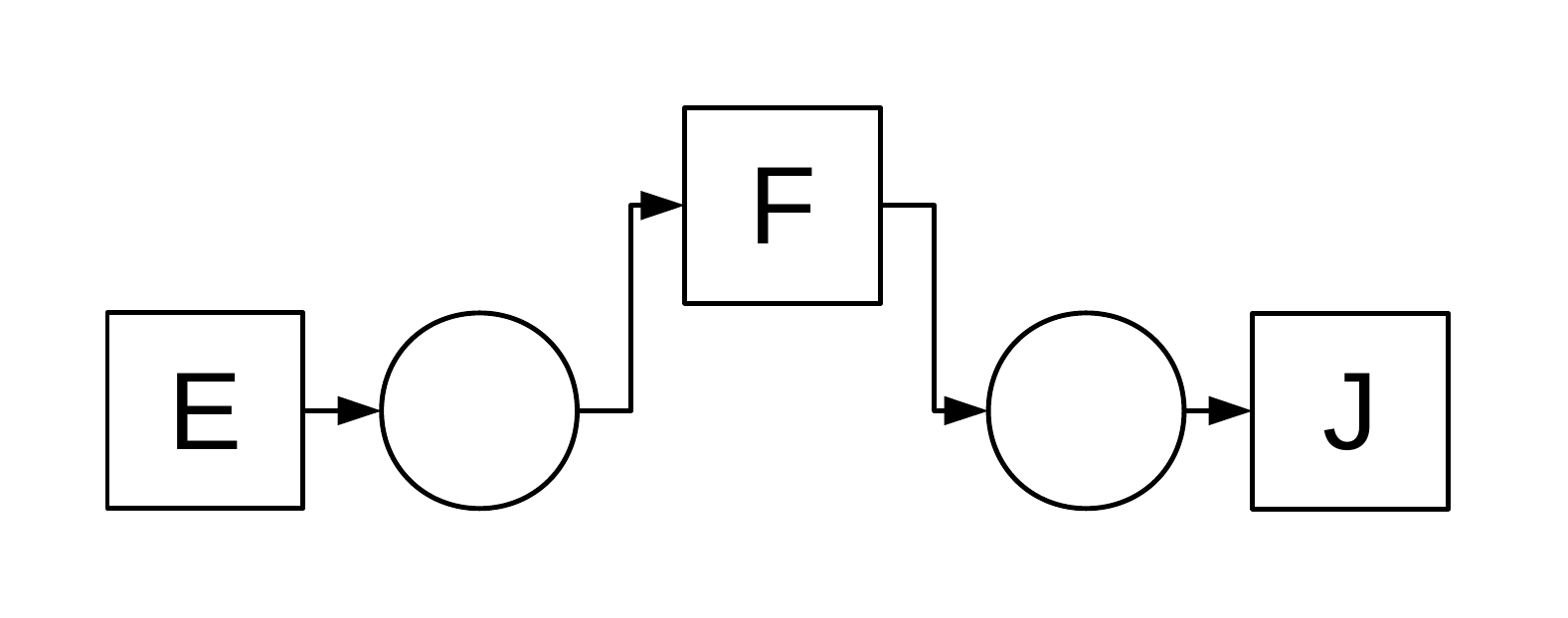}
                \label{fig:example-minimal-pattern-1}
            }
        
        \end{minipage}   
        \begin{minipage}[c]{.48\textwidth}
            
            \subfloat[\textit{M}-pattern of \texttt{J}.]{
                \includegraphics[width=0.95\textwidth]{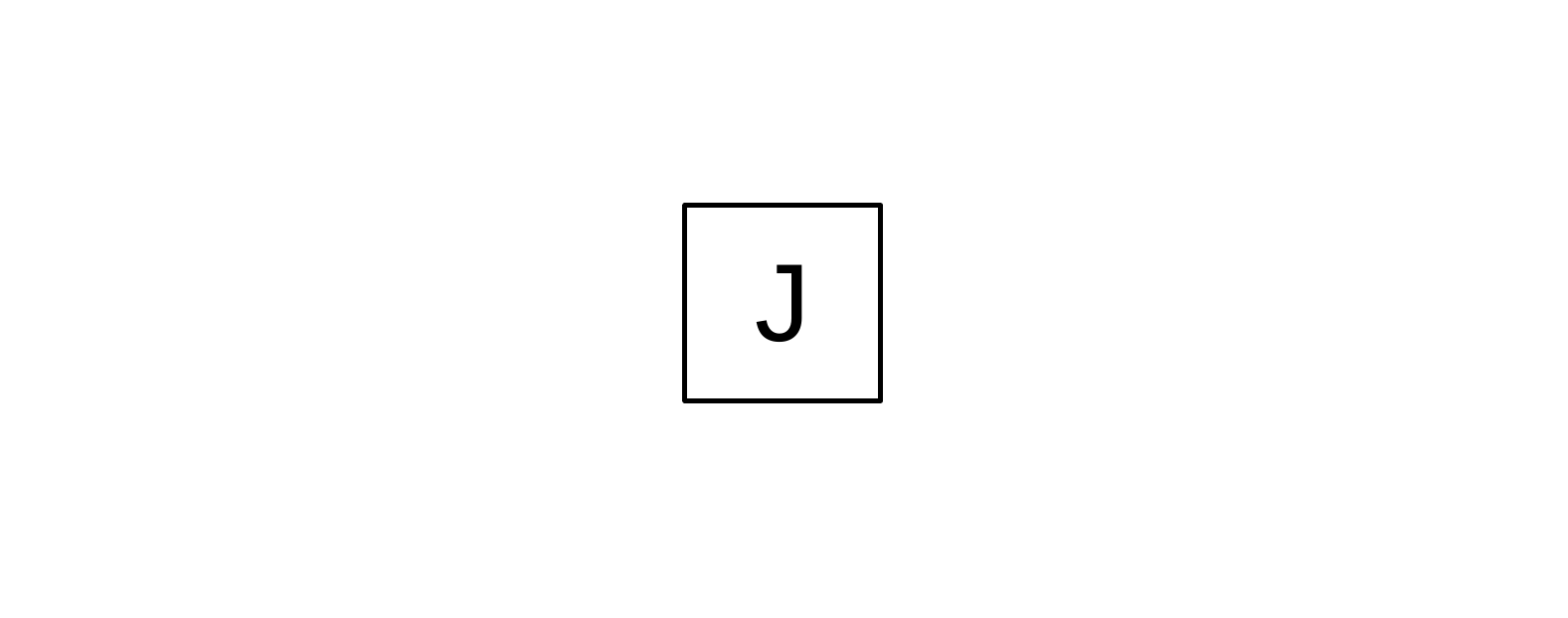}
                \label{fig:example-minimal-pattern-2}
            }
            
        \end{minipage}  
        
        \begin{minipage}[c]{.96\textwidth}
        
        \subfloat[\textit{M}-patterns of \texttt{A}.]{
            \includegraphics[width=0.95\textwidth]{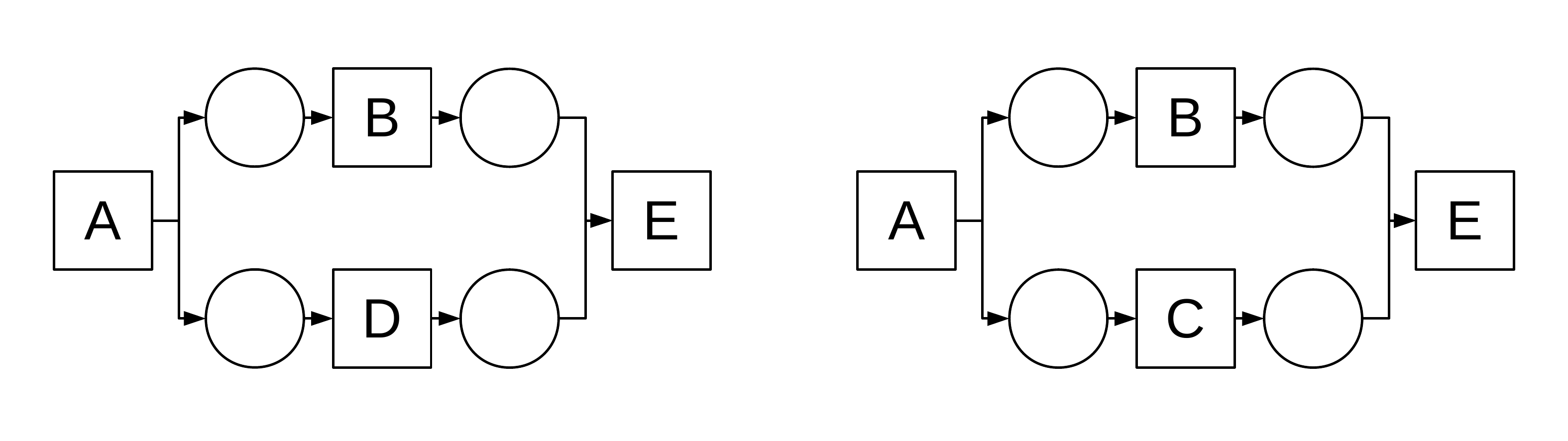}
            \label{fig:example-minimal-pattern-3}
        }
            
        \end{minipage}
        
    \end{minipage}

    \caption{A process model and three examples of \textit{M}-patterns.}
    \label{fig:example-minimal-pattern}
\end{figure}

In WoMine each task has associated, at least, one \textit{M}-pattern.
For a task $\alpha_{1}$, its \textit{M}-patterns include: \textit{i)} the structure executed always when $\alpha_{1}$ is executed ---1-path inputs and 1-path outputs---, and \textit{ii)} the parallel connections that $\alpha_{1}$ has.
If the inputs of $\alpha_{1}$ have a selection and no parallel choices, the inputs of $\alpha_{1}$ in the \textit{M}-pattern must be empty.
If $\alpha_{1}$ has a parallel connection in the inputs, there must be an \textit{M}-pattern with the parallel connection.
For each task $\alpha$ in the \textit{M}-pattern having only one possible input path, $\alpha$ must have the same path in the \textit{M}-pattern ---because this path will be always executed when $\alpha$ is executed.
Finally, for all the tasks not being reached from $\alpha_{1}$ ---not being in the successors $R^{+}(\alpha_{1})$---, except $\alpha_{1}$, if they have a selection in the inputs, they must have its inputs empty in the \textit{M}-pattern.
All these rules are similar for the outputs, but with the predecessors instead of the successors.

Fig.~\ref{fig:example-minimal-pattern} shows some \textit{M}-patterns of a model.
Fig.~\ref{fig:example-minimal-pattern-1} shows the \textit{M}-pattern of \texttt{F}: the process starts in \texttt{F} and expands the \textit{M}-pattern through \texttt{F} inputs and outputs, because both are formed by only one path.
The backwards expansion stops in \texttt{E} because its inputs are part of a selection.
Fig.~\ref{fig:example-minimal-pattern-2} depicts the \textit{M}-pattern of \texttt{J}.
It is formed only by itself, because its inputs are part of a selection and its outputs are empty.
Finally, Fig.~\ref{fig:example-minimal-pattern-3} presents the two \textit{M}-patterns of \texttt{A}.
As \texttt{A} is an AND-split with a selection, two \textit{M}-patterns are created, each one with one of the possible paths.

\begin{quotation}
    \begin{defn}[Candidate Arcs\label{def:candidate-arcs}]
    Let $C = (T, I, O)$ be a causal matrix representing a process model $W$.
    An arc $\langle \alpha_{i} \rightarrow \alpha_{j}\rangle \colon \alpha_{i}, \alpha_{j} \in T$ is part of the $A^{<}$ set, i.e., a candidate arc, if and only if:

    $\forall \varTheta \in O(\alpha_{i}) \colon \varTheta = \{\alpha_{j}\} \lor \alpha_{j} \not\in \varTheta$

    $\forall \varPhi \in I(\alpha_{j}) \colon \varPhi = \{\alpha_{i}\} \lor \alpha_{i} \not\in \varPhi$

    \end{defn}
\end{quotation}

The set of candidate arcs or $A^{<}$ is a subset of the arcs in the model which are not part of an AND structure.
For instance, all arcs of Fig.~\ref{fig:example-minimal-pattern-net}, but $\langle A \rightarrow B \rangle$, $\langle A \rightarrow C \rangle$, $\langle A \rightarrow D \rangle$, $\langle B \rightarrow E \rangle$, $\langle C \rightarrow E \rangle$ and $\langle D \rightarrow E \rangle$ are included in the $A^{<}$ set.

\begin{quotation}
    \begin{defn}[Compliance\label{def:compliance}]
        Given a trace $\tau \in L$ and a simple pattern $\mathit{SP}$ belonging to the process model, the trace is compliant with $\mathit{SP}$, denoted as $\mathit{SP} \vdash \tau$, when the execution of the trace in the process model contains the execution of the pattern, i.e., all arcs and tasks of $\mathit{SP}$ are executed in a correct order, and each task fires the execution of its outputs in the pattern.
    
    \end{defn}
\end{quotation}

\begin{figure}[t]

    \begin{minipage}[c]{.53\textwidth}
    
        \subfloat{
            \includegraphics[width=0.9\textwidth]{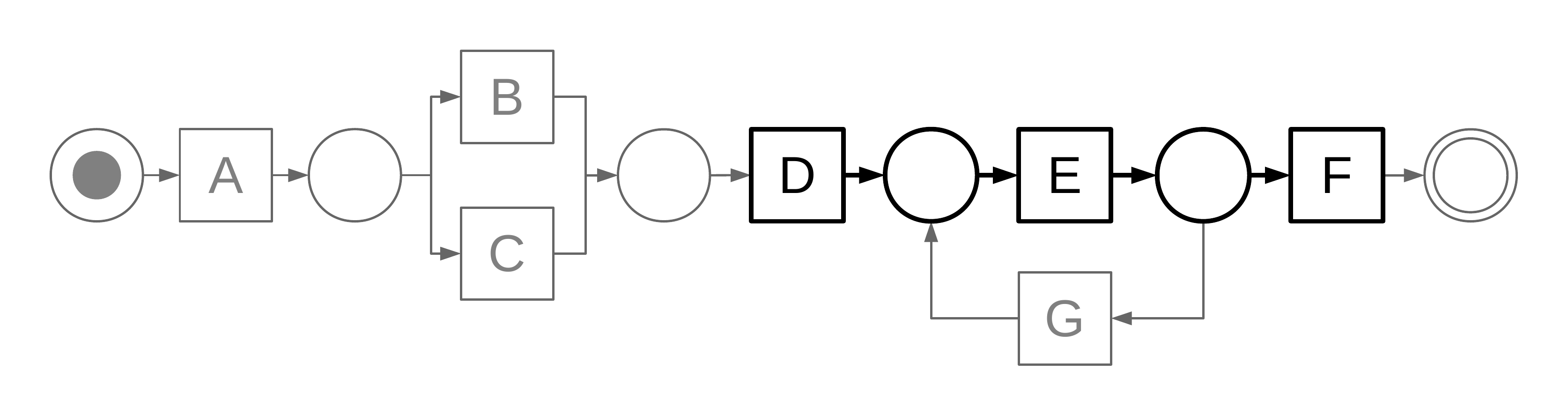}
        }
    
    \end{minipage}   
    \begin{minipage}[c]{.22\textwidth}
        
        \begin{tabular}{l}
            Trace:\\
            \texttt{A B D E G E F}
        \end{tabular}
        
    \end{minipage}   
    \begin{minipage}[c]{.24\textwidth}
        
        \begin{tabular}{l}
            Executed arcs:\\
            $\langle A \rightarrow B \rangle$, 
            $\langle B \rightarrow D \rangle$,\\
            $\langle D \rightarrow E \rangle$, 
            $\langle E \rightarrow G \rangle$,\\
            $\langle G \rightarrow E \rangle$, 
            $\langle E \rightarrow F \rangle$
        \end{tabular}

    \end{minipage}

    \caption{Process model with a pattern highlighted and a trace that is not compliant with the pattern due to the execution of the loop.}
    \label{fig:compliance}
\end{figure}

To check the compliance it is important that, while the pattern is being executed, each of its tasks triggers only the execution of its outputs.
There are cases where all the arcs and tasks of the pattern are executed in a correct order, but the execution of the pattern is disrupted in the middle of it (Fig.~\ref{fig:compliance}).
In the trace, the tasks and arcs of the pattern are executed in a proper order, but the trace is not compliant with the pattern because the sequence \texttt{D-E-F} is disrupted with the execution of \texttt{G}.

\begin{quotation}
    \begin{defn}[Frequecy of a Simple Pattern\label{def:frequency-pattern-simple-pattern}]
    Let $L$ be the set of traces of the process log.
    The frequency of a simple pattern $\mathit{SP}$ is the number of traces compliant with $\mathit{SP}$ divided by the size of the log, given by:
    
    \begin{equation}
    \mathit{freq}(\mathit{SP}) = \frac{\lvert \{ \tau \in L \colon \mathit{SP} \vdash \tau\} \lvert}{\lvert L \rvert}
    \end{equation}

    And the frequency of a pattern $P$ is the minimum frequency of the simple patterns it represents:
    
    \begin{equation}
    \mathit{freq}(P) = \min_{\forall \mathit{SP} \in P} \mathit{freq}(\mathit{SP})
    \end{equation}
    
    \end{defn}
\end{quotation}

\begin{quotation}
    \begin{defn}[Frequent Pattern\label{def:frequent-pattern}]
    Given a frequency threshold $\sigma \in \mathbb{R} \colon 0 < \sigma \leq 1$, a pattern $P$ is a frequent pattern if and only if $freq(P) \geq \sigma$.
    
    \end{defn}
\end{quotation}

    \section{WoMine\label{sec:algorithm}}

\begin{figure}[b!]
    \centering
    \includegraphics[width=0.95\textwidth]{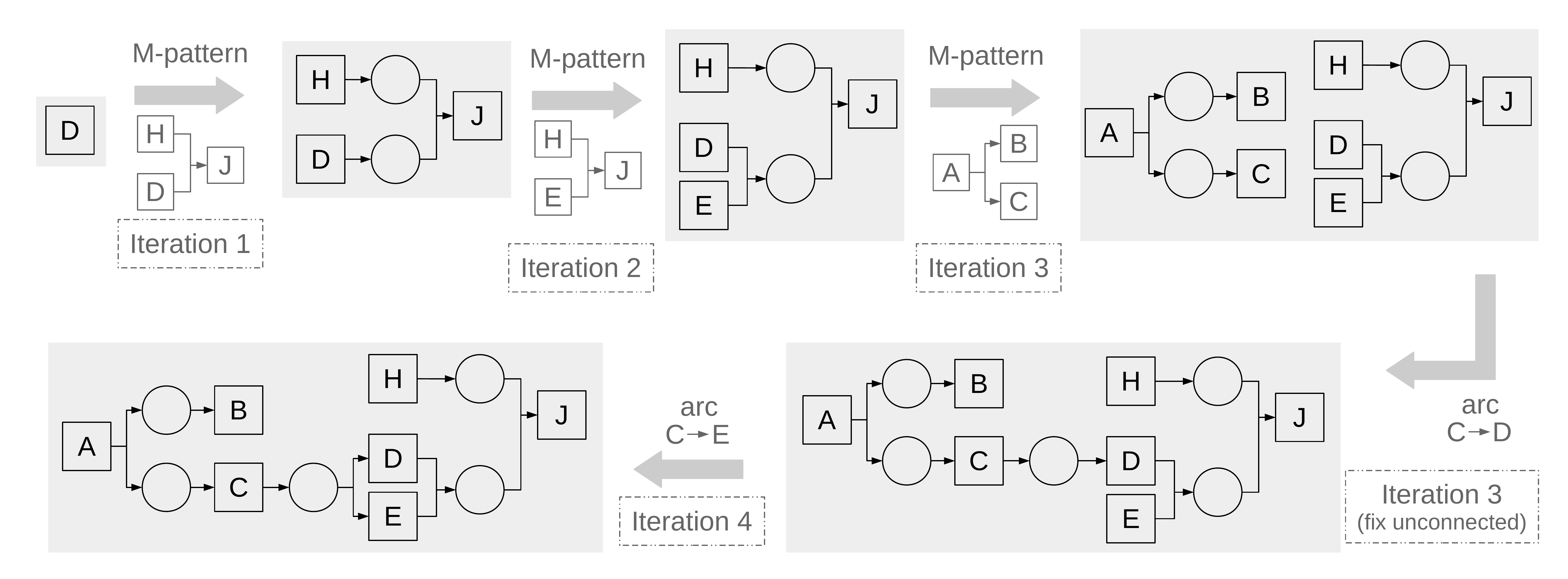}
        \caption{Example of the expansion of a pattern in 4 iterations. The process starts with the \textit{M}-pattern of \texttt{D}, being expanded with an \textit{M}-pattern of \texttt{J} in the first iteration. 
In the second iteration the new pattern is extended with another \textit{M}-pattern of \texttt{J}. 
The third iteration shows an addition of the \textit{M}-pattern of \texttt{A}, resulting in a non-connected pattern and, thus, fixed with the addition of an arc ($\langle C \rightarrow D \rangle$) which connects the two patterns.
Finally, the last iteration in the example is made with an arc ($\langle C \rightarrow E \rangle$), producing a pattern formed by two simple patterns.}
    \label{fig:example-expansion-pattern}
\end{figure}

Given a process model and a set of instances, i.e., executions of the process, the objective is to extract the subgraphs of the process model that are executed in a percentage of the traces over a threshold.
A simple approach might be a brute-force algorithm, checking the frequency of every existent subgraph inside the process model, and retrieving the frequent ones.
The computational cost of this approach makes it a non-viable option.
The algorithm presented in this paper performs an a priori search\footnote{An a priori search uses the previous knowledge, i.e., the a priori knowledge. It reduces the search space by pruning the exploration of those paths that will not finish in a valuable result.}~\cite{greco2006mining} starting with the frequent minimal patterns (Def.~\ref{def:minimal-pattern}) of the model.
In this search, there is a expansion stage done in two ways: \textit{i)} adding frequent \textit{M}-patterns not contained in the current pattern, and \textit{ii)} adding frequent arcs of the $A^{<}$ set (Def.~\ref{def:candidate-arcs}).
This expansion is followed by a pruning strategy that verifies the downward-closure property of support~\cite{agrawal1993mining} ---also known as anti-monotonicity.
This property ensures that if a pattern appears in a given number of traces, all patterns containing it will appear, at most, in the same number of traces.
Therefore, a pattern is removed of the expansion stage when it becomes infrequent, because it will never be contained again in a frequent pattern.

Fig.~\ref{fig:example-expansion-pattern} shows an example with 4 iterations of the expansion of a minimal pattern, assuming that the expanded patterns are frequent.
Although the example shows only one path of expansion, in each iteration the algorithm generates as many patterns as \textit{M}-patterns and arcs are successfully added.

\begin{algorithm}[t]
\linespread{1.0}
\caption{Main structure of WoMine.}
\label{alg:womine}
{\scriptsize

  \SetAlgoLined
  \DontPrintSemicolon
  
  \SetKwFunction{womine}{WoMine}
  
  \SetKwFunction{addM}{addFrequentMPattern}
  \SetKwFunction{addConn}{addFrequentConnection}
  \SetKwFunction{addArc}{addFrequentArcs}
  \SetKwFunction{searchLoopArcs}{searchStartloopEndloopArcs}
  
  \SetKwProg{algorithm}{Algorithm}{}{}
  \SetKwProg{function}{Function}{}{}
  
  \KwIn{A process model $W$, a set $T = \{T_1, T_2, \ldots, T_n\}$ of traces of $W$, and a threshold $thr$.}
  \KwOut{A set of maximum frequent patterns of $W$ w.r.t. $T$.}
  
  \algorithm{\womine{W, T, thr}}{
    $M \gets \{m \mid m \in W, m$ is an \textit{M}-pattern $\}$ \tcp{Def.~\ref{def:minimal-pattern}}\label{line:womine-initInitialize-start}
    $A^{<} \gets \{a \mid a \in W, a$ is a Candidate Arc $\}$ \tcp{Def.~\ref{def:candidate-arcs}}
    $\mathit{frequentArcs} \gets \{a \mid a \in A^{<}$, $a$ is frequent w.r.t. $T\}$\;
    $\mathit{frequentM} \gets \{m \mid m \in M$, isFrequentPattern($m$, $T$, $thr$) $\}$ \tcp{using Alg.~\ref{alg:isFreqPatt}}
    $\mathit{frequentPatterns} \gets \mathit{frequentM}$\;
    $currentPatterns \gets \mathit{frequentM}$\;\label{line:womine-initInitialize-end}
    \While{$currentPatterns \neq \emptyset$}{\label{line:start-iterative-process}
        $candidatePatterns \gets \emptyset$\;
        \ForAll{$p \in currentPatterns$}{
            $candidatePatterns \gets candidatePatterns$ $\cup$ \addArc{p}\;\label{line:add-arc}
            $complementaryM \gets \{m \mid m \in M$, $m \not\in p\}$\;\label{line:add-complementary-start}
            \ForAll{$m \in complementaryM$}{
                $candidatePatterns \gets candidatePatterns$ $\cup$ \addM{p, m}\;
            }\label{line:add-complementary-end}
        }
        $currentPatterns \gets \{p \mid p \in candidatePatterns,$ isFrequentPattern($p$, $T$, $thr$)$\}$ \tcp{using Alg.~\ref{alg:isFreqPatt}}\label{line:filter-infreq-patt}
        $\mathit{frequentPatterns} \gets \mathit{frequentPatterns} \cup currentPatterns$\;
    }
    Delete the redundant patterns of $\mathit{frequentPatterns}$ \tcp{Section~\ref{sec:post-simplification}}\label{line:simplifyPatterns}
    
    \Return{frequentPatterns}\;
  }{}

  \function{\addM{p, m}}{\label{line:womine-addFrequentEWPattern}
    $p' \leftarrow$ add $m$ to $p$\;
    \uIf{$p'$ is connected}{
        \Return{p'}\;
    }\Else{
        \Return{\addConn{p', p, m}}\;
    }\label{line:womine-addFrequentEWPattern-end}
  }
  
  \function{\addConn{p', p, m}}{\label{line:womine-addFrequentConn}
    $patterns \leftarrow \emptyset$\;
    \ForAll{$arc \in frequentArcs$}{
        \If{$( arc.source \in p$ \textup\&\textup\& $arc.destination \in m )$ $\mid\mid$
        $( arc.source \in m$ \textup\&\textup\& $arc.destination \in p )$}{\label{line:womine-iscompliant-1}
            $q \leftarrow$ add $arc$ to $p'$\;
            $patterns \leftarrow patterns \cup q$\;
        }
    }
    \KwRet $patterns$\;\label{line:womine-addFrequentConn-end}
  }
  
  \function{\addArc{p}}{\label{line:womine-addFrequentArc}
    $patterns \leftarrow \emptyset$\;
    \ForAll{$arc \in freqArcs$}{
        \If{$ arc \not\in p$ \textup\&\textup\& $arc.source \in p$ \textup\&\textup\& $arc.destination \in p$}{\label{line:womine-iscompliant-2}
            $q \leftarrow$ add $arc$ to $p$\;
            $patterns \leftarrow patterns \cup q$\;
        }
    }
    \KwRet $patterns$\;\label{line:womine-addFrequentArc-end}
  }

}
\end{algorithm}

The pseudocode in Alg.~\ref{alg:womine} shows the main structure of the search made by WoMine.
First, the frequent arcs of $A^{<}$ and the frequent \textit{M}-patterns are initialized using the algorithm described in Section~\ref{sec:measure-frequency} to measure the frequency ---\textit{M} is the set with all \textit{M}-patterns of the model.
These \textit{M}-patterns will be used to start the iterative stage, and to expand other patterns with them.
Also, the final set is initialized with them because they are valid frequent patterns (Alg.~\ref{alg:womine}:\ref{line:womine-initInitialize-start}-\ref{line:womine-initInitialize-end}).

Afterwards, the iterative part starts (Alg.~\ref{alg:womine}:\ref{line:start-iterative-process}).
In this stage, an expansion of each of the current patterns is done, followed by a filtering of the frequent patterns.
The expansion by adding frequent arcs of the $A^{<}$ set (Alg.~\ref{alg:womine}:\ref{line:add-arc}) is done with the function \texttt{addFrequentArcs} (Alg.~\ref{alg:womine}:\ref{line:womine-addFrequentArc}-\ref{line:womine-addFrequentArc-end}).
The other expansion, the addition of \textit{M}-patterns that are not in the current pattern (Alg.~\ref{alg:womine}:\ref{line:add-complementary-start}-\ref{line:add-complementary-end}), is done with the function \texttt{addFrequentMPattern} (Alg.~\ref{alg:womine}:\ref{line:womine-addFrequentEWPattern}-\ref{line:womine-addFrequentEWPattern-end}).
If the joint pattern is unconnected, it generates as many patterns as possible by adding arcs that connect the unconnected parts with the function \texttt{addFrequentConnection} (Alg.~\ref{alg:womine}:\ref{line:womine-addFrequentConn}-\ref{line:womine-addFrequentConn-end}).
Once the expansion is completed, the obtained patterns are filtered to delete the infrequent ones (Alg.~\ref{alg:womine}:\ref{line:filter-infreq-patt}).
Finally, once the iterative stage finishes, a simplification is made to delete the patterns which provide redundant information (Alg.~\ref{alg:womine}:\ref{line:simplifyPatterns}).
This simplification stage is explained in detail in Section~\ref{sec:post-simplification}.

WoMine is a robust algorithm, even for process models with low fitness, precision or generalization, as it extracts the patterns from the model, but measures the frequency with the log.
If a structure is supported by the log, but it does not appear in the model (low fitness), it will not be considered as a frequent pattern.
Anyway, this situation is irrelevant because, unless the model has a very low fitness, the unsupported structures will have low frequency.
Moreover, the patterns detected by WoMine are not affected by models with high generalization ---models that allow behaviour not recorded in the log---: the non-existent structures in the log have a frequency of 0 and, thus, will never be detected by WoMine.

    \section{Measuring the Frequency of a Pattern\label{sec:measure-frequency}}

\begin{algorithm}[t]
\linespread{1.0}
\caption{Check if a given pattern is executed more times than a threshold.}
\label{alg:isFreqPatt}
{\footnotesize
    
    \SetAlgoLined
    \DontPrintSemicolon
    
    \SetKwFunction{isFreqPat}{isFrequentPattern}
    \SetKwFunction{getFreqPattern}{getPatternFrequency}
    \SetKwFunction{isPatternExecuted}{isTraceCompliant}
    \SetKwFunction{simulateExec}{simulateExecutionInPattern}
    \SetKwFunction{restartAndTry}{restartAndTryAgain}

    \SetKwProg{algorithm}{Algorithm}{}{}
    \SetKwProg{function}{Function}{}{}
    
    \KwIn{A set $T = {T_{1}, ..., T_{N}}$ of traces, a pattern $pattern$ to measure its frequency w.r.t. $T$ and a threshold to establish the bound of frequency.}
    \KwOut{A Boolean value indicating if the pattern is frequent or not.}
  
    \algorithm{\isFreqPat{pattern, T, threshold}}{
        $simplePatterns \gets$ generate the simple patterns of $pattern$\;\label{line:get-combinations}
        $\mathit{frequencies} \gets \emptyset$\;
        \ForAll{$simplePattern \in simplePatterns$}{
            $\mathit{frequencies} \gets \mathit{frequencies}$ $\cup$ \getFreqPattern{simplePattern, T}\;
        }\label{line:end-get-combinations}
        $minFreq \gets 0$\;
        \If{frequencies.length $> 0$}{
        	$minFreq \gets $ minimum of $\mathit{frequencies}$\;
        }
        $realFreq \gets minFreq / T.length$\;
        \Return{$realFreq \geq threshold$}\;\label{line:return-is-infrequent}
    }

    \function{\getFreqPattern{pattern, T}}{
        $executed \gets 0$\;
        \ForAll{$trace \in T$}{\label{line:start-foreach-traces}
            \If{\isPatternExecuted{pattern, trace}}{
                $executed \gets executed + 1$\;
            }
        }\label{line:end-foreach-traces}
        \Return{executed}\;
    }

    \function{\isPatternExecuted{pattern, trace}}{\label{line:func-is-pattern-executed}
    
        \ForAll{$task \in trace$}{\label{line:foreach-tasks}
            Execute $task$ in the process model\;
            $sources \gets$ get the tasks that fired the execution of $task$\;\label{line:get-firing-tasks}
            \simulateExec{sources, task, pattern}\;\label{line:replicate-execution}
            
            \If{pattern {\upshape has been successfully executed}}{\label{line:has-been-pattern-executed}
                \Return{true}\;
            }
        }
        
        \Return{false}\;\label{line:pattern-not-executed}
    }
    
}
\end{algorithm}

In each step of the iterative process, WoMine reduces the search space by pruning the infrequent patterns (Alg.~\ref{alg:womine}:\ref{line:filter-infreq-patt}).
For this, an algorithm to check the frequency of a pattern is needed (Alg.~\ref{alg:isFreqPatt}).
Following Defs.~\ref{def:frequency-pattern-simple-pattern} and~\ref{def:frequent-pattern}, the algorithm generates the simple patterns of a pattern and checks the frequency of each one (Alg.~\ref{alg:isFreqPatt}:\ref{line:get-combinations}-\ref{line:end-get-combinations}).
After calculating the frequency of the simple patterns, the function checks if this is considered frequent w.r.t. the threshold and returns the corresponding value (Alg.~\ref{alg:isFreqPatt}:\ref{line:return-is-infrequent}).
The frequency of a simple pattern is measured in the function \texttt{getPatternFrequency} by parsing all the traces and checking how many of them are compliant with it (Alg.~\ref{alg:isFreqPatt}:\ref{line:start-foreach-traces}-\ref{line:end-foreach-traces}).
Finally, to check if a trace is compliant with a simple pattern, the function \texttt{isTraceCompliant} is executed: it goes over the tasks in the trace (Alg.~\ref{alg:isFreqPatt}:\ref{line:foreach-tasks}), simulating its execution in the model, and retrieving the tasks that have fired the current one (Alg.~\ref{alg:isFreqPatt}:\ref{line:get-firing-tasks}-\ref{line:replicate-execution}).
The simulation (\texttt{simulateExecutionInPattern}) consists in a replay of the trace, checking if the pattern is executed correctly.

\begin{figure}[t!]
    \centering
    \subfloat[Petri net of a process model with a pattern highlighted in black (the unnamed task is an invisible task, i.e., a task that is fired automatically to simulate the arc $\langle C \rightarrow H \rangle$).]{
        \includegraphics[width=0.25\textwidth]{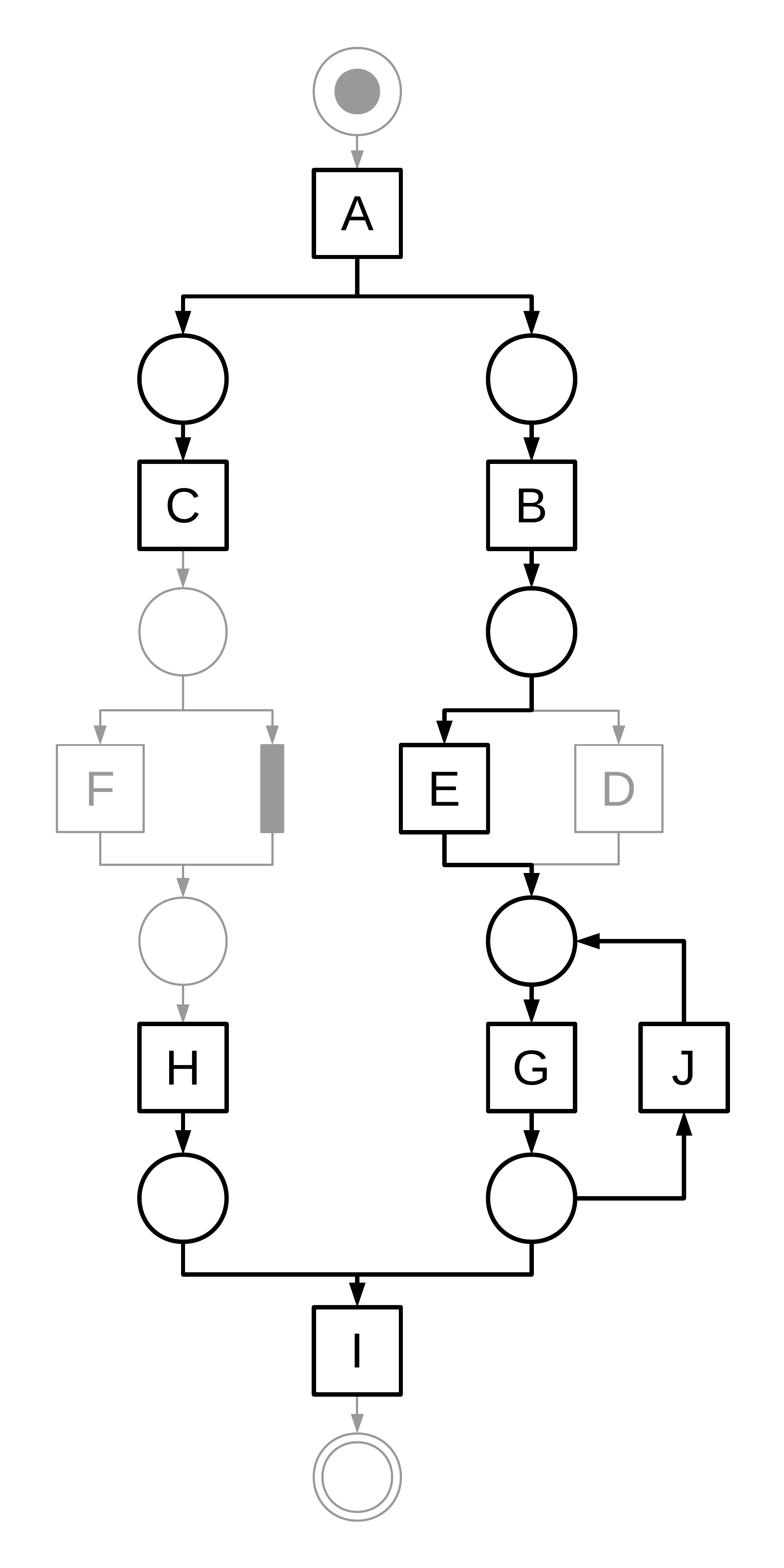}
        \label{fig:petri-net-for-is-pattern-executed}
    }
    \hfill
    \subfloat[Check of the execution of a trace for the pattern highlighted in Fig.~\ref{fig:petri-net-for-is-pattern-executed}: '\#' is the step of the algorithm; \textit{'executed task'} is the task currently executed; \textit{'executed arcs'} is the set with the arcs belonging to the pattern which execution was correctly saved; \textit{'Last executed tasks'} is the set of tasks which have not fired an entire set of their outputs.]{
        \small
        \begin{tabular}[b]{|r|c|l|l|}
            \hline
            \multicolumn{4}{|c|}{\textbf{Trace:} \texttt{A B E G J G C F H I}}                                                                                                                                                                                                                                                                                                                                                                                                                                                                                                                                     \\ \hline
            \multicolumn{4}{|c|}{\textbf{Initial tasks}: \texttt{[A, H]}            \textbf{End tasks}: \texttt{[C, I]}}                                                                                                                                                                                                                                                                                                                                                                                                                                                                                           \\ \hline
            \multicolumn{1}{|c|}{\textbf{\#}} & \textbf{executed task} & \multicolumn{1}{c|}{\textbf{executed arcs}}                                                                                                                                                                                                                                                                          & \multicolumn{1}{c|}{\textbf{\begin{tabular}[c]{@{}c@{}}last executed\\ tasks\end{tabular}}} \\ \hline
            0                                 & -                                                                & $\emptyset$                                                                                                                                                                                                                                                                                                          & $\emptyset$                                                                                                                                   \\ \hline
            1                                 & A                                                                & $\emptyset$                                                                                                                                                                                                                                                                                                          & A                                                                                                                                               \\ \hline
            2                                 & B                                                                & $\langle A \rightarrow B \rangle$                                                                                                                                                                                                                                                                                    & A, B                   
            \\ \hline                                                           
            3                                 & E                                                                & $\langle A \rightarrow B \rangle, \langle B \rightarrow E \rangle$                                                                                                                                                                                                                                                   & A, E                                                                                                                                                   \\ \hline
            4                                 & G                                                                & $\langle A \rightarrow B \rangle, \langle B \rightarrow E \rangle, \langle E \rightarrow G \rangle$                                                                                                                                                                                                                  & A, G                                                                                                                                              \\ \hline
            5                                 & J                                                                & $\langle A \rightarrow B \rangle, \langle B \rightarrow E \rangle, \langle E \rightarrow G \rangle, \langle G \rightarrow J \rangle$                                                                                                                                                                                 & A, J                                                                                                                                           \\ \hline
            6                                 & G                                                                & \begin{tabular}[c]{@{}l@{}}$\langle A \rightarrow B \rangle, \langle B \rightarrow E \rangle, \langle E \rightarrow G \rangle, \langle G \rightarrow J \rangle,$\\ $\langle J \rightarrow G \rangle$\end{tabular}                                                                                                    & A, G                                                                                                                                                  \\ \hline
            7                                 & C                                                                & \begin{tabular}[c]{@{}l@{}}$\langle A \rightarrow B \rangle, \langle B \rightarrow E \rangle, \langle E \rightarrow G \rangle, \langle G \rightarrow J \rangle,$\\ $\langle J \rightarrow G \rangle, \langle A \rightarrow C \rangle$\end{tabular}                                                                   & G, C                                                                                                                                              \\ \hline
            8                                 & F                                                                & \begin{tabular}[c]{@{}l@{}}$\langle A \rightarrow B \rangle, \langle B \rightarrow E \rangle, \langle E \rightarrow G \rangle, \langle G \rightarrow J \rangle,$\\ $\langle J \rightarrow G \rangle, \langle A \rightarrow C \rangle$\end{tabular}                                                                   & G, C                                                                                                                                                    \\ \hline
            9                                 & H                                                                & \begin{tabular}[c]{@{}l@{}}$\langle A \rightarrow B \rangle, \langle B \rightarrow E \rangle, \langle E \rightarrow G \rangle, \langle G \rightarrow J \rangle,$\\ $\langle J \rightarrow G \rangle, \langle A \rightarrow C \rangle$\end{tabular}                                                                   & G, C, H                                                                                                                                                     \\ \hline
            10                                & I                                                                & \begin{tabular}[c]{@{}l@{}}$\langle A \rightarrow B \rangle, \langle B \rightarrow E \rangle, \langle E \rightarrow G \rangle, \langle G \rightarrow J \rangle,$\\ $\langle J \rightarrow G \rangle, \langle A \rightarrow C \rangle, \langle G \rightarrow I \rangle, \langle H \rightarrow I \rangle$\end{tabular} & C, I                                                                                                                                                       \\ \hline
        \end{tabular}
        \label{fig:example-trace}
    }
    \caption{An example that shows how the algorithm checks if a trace is compliant with a pattern of the process model.}
    \label{fig:petri-net-and-trace}
\end{figure}

With the current task ---the fired one--- and the tasks that have fired it ---the firing tasks, retrieved by the simulation---, the executed tasks and arcs are saved, in order to analyse and to detect if the execution of the pattern is being disrupted before it is completed (Alg.~\ref{alg:isFreqPatt}:\ref{line:replicate-execution}).
Fig.~\ref{fig:petri-net-and-trace} shows an example of this process.
The algorithm starts (\#0) with the sets of the \textit{executed arcs} and \textit{last executed tasks} empty.
The first step (\#1) executes \texttt{A}.
There are no firing tasks because \texttt{A} is the initial task of the process model.
As \texttt{A} is also one of the initial tasks of the pattern, it is saved correctly in the \textit{last executed tasks} set.

The following task (\#2) in the trace is \texttt{B}.
As there is only one firing task (\texttt{A}), a single arc is executed ($\langle A \rightarrow B \rangle$).
The arc is added to the \textit{executed arcs} set, and the task \texttt{B} to the \textit{last executed tasks} set.
The \texttt{A} task is not deleted because the set of outputs is formed by \texttt{\{B, C\}}, and \texttt{C} is still pending.

The next four steps, tasks \texttt{E} (\#3), \texttt{G} (\#4), \texttt{J} (\#5) and \texttt{G} (\#6), will have the same behaviour.
They have only one firing task, i.e., one executed arc.
The arcs are in the pattern and their source tasks are in the \textit{last executed tasks} set, because they were executed before them.
Hence, after adding and removing these tasks from the last executed ones, \texttt{G} is the remaining one.
After this process, the following task is \texttt{C} (\#7).
Its execution has the same behaviour as the execution of \texttt{B}, but with the deletion of \texttt{A} from the \textit{last executed tasks}, because the set of outputs \texttt{\{B, C\}} has been fired.

At step \#8, only the source task of the arc $\langle C \rightarrow F \rangle$ is in the pattern.
In this case, the source of the arc is one of the end tasks and, thus, the pattern finishes its execution in that branch.
The execution is done with no action in the sets \textit{executed arcs} and \textit{last executed tasks}.

In the next iteration (\#9) only the target task of the executed arc $\langle F \rightarrow H \rangle$ is in the pattern.
As the target is one of the initial tasks, the pattern starts to be executed in that branch.
Thus, in a similar way to \texttt{A}, \texttt{H} is added to the \textit{last executed tasks} set.

Finally (\#10), \texttt{I} has two firing tasks and, thus, two arcs are executed.
In both cases, the source task of the arcs ---\texttt{G} and \texttt{H}--- is in the \textit{last executed tasks} set, and the arc is in the pattern.
Thus, a simple addition of \texttt{I} to the \textit{last executed tasks} set is done when the last of its branches is executed.

At the end of each step, the algorithm checks if the pattern has been correctly executed (Alg.~\ref{alg:isFreqPatt}:\ref{line:has-been-pattern-executed}), i.e., all its arcs have been correctly executed and the \textit{last executed tasks} set corresponds with the end tasks of the pattern.
Unlike the other steps, this testing has a positive result when \texttt{I} is executed.
Thus, the trace is compliant with the pattern.

The process of saving the executed arcs and tasks has to be restarted when the executed arc is disrupting the execution of the pattern.
For instance, in step \#3, if the arc $\langle B \rightarrow D \rangle$ was executed instead of $\langle B \rightarrow E \rangle$, this would cause this saving process to go back by removing the arcs and tasks of the failed path and to continue with the trace in order to check if the execution of the pattern is resumed later.
When an arc outside the pattern ---for example of a concurrent branch--- is executed, the analysis is not disrupted because the execution does not correspond to the pattern.
This analysis is able to recognize the correct execution of a pattern in 1-safe Petri nets\footnote{A Petri net is 1-safe when the value of the places can be binary, i.e., there can be only one mark in a place at the same time.}.

    \section{Simplifying the Result Set of Patterns\label{sec:post-simplification}}

\begin{figure}[t]
    \centering
    \subfloat[Simple pattern formed by a sequence.]{
        \includegraphics[width=0.45\textwidth]{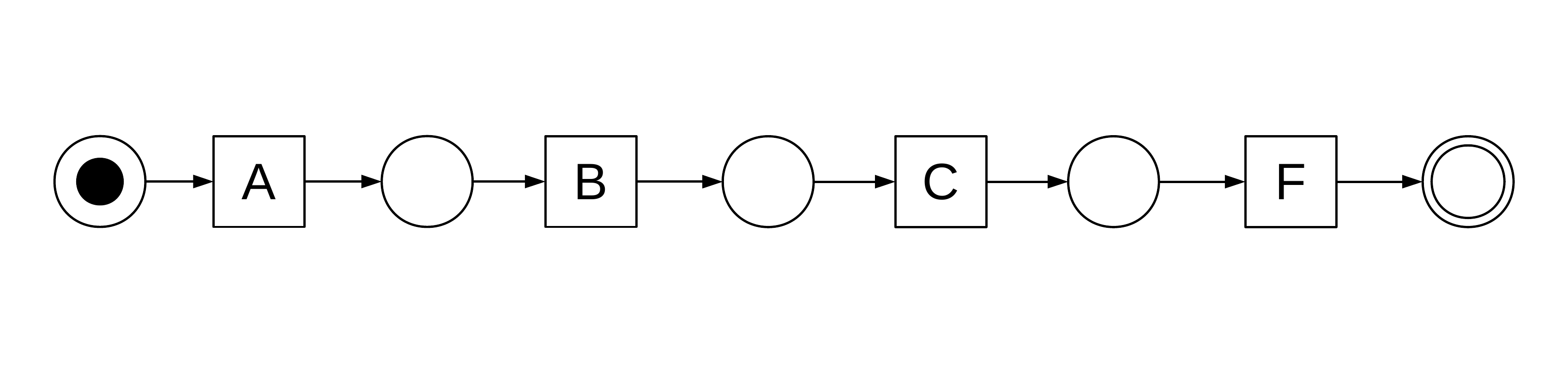}
        \label{fig:example-why-simplification-no-loop}
    }
    \hfill
    \subfloat[Simple pattern with a 2-length loop.]{
        \includegraphics[width=0.45\textwidth]{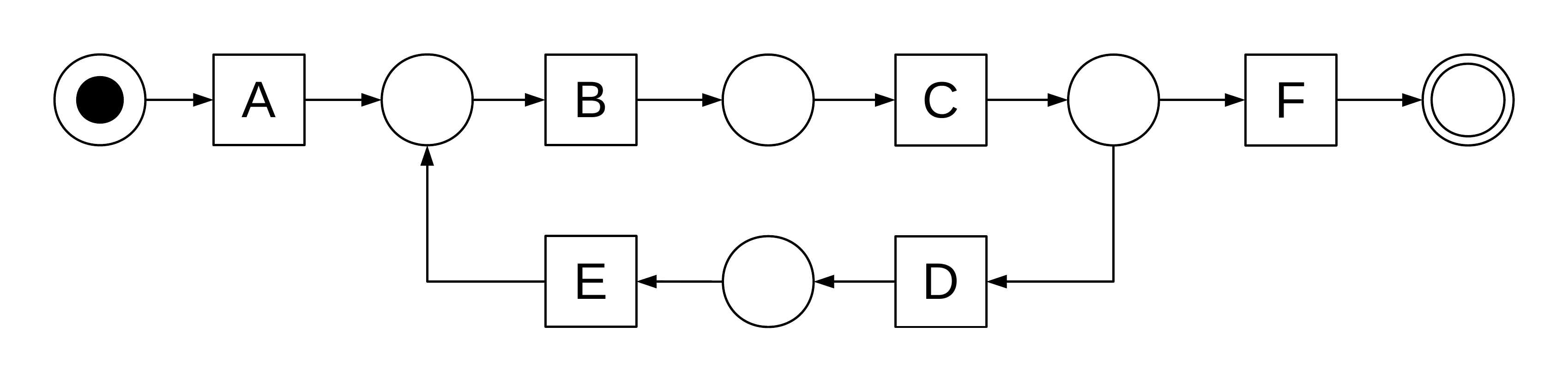}
        \label{fig:example-why-simplification-with-loop}
    }
    \caption{An example where the naive simplification fails. 
    With this naive technique, assuming a scenario where~\ref{fig:example-why-simplification-no-loop} and~\ref{fig:example-why-simplification-with-loop} were frequent patterns,~\ref{fig:example-why-simplification-no-loop} would be removed from the frequent results ---because~\ref{fig:example-why-simplification-with-loop} would be obtained, among others, by an expansion of~\ref{fig:example-why-simplification-no-loop}.
    Thus, the apparition of~\ref{fig:example-why-simplification-with-loop} as a frequent pattern would mean that both behaviours, the sequence (\textit{A-B-C-F}) and the pattern with the loop (\textit{A-B-C-D-E-B-C-F}), are frequent.
Therefore, with the naive technique, it is impossible to indicate that the pattern with the loop appears frequently in the traces, but the sequence does not.
Because the apparition of pattern~\ref{fig:example-why-simplification-with-loop} in the results indicates both behaviours as frequent.
    }
    \label{fig:example-why-simplification}
\end{figure}

The result set of the a priori search has a high redundancy.
This is because there are patterns in the $k$-th iteration which are expanded and thus are subpatterns of those in the $k+1$-th iteration.
A naive approach to reduce the redundancy generated by the expansion might be to remove the patterns from iteration $k$-th that are expanded in iteration $k+1$-th.
But, with the existence of loops, this naive approach might fail (see Fig.~\ref{fig:example-why-simplification} for an example).

In Womine, the simplification process deletes the patterns contained into others, but only when the behaviour of the contained pattern is also included in the other pattern.
For this, each pattern is compared with its previous patterns in the expansion.
If all tasks and arcs of a previous pattern are contained into the current one, and there is no new loops in the current pattern, the previous one is deleted.
For this, an algorithm to detect loop arcs (Section~\ref{subsec:search-loop-arcs}) is executed for the current pattern.
For instance, if both~\ref{fig:example-why-simplification-with-loop} and~\ref{fig:example-why-simplification-no-loop} were in the results set, WoMine would detect that~\ref{fig:example-why-simplification-no-loop} is inside~\ref{fig:example-why-simplification-with-loop}, but, as the difference between them begins in the start of a loop and finishes in the end, the pattern is not deleted.
The next section explains the approach designed to identify the arcs starting and closing a loop.

\subsection{Identification of the start and end arcs of a loop\label{subsec:search-loop-arcs}}

\begin{quotation}
    \begin{defn}[Startloop arc]
    
    Given a process model composed by a set of tasks $T$, and a set of arcs $E$, a \textit{Startloop} arc $e \in E$ is an arc that starts a loop, i.e., $e$ starts a path that will result, inevitably, in a return to a previous task in the process.
    
    \end{defn}
\end{quotation}
\begin{quotation}
    \begin{defn}[Endloop arc]
    
    Given a process model composed by a set of tasks $T$, and a set of arcs $E$, an \textit{Endloop} arc $e \in E$ is an arc that ends a loop, i.e., the transition through $e$ implies the immediate closing of a loop and a return to a previous task in the process.
    
    \end{defn}
\end{quotation}

An example of a Startloop arc is $\langle C \rightarrow D \rangle$ in Fig.~\ref{fig:example-why-simplification-with-loop}, while $\langle E \rightarrow B \rangle$ is an Endloop arc.
Alg.~\ref{alg:searchLoopArcs} identifies the Startloop and Endloop arcs of a pattern.
The approach is an iterative process with two different phases in each iteration.
First, it searches the Startloop arcs (Alg.~\ref{alg:searchLoopArcs}:\ref{line:searchloops-firstphase-1}) and then, based on these arcs, it looks for the arcs that close the loops ---Endloop arcs--- (Alg.~\ref{alg:searchLoopArcs}:\ref{line:searchloops-secondphase-1}-\ref{line:searchloops-secondphase-2}).

\subsubsection{Startloop search}

The search starts with the initial tasks of the pattern and goes forward until it reaches a task with more than one output.
When this happens, an analysis trying to close a loop ---reach the current task again--- is thrown for each output arc.
The analysis goes forward through non-Startloop arcs and finishes when it reaches the current task or the end of the pattern.
If the analysis reaches the starting task, the arc is marked as Startloop.

Table~\ref{tab:simulation-first-phase} shows an example of this search for the pattern of Fig.~\ref{fig:example-why-simplification-with-loop}.
It starts at \texttt{A} (Alg.~\ref{alg:searchLoopArcs}:\ref{line:searchloops-initialNode}), and stops at \texttt{C} in step \#3, because it has more than one output (Alg.~\ref{alg:searchLoopArcs}:\ref{line:searchloops-goforward}).
With its output arcs, $\langle C \rightarrow F \rangle$ and $\langle C \rightarrow D \rangle$, a subanalysis to detect if any of them is the beginning of a loop starts (Alg.~\ref{alg:searchLoopArcs}:\ref{line:searchloops-throwAnalysis}).
This analysis performs a depth-first search going forward through the arcs that are not detected as Startloop, trying to find the source of the arc that started the analysis, i.e., trying to close the loop.
With  $\langle C \rightarrow F \rangle$ it reaches the end of the model and stops the search (\#4).
But with $\langle C \rightarrow D \rangle$ it closes the loop reaching \texttt{C}, after going through \texttt{D} (\#5), \texttt{E} (\#6) and \texttt{B} (\#7).
In this case, the arc under analysis, $\langle C \rightarrow D \rangle$, is marked as Startloop (\#8).

\begin{table}[H]
\centering
{\footnotesize

    \begin{tabular}{|c|c|l|c|c|c|l|}
        \hline
        \multirow{2}{*}{\textbf{\#}} & \multirow{2}{*}{\textbf{\begin{tabular}[c]{@{}c@{}}task under\\ analysis\end{tabular}}} & \multicolumn{1}{c|}{\multirow{2}{*}{\textbf{outputs}}} & \multicolumn{3}{c|}{\textbf{subanalysis}}                                    & \multicolumn{1}{c|}{\multirow{2}{*}{\textbf{action}}}       \\ \cline{4-6}
                                     &                                                                                         & \multicolumn{1}{c|}{}                                  & \textbf{possible Startloop arc}   & \textbf{current task} & \textbf{outputs} & \multicolumn{1}{c|}{}                                       \\ \hline
        1                            & A                                                                                       & B                                                      & -                                 & -                     & -                & only one output, continue                                   \\ \hline
        2                            & B                                                                                       & C                                                      & -                                 & -                     & -                & only one output, continue                                   \\ \hline
        3                            & C                                                                                       & F, D                                                   & -                                 & -                     & -                & two outputs, start subanalysis searching for C              \\ \cline{1-1} \cline{4-7} 
        4                            &                                                                                         &                                                        & $\langle C \rightarrow F \rangle$ & F                     & $\emptyset$      & reached end of model, C not found                        \\ \cline{1-1} \cline{4-7} 
        5                            &                                                                                         &                                                        & $\langle C \rightarrow D \rangle$ & D                     & E                & C not found, continue with outputs                          \\ \cline{1-1} \cline{5-7} 
        6                            &                                                                                         &                                                        &                                   & E                     & B                & C not found, continue with outputs                          \\ \cline{1-1} \cline{5-7} 
        7                            &                                                                                         &                                                        &                                   & B                     & C                & C found, set $\langle C \rightarrow D \rangle$ as Startloop \\ \hline
        8                            & -                                                                                       & \multicolumn{1}{c|}{-}                                 & -                                 & -                     & -                & No more tasks, end of first phase                           \\ \hline
    \end{tabular}
    \caption{Startloop search for the pattern of Fig.~\ref{fig:example-why-simplification-with-loop}: \textit{'task under analysis'} is the task which output arcs are currently examined; \textit{'outputs'} are the outputs of the task under analysis, i.e., the targets of the arcs; \textit{'possible Startloop arc'} is the arc being analysed; \textit{'current task'} is the current task in the search of the source task of the arc; \textit{'outputs'} is the set of outputs of the 'current task', i.e., the next in the analysis; \textit{'action'} is a description of the action at the end of each iteration.}
    \label{tab:simulation-first-phase}
}
\end{table}

\begin{algorithm}[b!]
\linespread{1.0}
\caption{Identify the Startloop and Endloop arcs of a pattern or process model.}
\label{alg:searchLoopArcs}
{\scriptsize

  \SetAlgoLined
  \DontPrintSemicolon
  
  \SetKwFunction{searchAlg}{searchLoopArcs}
  \SetKwFunction{searchStart}{searchStartloopArcs}
  
  \SetKwFunction{arriveStart}{arriveStartWithoutLoop}
  \SetKwFunction{continueOutputs}{continueWithOutputs}
  \SetKwFunction{analizeSplit}{analizeSplit}
  \SetKwFunction{extendExpl}{extendExploration}
  
  \SetKwProg{algorithm}{Algorithm}{}{}
  \SetKwProg{function}{Function}{}{}
  
  \KwIn{A pattern $p$}
  \KwOut{The pattern $p$ with the Startloop and Endloop arcs identified}
  
  \algorithm{\searchAlg{p}}{
    $startloopArcs \gets \emptyset$\;
    \Do{$startloopArcs \neq \emptyset$}{
        $startloopArcs \gets$ \searchStart{p, startloopArcs}\;\label{line:searchloops-firstphase-1}
        \ForAll{$arc \in startloopArcs$}{\label{line:searchloops-secondphase-1}
            $startTask \gets arc.target$\;
            \uIf{\arriveStart{startTask, startloopArcs}}{\label{line:searchloops-reachStartFromDestination}
                set $arc$ as Endloop\;
            }\Else{
                \continueOutputs{startTask, startloopArcs}\;\label{line:searchloops-continueOutputsOfDestination}
            }
        }\label{line:searchloops-secondphase-2}
    }
    \Return{p}
  }{}
  
  \function{\searchStart{p, previousStartloops}}{
    $initialTasks \gets \emptyset$\;
    \uIf{$\mathit{previousStartloops} = \emptyset$}{
        $initialTasks \gets$ start tasks of $p$\;\label{line:searchloops-initialNode}
    }\Else{
        $initialTasks \gets$ targets of arcs in $\mathit{previousStartloops}$\;\label{line:searchloops-initialNodes}
    }
    \ForAll{$task \in initialTasks$}{
        Go forward through the outputs while there is only one\;\label{line:searchloops-goforward}
        \ForAll{$output \in$ {\upshape sorted} $task.outputs$}{
            \analizeSplit{task, output}\;\label{line:searchloops-throwAnalysis}
        }
    }
    \KwRet the set of new Startloops\;
  }
  
  \function{\continueOutputs{previousTask, startloopArcs}}{
    \ForAll{$output \in previousTask.outputs$}{
        \uIf{\arriveStart{output, startloopArcs}}{\label{line:searchloops-checkReachStart}
            set $\langle previousTask \rightarrow output \rangle$ as Endloop\;
        }\Else{
            \If{$output$ {\upshape has not been explored}}{
                \continueOutputs{output, startloopArcs}\;\label{line:searchloops-continueOutputs}
            }
        }
    }
  }
}
\end{algorithm}

\subsubsection{Endloop search}

This search starts with the target task of each Startloop detected in the same iteration.
For each task, it goes forward through its output arcs by analysing their target tasks.
The analysis continues until a task that can reach, going backwards, the start of the pattern is found.
When the start is reached, the current arc is marked as Endloop.
In this backwards search, the algorithm cannot go through a Startloop arc detected in the same iteration.

Table~\ref{tab:simulation-second-phase} shows an example of this search for the pattern of Fig.~\ref{fig:example-why-simplification-with-loop}.
As there is only one Startloop detected in this iteration ($\langle C \rightarrow D \rangle$), the search begins with it.
The search tries to go backwards from \texttt{D} (the target of the Startloop arc) in step \#1 (Alg.~\ref{alg:searchLoopArcs}:\ref{line:searchloops-reachStartFromDestination}), but as the unique input arc is one of the detected Startloop arcs, it stops and goes forward to \texttt{E} in step \#2 (Alg.~\ref{alg:searchLoopArcs}:\ref{line:searchloops-continueOutputsOfDestination}).
From \texttt{E} the same happens: it goes backwards to \texttt{D} (\#3) and stops again.
Finally, it searches from \texttt{B} (\#4), where the algorithm is able to reach the initial task going backwards through the $\langle A \rightarrow B \rangle$ arc (\#7). Therefore, the current arc, $\langle E \rightarrow B \rangle$, is marked as Endloop.

\begin{table}[H]
\centering
{\footnotesize

    \begin{tabular}{|l|l|l|l|l|l|l|}
        \hline
        \multicolumn{1}{|c|}{\textbf{\#}} & \multicolumn{1}{c|}{\textbf{Startloop}}              & \multicolumn{1}{c|}{\textbf{\begin{tabular}[c]{@{}c@{}}possible\\ Endloop\end{tabular}}} & \multicolumn{1}{c|}{\textbf{\begin{tabular}[c]{@{}c@{}}task to\\ reach the\\ start from\end{tabular}}} & \multicolumn{1}{c|}{\textbf{\begin{tabular}[c]{@{}c@{}}pending\\ to analyse\end{tabular}}} & \multicolumn{1}{c|}{\textbf{\begin{tabular}[c]{@{}c@{}}input \\ under\\ analysis\end{tabular}}} & \multicolumn{1}{c|}{\textbf{action}}                                                            \\ \hline
        1                                 & \multirow{7}{*}{$ \langle C \rightarrow D \rangle $} & $ \langle C \rightarrow D \rangle $                                                      & D                                                                                                      & C                                                                                          & C                                                                                               & $ \langle C \rightarrow D \rangle \in$ detected Startloop arcs, cannot go back through this path \\ \cline{1-1} \cline{3-7} 
        2                                 &                                                      & $ \langle D \rightarrow E \rangle $                                                      & E                                                                                                      & D                                                                                          & D                                                                                               & keep going back                                                                                 \\ \cline{1-1} \cline{4-7} 
        3                                 &                                                      &                                                                                          & D                                                                                                      & C                                                                                          & C                                                                                               & $ \langle C \rightarrow D \rangle \in$ detected Startloop arcs, cannot go back through this path \\ \cline{1-1} \cline{3-7} 
        4                                 &                                                      & $ \langle E \rightarrow B \rangle $                                                      & B                                                                                                      & A, E                                                                                       & E                                                                                               & keep going back                                                                                 \\ \cline{1-1} \cline{4-7} 
        5                                 &                                                      &                                                                                          & E                                                                                                      & A, D                                                                                       & D                                                                                               & keep going back                                                                                 \\ \cline{1-1} \cline{4-7} 
        6                                 &                                                      &                                                                                          & D                                                                                                      & A, C                                                                                       & C                                                                                               & $ \langle C \rightarrow D \rangle \in$ detected Startloop arcs, cannot go back through this path \\ \cline{1-1} \cline{4-7} 
        7                                 &                                                      &                                                                                          & A                                                                                           & -                                                                                          & -                                                                                               & Start of model reached, $ \langle E \rightarrow B \rangle $ marked as Endloop arc               \\ \hline
    \end{tabular}
    \caption{Endloop search for the pattern of Fig.~\ref{fig:example-why-simplification-with-loop}: \textit{'startloop'} is the Startloop, of the detected ones in this iteration, to start the search; \textit{'possible Endloop'} is the arc considered as Endloop if the start of the pattern is reached from its target task; \textit{'task to reach the start from'} is the task from which the search is trying to reach the start in that step; \textit{'pending to analyse'} is the set of input tasks that have not been analysed in the search for the start of the pattern; \textit{'input under analysis'} is the task currently being analysed in the search for the start of the pattern; \textit{'action'} is a description of the action at the end of each iteration.}
    \label{tab:simulation-second-phase}
}
\end{table}

The iterative nature of the algorithm allows it to find loops inside other loops, and to detect also multiple Startloop or Endloop arcs for the same loop, i.e., loops with more than one input or more than one output.

    \section{Experimentation\label{sec:experiments}}

The validation of the presented approach has been done with different types of event logs.
Subsection~\ref{subsec:self-created-logs} presents the results of the comparison between WoMine and the state of the art techniques for 5 process models.
Subsection~\ref{subsec:de-medeiros-logs} discusses, for 20 logs from~\cite{phdGenetic}, the extracted frequent patterns and their evolution as the threshold varies.
Finally, in Subsection~\ref{subsec:bpic-logs}, we prove the performance of WoMine over complex real logs and compare the impact of the model quality in the extraction of patterns using several Business Process Intelligence Challenge's logs.

These experiments have been executed in a laptop (Lenovo G500) with an Intel i7-3612QM (2.1 GHz) processor and 8GB of RAM (1600 MHz)\footnote{The algorithm can be tested and downloaded from \url{http://tec.citius.usc.es/processmining/womine/}}.

\subsection{Comparison between WoMine and the state of the art approaches\label{subsec:self-created-logs}}

In this comparison, 5 process models with the most common control structures have been used.
These models present scenarios where WoMine is able to retrieve frequent patterns, while any of the other techniques fails.
For each model, two Petri nets will be presented: one with a highlighted frequent pattern extracted by WoMine, and another one with the frequent structure extracted by heat maps.
The structure obtained by heat maps is retrieved establishing a threshold, and highlighting all the arcs with a frequency over it.
The chosen threshold is the one that allows to get the structure closest to the frequent pattern extracted by WoMine.
Table~\ref{tab:comparison-experimentation} shows the results of these techniques for the 5 process models.

\begin{table}[H]
    \centering
    {\small
        \begin{tabular}{rl|c|c|c|c|c|}
            \cline{3-7}
                                                  &  & \multicolumn{5}{c|}{\textbf{Examples}}                                                                                                                                                                                                                                          \\ \cline{3-7} 
                                                  &  & \textbf{\#1 (Fig.~\ref{fig:experimentation-1})} & \textbf{\#2 (Fig.~\ref{fig:experimentation-2})} & \textbf{\#3 (Fig.~\ref{fig:experimentation-3})} & \textbf{\#4 (Fig.~\ref{fig:experimentation-5})} & \textbf{\#5 (Fig.~\ref{fig:experimentation-4})} \\ \hline
            
            \multicolumn{1}{|r|}{\textbf{WoMine}}         &  & \textbf{+}                                          & \textbf{+}                                          & \textbf{+}                              & \textbf{+}                              & \textbf{+}                              \\ \hline
            \multicolumn{1}{|r|}{Heat Maps}               &  & $\pm$                                               & -                                                   & -                                       & +                                       & -                                       \\ \hline
            \multicolumn{1}{|r|}{$w$-find}                &  & +                                                   & $\pm$                                               & -                                       & -                                       & -                                       \\ \hline
            \multicolumn{1}{|r|}{Local Process Mining}    &  & +                                                   & $\pm$                                               & $\pm$                                   & -                                       & $\pm$                                   \\ \hline
            \multicolumn{1}{|r|}{Episode Mining}          &  & +                                                   & $\pm$                                               & -                                       & -                                       & -                                       \\ \hline
            \multicolumn{1}{|r|}{SPM (PrefixSpan)}        &  & +                                                   & -                                                   & $\pm$                                   & -                                       & -                                       \\ \hline
            \multicolumn{1}{|r|}{Tree Mining}             &  & +                                                   & -                                                   & $\pm$                                   & -                                       & -                                       \\ \hline
        
        \end{tabular}
        \caption{Comparison between WoMine and other state of the art techniques for 5 process models: \textit{'+'} stands for a correct frequent pattern extraction; \textit{'-'} stands for a non extraction of the frequent pattern, and \textit{'$\pm$'} stands for an incorrect extraction of the frequent pattern (similar but wrong structure or wrong frequency).}
        \label{tab:comparison-experimentation}
    }
    
\end{table}

\begin{figure}[H]
    \centering
    \subfloat[WoMine pattern (highlighted) for a threshold of 40\%.]{
        \includegraphics[width=0.42\textwidth]{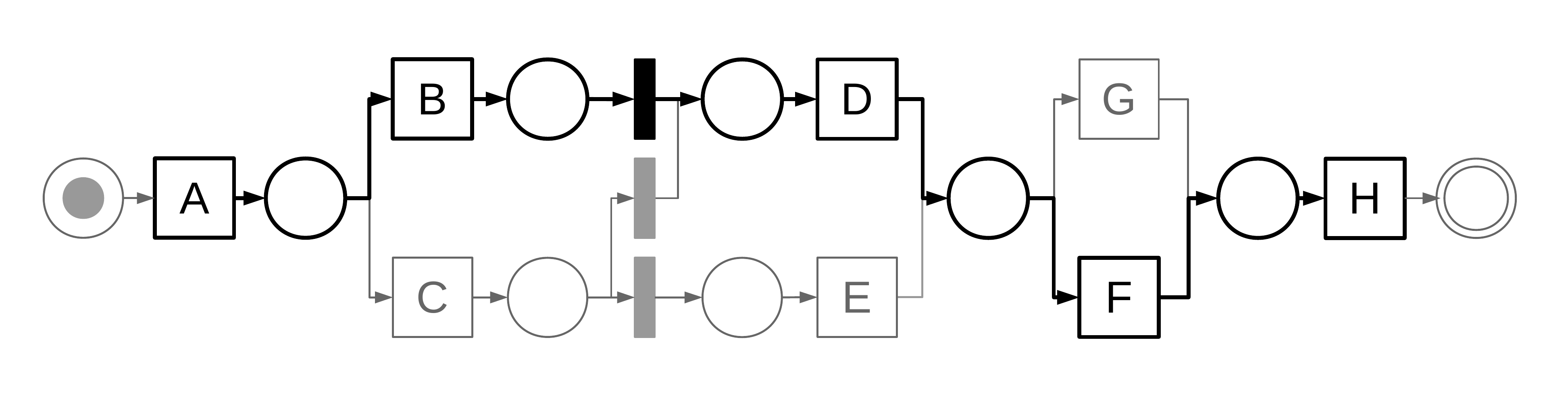}
        \label{fig:experimentation-custom-1}
    }
    \hfill
    \subfloat[Heat maps pattern (highlighted) for a threshold of 40.
    The pattern includes infrequent paths, for instance \texttt{A-C-D-H-F}.
    These arcs are frequent individually, but not the sequence itself.
    This happens because the executions of $\langle A \rightarrow C \rangle$ are distributed in traces through \texttt{D-G} and \texttt{E}.]{
        \includegraphics[width=0.42\textwidth]{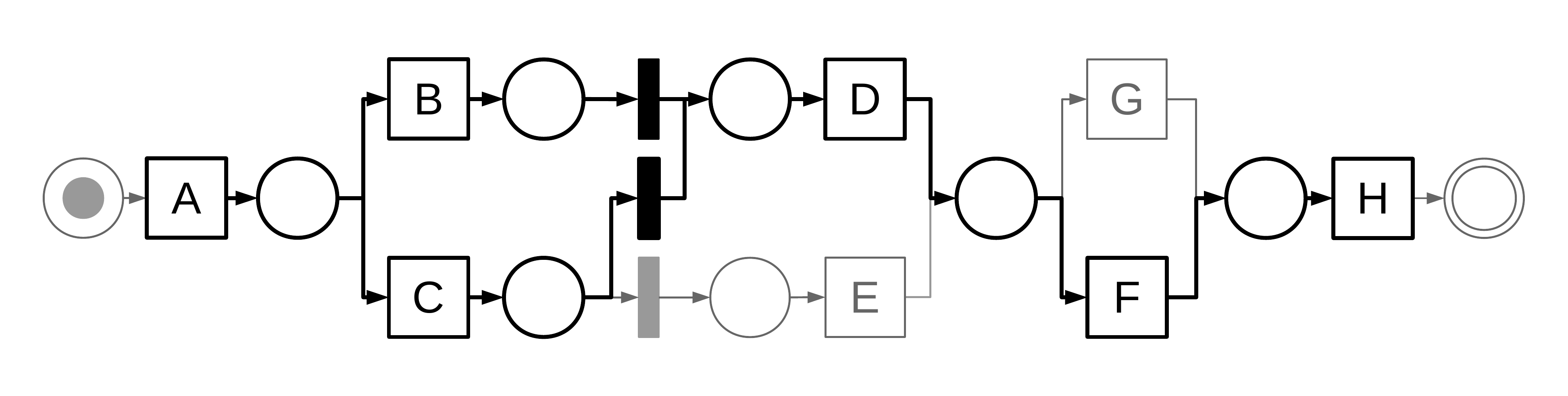}
        \label{fig:experimentation-custom-1-arcs}
    }
    \caption{Results of WoMine and heat maps for a process model with several selections.}
    \label{fig:experimentation-1}
\end{figure}

The first process model (Fig.~\ref{fig:experimentation-1}) has several selections.
WoMine finds a pattern appearing in the 40\% of the traces (Fig.~\ref{fig:experimentation-custom-1}).
On the contrary, the heat maps discovers, as frequent, paths that are not frequent.
The other approaches ---$w$-find, local process mining, episode mining, sequential pattern mining (PrefixSpan), and the tree mining approach--- detect the same pattern as WoMine.

\begin{figure}[H]
    \centering
    \subfloat[WoMine pattern (highlighted) for a threshold of 70\%. 
The traces compliant to this pattern are those where the loop is not executed.]{
        \includegraphics[width=0.42\textwidth]{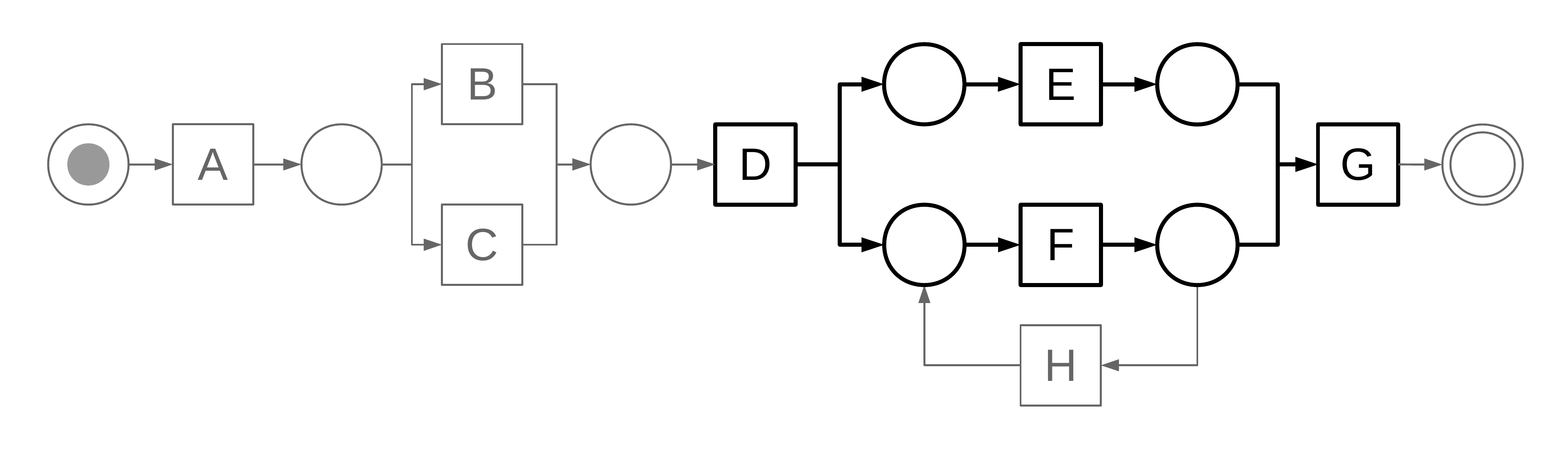}
        \label{fig:experimentation-custom-2}
    }
    \hfill
    \subfloat[Heat maps pattern (highlighted) for a threshold of 70.
    The parallel structure with the loop is not frequent.
    The loop is executed several times in the same trace, increasing its absolute frequency, but not the frequency of the whole pattern.]{
        \includegraphics[width=0.42\textwidth]{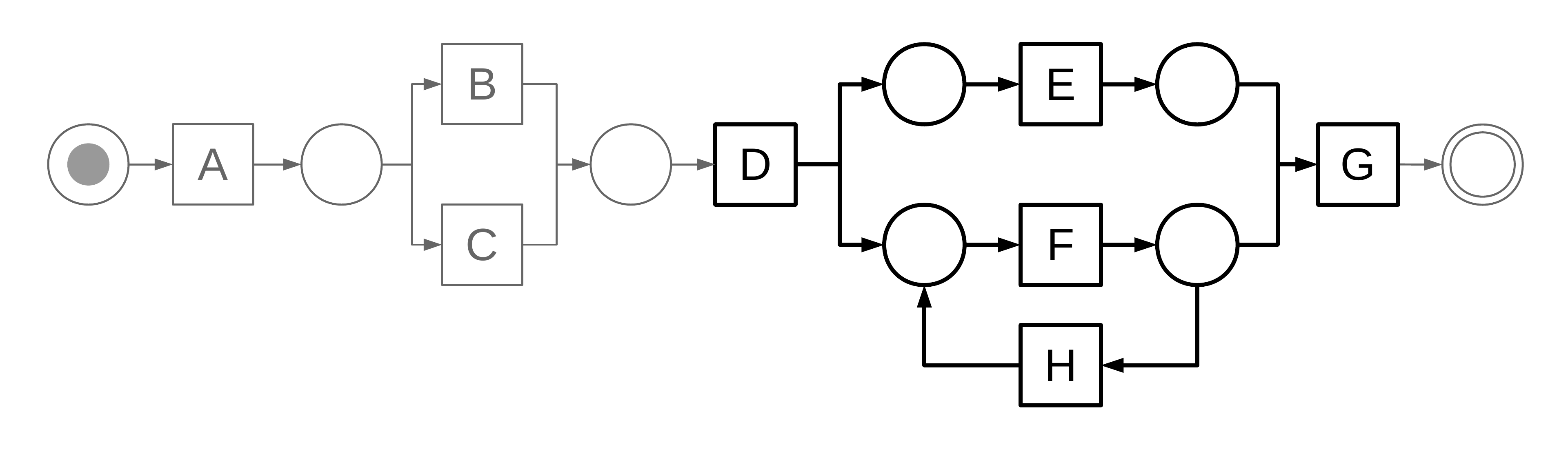}
        \label{fig:experimentation-custom-2-arcs}
    }
    \caption{Results of WoMine and heat maps for a process model with a 1-length loop inside a branch with a parallel structure.}
    \label{fig:experimentation-2}
\end{figure}

Fig.~\ref{fig:experimentation-2} presents the second process model, which has a loop.
WoMine finds a frequent pattern appearing in the 70\% of the traces (Fig.~\ref{fig:experimentation-custom-2}).
The heat maps approach, nevertheless, retrieves as frequent the structure with the execution of the loop (Fig.~\ref{fig:experimentation-custom-2-arcs}).
$w$-find gets the same pattern as WoMine, but with a wrong frequency.
As has been explained before, when \texttt{H} is executed in a trace, $w$-find can not distinguish if it is a loop disrupting the execution of the pattern, or a task in other parallel branch of the model.
In the local process mining technique, the pattern is also registered as executed in the traces with \texttt{H}, retrieving the same result as $w$-find.
The episode mining approach also retrieves, among other patterns with the same tasks but different relations, this pattern with a wrong frequency due to the same reason.
PrefixSpan and the tree mining approach cannot get structures with parallels because they interpret that traces with \texttt{E} before \texttt{F} are different than those with \texttt{F} before \texttt{E}.

\begin{figure}[H]
    \centering
    \subfloat[WoMine pattern (highlighted) for a threshold of 55\%. 
    This structure corresponds the instances of the process model where \texttt{E} and \texttt{I} are not executed, and the loop of \texttt{J} is executed at least one time.]{
        \includegraphics[width=0.42\textwidth]{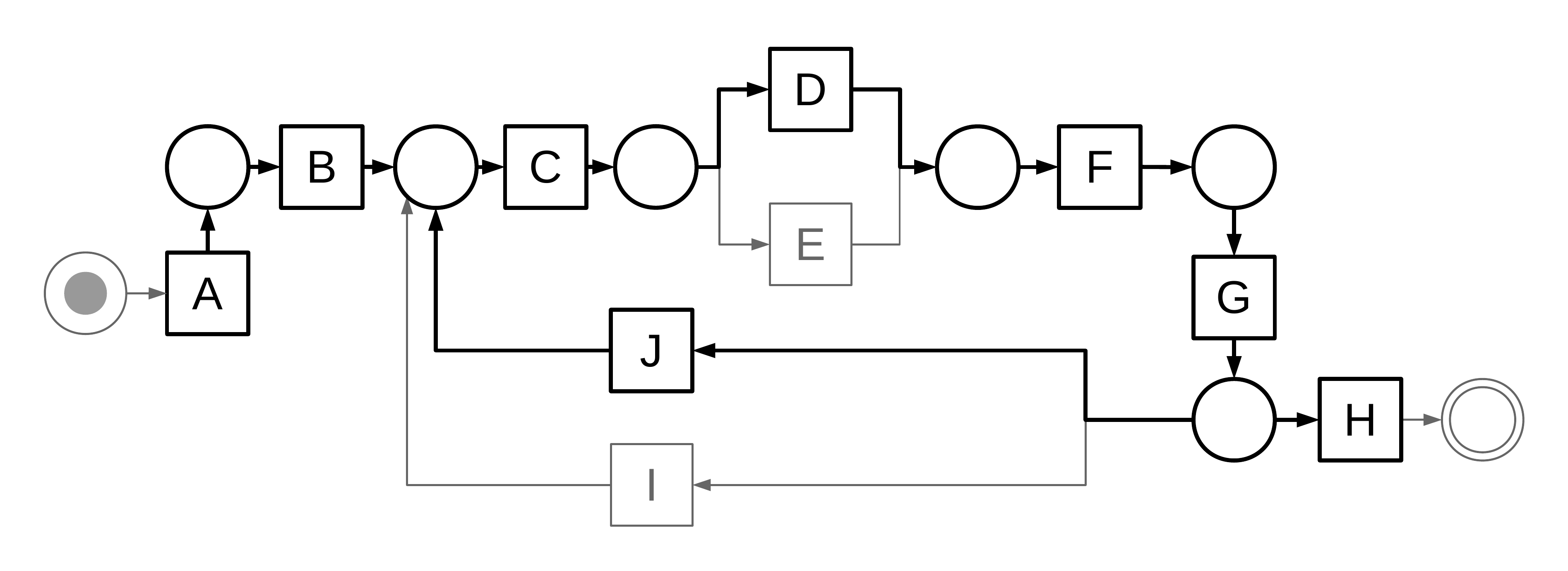}
        \label{fig:experimentation-custom-3}
    }
    \hfill
    \subfloat[Heat maps pattern (highlighted) for a threshold of 100.
    The arcs of the \texttt{I} loop are individually frequent, but the highlighted structure is not.]{
        \includegraphics[width=0.42\textwidth]{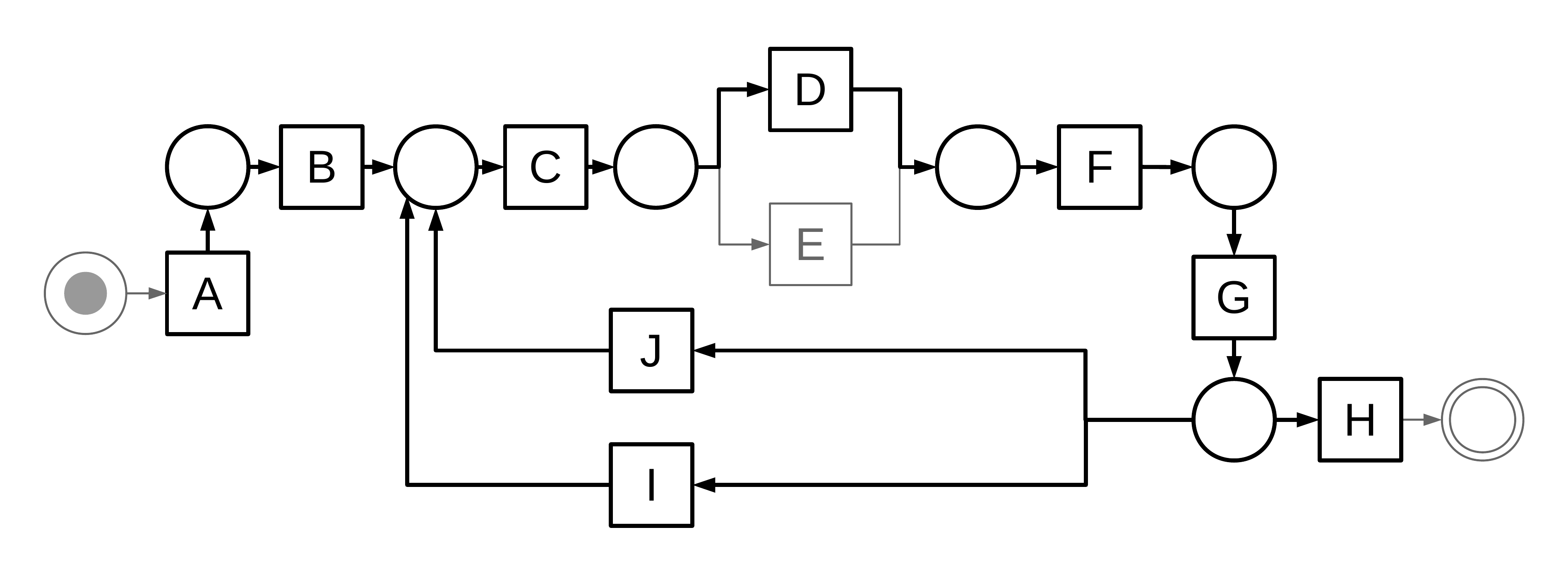}
        \label{fig:experimentation-custom-3-arcs}
    }
    \caption{Results of WoMine and heat maps for a process model composed by a sequence with a selection, and two loops sharing the start and end tasks.
    The loops give the model the ability to execute both branches of the selection in the same trace, and more than once.}
    \label{fig:experimentation-3}
\end{figure}

In the third scenario, Fig.~\ref{fig:experimentation-custom-3} depicts interesting behaviour found by WoMine.
In the 55\% of the traces, the pattern performs the sequence through \texttt{D}, followed by the loop with \texttt{J} and \texttt{D} again, without including neither \texttt{I} nor \texttt{E}.
On the contrary, the heat maps (Fig.~\ref{fig:experimentation-custom-3-arcs}) highlights wrongly the pattern with the two loops as frequent.
The $w$-find approach cannot retrieve the pattern found by WoMine due to the existence of a loop.
With the local process mining approach, the pattern is found, but the frequency is incorrect when the loop (\texttt{J}) is not executed, or when \texttt{I} is executed, because the final mark is reached in both cases.
The episode mining approach cannot find this pattern in a feasible time, due to the number of combinations among tasks that it checks.
PrefixSpan gets the pattern retrieved by WoMine replacing the loop with a sequence with one repetition, i.e., duplicating the tasks in the sequence.
Also, PrefixSpan does not ensure that \texttt{I} or \texttt{E} are not executed.
The tree mining approach presents the same drawback, but in a tree structure.

\begin{figure}[H]
    \centering
    \subfloat[WoMine pattern (highlighted) for a threshold of 65\%.]{
        \includegraphics[width=0.42\textwidth]{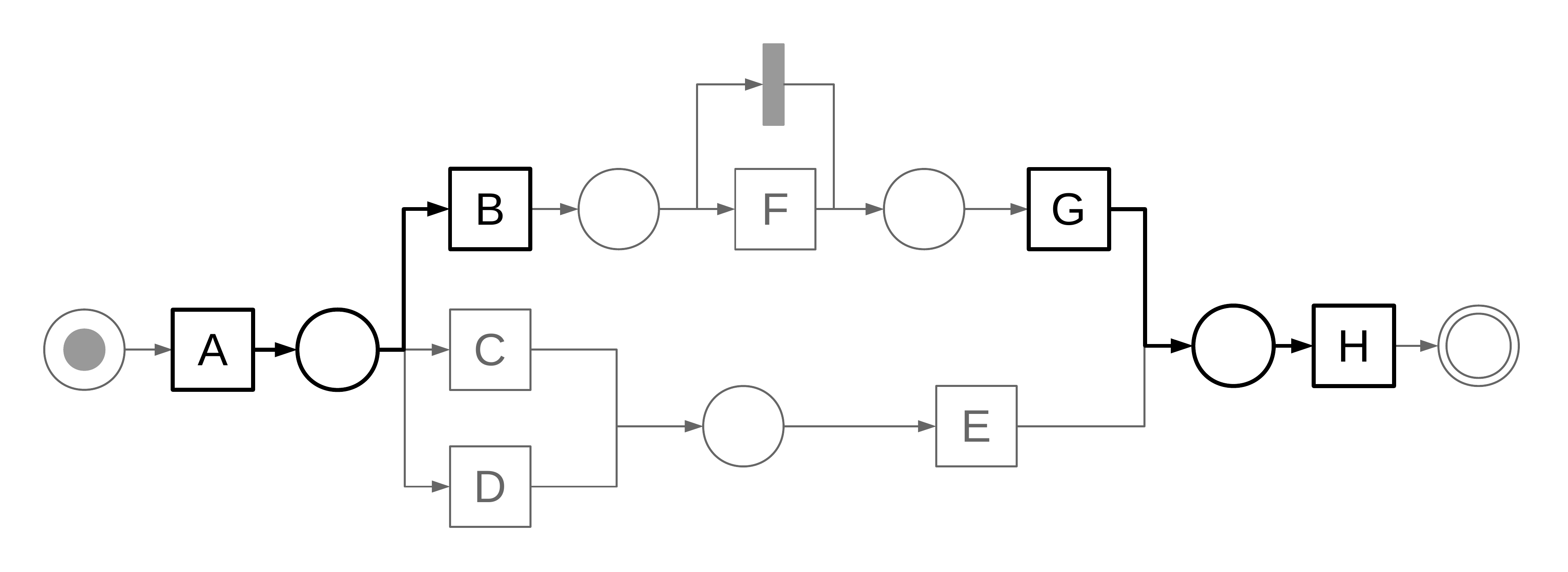}
        \label{fig:experimentation-custom-5}
    }
    \hfill
    \subfloat[Results of the heat maps approach applied with a threshold of 65. The result is correct and agrees with WoMine's.]{
        \includegraphics[width=0.42\textwidth]{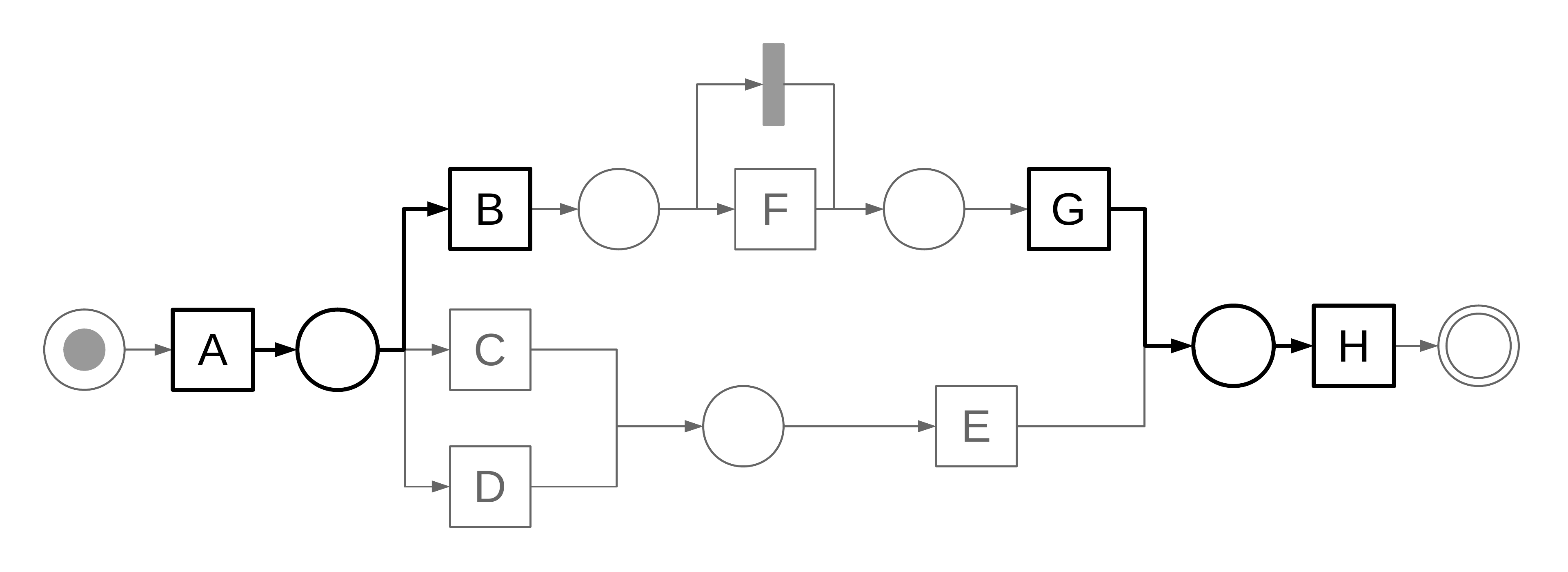}
        \label{fig:experimentation-custom-5-arcs}
    }
    \caption{Results of WoMine and heat maps for a process model with a selection of three branches, having one of them an optional task.}
    \label{fig:experimentation-5}
\end{figure}

WoMine finds, for the process model in Fig.~\ref{fig:experimentation-5}, two frequent patterns composed each one by an arc (Fig.~\ref{fig:experimentation-custom-5}) and separated by a selection.
The knowledge extracted by heat maps (Fig.~\ref{fig:experimentation-custom-5-arcs}) agrees with the result of WoMine.
All the other techniques retrieve, among others, the sequence \texttt{A-B-G-H} as frequent with a 65\%.
This is because they do not take into account the process model while checking the frequency of the pattern, and they consider that the execution of \texttt{F} is due to the other parallel branch in the model.

\begin{figure}[t]
    \centering
    \subfloat[WoMine pattern (highlighted) for a threshold of 55\%.
    The pattern consists in a parallel structure with \texttt{C} in the upper branch, the loop of \texttt{I-J}, and passing again through \texttt{C}.]{
        \includegraphics[width=0.42\textwidth]{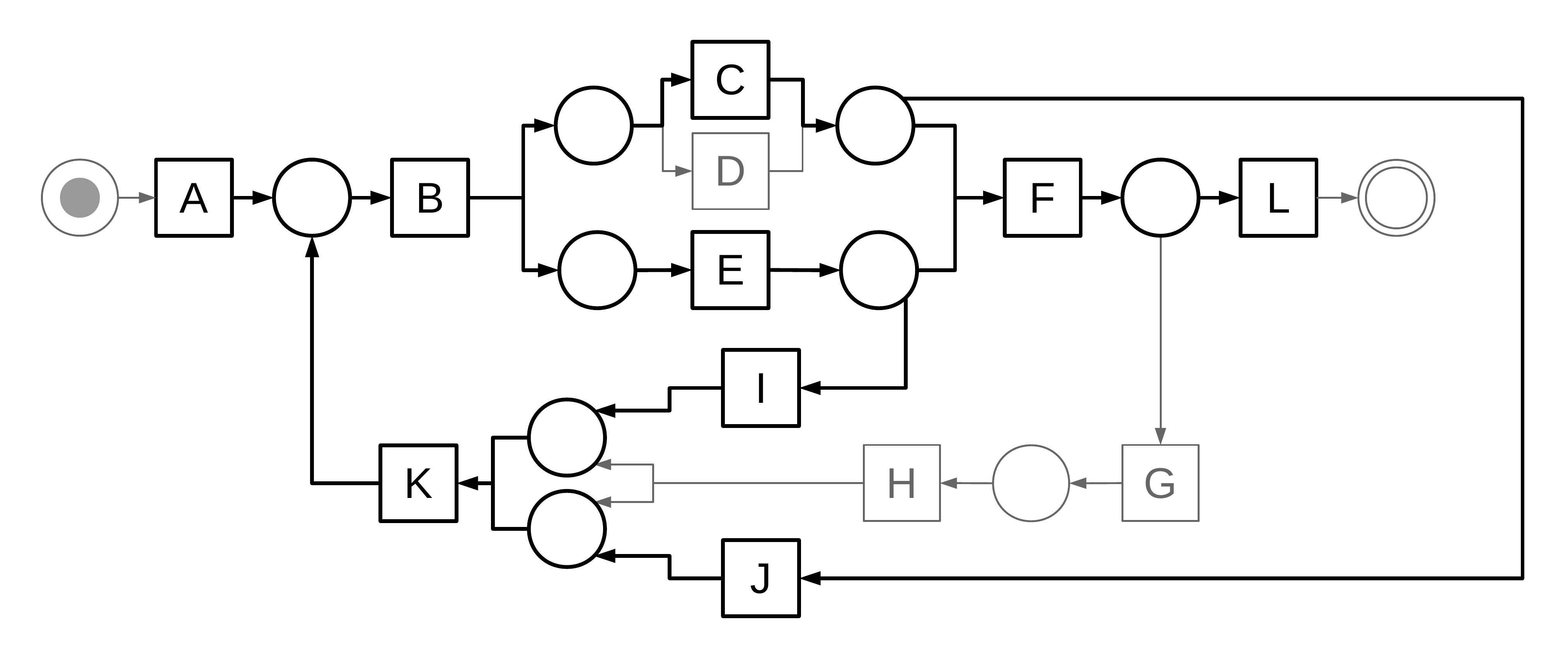}
        \label{fig:experimentation-custom-4}
    }
    \hfill
    \subfloat[Heat maps pattern (highlighted) for a threshold of 55.
    The repetition of the loops increases the frequency of other arcs, without building a recognizable pattern.]{
        \includegraphics[width=0.42\textwidth]{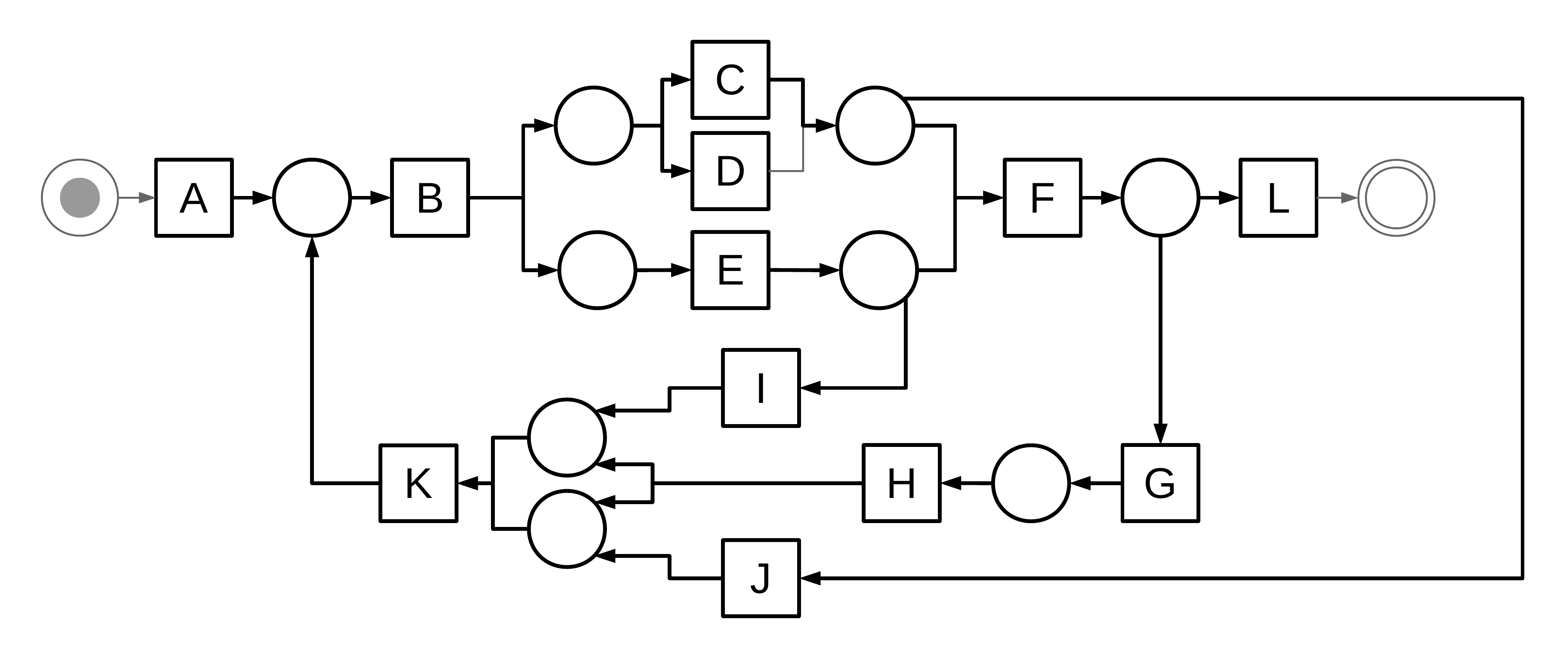}
        \label{fig:experimentation-custom-4-arcs}
    }
    \caption{Results of WoMine and heat maps for a process model with two loops sharing the end task, and one of them starting in a parallel structure.}
    \label{fig:experimentation-4}
\end{figure}

Finally, in the last case (Fig.~\ref{fig:experimentation-4}), Fig.~\ref{fig:experimentation-custom-4} shows a frequent structure executed in the 55\% of the traces.
The structural complexity of this pattern makes impossible its extraction by the heat maps approach (Fig.~\ref{fig:experimentation-custom-4-arcs}).
Similar to the Fig.~\ref{fig:experimentation-3} case, $w$-find cannot retrieve the pattern, local process mining retrieves it with a wrong frequency, and the episode mining approach fails due to the high number of possible relations between the tasks.
PrefixSpan and the tree mining approach cannot detect this pattern due to the parallel structures.

In summary, the state of the art algorithms fail to detect several patterns that can be retrieved with WoMine.
The heat maps approach is very simple, and allows to extract interesting behaviour in a quick way, but with the inability to detect which task causes the firing of another task.
The $w$-find algorithm uses the model to extract frequent patterns but, the approach to check the compliance of a trace with a pattern fails in some cases.
Loops and choices are also unsupported by this algorithm.
The local process mining technique does not ensure that the pattern was entirely executed, without disruptions in the middle of its execution.
Moreover, it does not ensure that the loops of a pattern were executed when the final marking is reached either.
The episode mining approach does not use the process model, which causes the extraction of a high number of similar patterns varying the relations between the tasks.
Also, two sequences separated by infrequent selections are detected as one single sequence by this technique.
The sequential pattern mining search, represented by PrefixSpan, has a similar problem.
Moreover, it can only detect sequences.
Finally, the tree mining approach obtains results similar to sequential pattern mining, but searching in a tree structure.

\subsection{Characteristics of the patterns and analysis of runtimes\label{subsec:de-medeiros-logs}}

\begin{table}[b!]
    \centering
    {\scriptsize
        \begin{tabular}{l|r|r|r|r|r|r|r|r|r|}

        \cline{2-10}
        & \multicolumn{9}{c|}{Threshold : 40\%}                \\ \cline{2-10} 
        & \multicolumn{2}{c|}{runtime (secs)} & \multicolumn{1}{c|}{\multirow{2}{*}{\#patt}}  & \multicolumn{1}{c|}{\multirow{2}{*}{frequency}}   & \multicolumn{1}{c|}{\multirow{2}{*}{\#tasks}}   & \multicolumn{1}{c|}{\multirow{2}{*}{\#sequences}}   & \multicolumn{1}{c|}{\multirow{2}{*}{\#choices}} & \multicolumn{1}{c|}{\multirow{2}{*}{\#parallels}} & \multicolumn{1}{c|}{\multirow{2}{*}{\#loops}}   \\ \cline{2-3} 
        & \multicolumn{1}{c|}{pre}  & \multicolumn{1}{c|}{alg}                      & \multicolumn{1}{c|}{}                             & \multicolumn{1}{c|}{}                         & \multicolumn{1}{c|}{}                             & \multicolumn{1}{c|}{}                         & \multicolumn{1}{c|}{}                           &  \multicolumn{1}{c|}{}            & \multicolumn{1}{c|}{}                         \\ \hline

        \multicolumn{1}{|l|}{g2}  & 0.005 &  0.011 & 2 & 0.48$\pm$0.04 & 4.50$\pm$0.71 & 1.00$\pm$0.00 & 0$\pm$0 & 0$\pm$0 & 0$\pm$0 \\ \hline
        \multicolumn{1}{|l|}{g3}  & 0.060 &  6.800 & 4 & 0.45$\pm$0.04 & 14.25$\pm$5.12 & 1.75$\pm$0.96 & 1.50$\pm$1.29 & 3.00$\pm$0.82 & 0.25$\pm$0.50 \\ \hline
        \multicolumn{1}{|l|}{g4}  & 0.010 &  0.236 & 4 & 0.62$\pm$0.25 & 4.50$\pm$2.38 & 0.75$\pm$0.50 & 0.25$\pm$0.50 & 0$\pm$0 & 0.25$\pm$0.50 \\ \hline
        \multicolumn{1}{|l|}{g5}  & 0.006 &  0.066 & 3 & 0.50$\pm$0.03 & 8.00$\pm$5.29 & 1.00$\pm$1.00 & 0.67$\pm$0.58 & 1.33$\pm$1.15 & 0$\pm$0 \\ \hline
        \multicolumn{1}{|l|}{g6}  & 0.007 &  0.084 & 3 & 0.67$\pm$0.29 & 6.67$\pm$3.06 & 0.67$\pm$0.58 & 0.67$\pm$1.15 & 0.67$\pm$1.15 & 0$\pm$0 \\ \hline
        \multicolumn{1}{|l|}{g7}  & 0.046 &  6.029 & 8 & 0.48$\pm$0.04 & 16.00$\pm$2.20 & 1.75$\pm$0.46 & 1.00$\pm$1.20 & 1.75$\pm$0.71 & 0.25$\pm$0.46 \\ \hline
        \multicolumn{1}{|l|}{g8}  & 0.015 &  0.082 & 4 & 0.72$\pm$0.33 & 4.25$\pm$1.26 & 0.75$\pm$0.50 & 0.25$\pm$0.50 & 0.50$\pm$1.00 & 0$\pm$0 \\ \hline
        \multicolumn{1}{|l|}{g9}  & 0.007 &  0.039 & 3 & 0.50$\pm$0.02 & 6.33$\pm$0.58 & 1.00$\pm$0.00 & 0.33$\pm$0.58 & 0.33$\pm$0.58 & 0$\pm$0 \\ \hline
        \multicolumn{1}{|l|}{g10} & 0.006 &  0.043 & 1 & 0.49$\pm$0.00 & 14.00$\pm$0.00 & 1.00$\pm$0.00 & 2.00$\pm$0.00 & 2.00$\pm$0.00 & 0$\pm$0 \\ \hline
        \multicolumn{1}{|l|}{g12} & 0.009 &  0.061 & 3 & 0.67$\pm$0.29 & 6.00$\pm$3.00 & 1.00$\pm$0.00 & 0.33$\pm$0.58 & 0.67$\pm$1.15 & 0$\pm$0 \\ \hline
        \multicolumn{1}{|l|}{g13} & 0.002 &  0.025 & 1 & 0.48$\pm$0.00 & 13.00$\pm$0.00 & 2.00$\pm$0.00 & 2.00$\pm$0.00 & 0$\pm$0 & 0$\pm$0 \\ \hline
        \multicolumn{1}{|l|}{g14} & 0.019 &  0.179 & 5 & 0.59$\pm$0.23 & 6.00$\pm$1.87 & 1.20$\pm$0.45 & 0$\pm$0 & 0$\pm$0 & 0.60$\pm$0.55 \\ \hline
        \multicolumn{1}{|l|}{g15} & 0.013 &  0.009 & 3 & 0.76$\pm$0.24 & 2.67$\pm$1.15 & 0.33$\pm$0.58 & 0$\pm$0 & 0$\pm$0 & 0$\pm$0 \\ \hline
        \multicolumn{1}{|l|}{g19} & 0.002 &  0.015 & 1 & 0.47$\pm$0.00 & 11.00$\pm$0.00 & 2.00$\pm$0.00 & 1.00$\pm$0.00 & 2.00$\pm$0.00 & 0$\pm$0 \\ \hline
        \multicolumn{1}{|l|}{g20} & 0.022 &  0.235 & 6 & 0.58$\pm$0.21 & 4.67$\pm$2.66 & 0.83$\pm$0.75 & 0.50$\pm$0.55 & 0$\pm$0 & 0$\pm$0 \\ \hline
        \multicolumn{1}{|l|}{g21} & 0.001 &  0.007 & 2 & 0.75$\pm$0.36 & 5.00$\pm$4.24 & 0.50$\pm$0.71 & 0.50$\pm$0.71 & 0$\pm$0 & 0$\pm$0 \\ \hline
        \multicolumn{1}{|l|}{g22} & 0.001 &  0.006 & 1 & 0.44$\pm$0.00 & 8.00$\pm$0.00 & 1.00$\pm$0.00 & 1.00$\pm$0.00 & 1.00$\pm$0.00 & 0$\pm$0 \\ \hline
        \multicolumn{1}{|l|}{g23} & 0.052 &  0.424 & 6 & 0.82$\pm$0.28 & 3.00$\pm$0.89 & 0.33$\pm$0.52 & 0$\pm$0 & 0$\pm$0 & 0.33$\pm$0.52 \\ \hline
        \multicolumn{1}{|l|}{g24} & 0.002 &  0.014 & 4 & 0.65$\pm$0.24 & 3.50$\pm$1.91 & 0$\pm$0 & 0.25$\pm$0.50 & 0.75$\pm$0.96 & 0$\pm$0 \\ \hline
        \multicolumn{1}{|l|}{g25} & 0.041 &  0.326 & 4 & 0.49$\pm$0.04 & 6.50$\pm$4.43 & 0.50$\pm$0.58 & 0.50$\pm$0.58 & 1.25$\pm$2.50 & 0.25$\pm$0.50 \\ \hline

        \end{tabular}
        \caption{Behavioral structure of the frequent patterns extracted, for a threshold of 40\%, from the process models in~\cite{phdGenetic}. The values show the runtimes for the preprocessing and for the algorithm; the number of patterns retrieved; the average and standard deviation of the frequency, number of tasks, number of sequences, number of choices, number of parallels and number of loops per pattern.}
        \label{tab:results-1}
    }
\end{table}

This subsection presents the results obtained by WoMine in a set of process models for different thresholds.
Tables~\ref{tab:results-1} and~\ref{tab:results-2} show the result of WoMine for 20 process models and three different thresholds.
For instance, for process model \textit{g25}, with a threshold of the 40\%, WoMine discovered four patterns.
These patterns have a frequency close to the 50\% and, in average, 6.5 tasks per pattern, with a standard deviation of 4.43, containing all kind of structures ---sequences, choices, parallels and loops.
As explained in Sec.~\ref{sec:measure-frequency}, the algorithm needs to execute the trace in the model to retrieve the executed arcs.
This process is independent of the threshold ---it only depends on the traces (log) and on the model.
Thus, the runtime is divided in two parts to distinguish this preprocessing time and the time spent by the  algorithm.

The event logs were randomly generated ---up to 300 traces--- from process models with different complexity levels, ranging from 20 to 30 unique tasks, and containing loops, parallelisms, selections, etc. A more detailed description of the behavioral structures of these process models can be found in~\cite{phdGenetic,vazquez2015prodigen}.
Furthermore, as WoMine takes as starting point a process model and an event log, we used ProDiGen~\cite{vazquez2015prodigen} over this set of event logs to retrieve the process model.

As can be seen, WoMine is able to retrieve frequent patterns with all type of structures.
When the threshold increases, WoMine obtains less and simpler patterns.
This is because, as the minimum frequency is increased, more patterns become infrequent and stop belonging to the result set, which also reduces the possibilities for growing the patterns.
Nevertheless, there are some cases where the number of frequent patterns becomes higher when the threshold increases (g10, g13, g21, etc.).
This happens when a large pattern is splitted due to the increase of the threshold, as some parts of it become infrequent and trigger a disjointed structure.
For instance, model \textit{g10} has a frequent pattern with 14 tasks for a threshold of 40\%, and two patterns for 60\% but with an average of 2.5 tasks per pattern.

\begin{table}[H]
    \centering
    {\scriptsize
        \begin{tabular}{l|r|r|r|r|r|r|r|r|r|}

        \cline{2-10}
        & \multicolumn{9}{c|}{Threshold : 60\%}              \\ \cline{2-10} 
        & \multicolumn{2}{c|}{runtime (secs)} & \multicolumn{1}{c|}{\multirow{2}{*}{\#patt}}  & \multicolumn{1}{c|}{\multirow{2}{*}{frequency}}   & \multicolumn{1}{c|}{\multirow{2}{*}{\#tasks}}   & \multicolumn{1}{c|}{\multirow{2}{*}{\#sequences}}   & \multicolumn{1}{c|}{\multirow{2}{*}{\#choices}} & \multicolumn{1}{c|}{\multirow{2}{*}{\#parallels}} & \multicolumn{1}{c|}{\multirow{2}{*}{\#loops}}   \\ \cline{2-3} 
        & \multicolumn{1}{c|}{pre}  & \multicolumn{1}{c|}{alg}                      & \multicolumn{1}{c|}{}                             & \multicolumn{1}{c|}{}                         & \multicolumn{1}{c|}{}                             & \multicolumn{1}{c|}{}                         & \multicolumn{1}{c|}{}                           &  \multicolumn{1}{c|}{}            & \multicolumn{1}{c|}{}                         \\ \hline

        \multicolumn{1}{|l|}{g2}  & 0.005 &  0.006 & 2 & 0.87$\pm$0.18 & 3.50$\pm$0.71 & 1.00$\pm$0.00 & 0$\pm$0 & 0$\pm$0 & 0$\pm$0 \\ \hline
        \multicolumn{1}{|l|}{g3}  & 0.060 &  0.609 & 4 & 0.83$\pm$0.20 & 6.75$\pm$4.11 & 1.00$\pm$0.00 & 0.50$\pm$1.00 & 1.25$\pm$1.50 & 0.25$\pm$0.50 \\ \hline
        \multicolumn{1}{|l|}{g4}  & 0.010 &  0.011 & 4 & 1.00$\pm$0.00 & 3.25$\pm$1.89 & 0.50$\pm$0.58 & 0$\pm$0 & 0$\pm$0 & 0$\pm$0 \\ \hline
        \multicolumn{1}{|l|}{g5}  & 0.008 &  0.012 & 3 & 0.89$\pm$0.19 & 4.33$\pm$0.58 & 0.67$\pm$0.58 & 0$\pm$0 & 0.67$\pm$1.15 & 0$\pm$0 \\ \hline
        \multicolumn{1}{|l|}{g6}  & 0.007 &  0.014 & 2 & 1.00$\pm$0.00 & 4.00$\pm$2.83 & 0.50$\pm$0.71 & 0$\pm$0 & 0$\pm$0 & 0$\pm$0 \\ \hline
        \multicolumn{1}{|l|}{g7}  & 0.046 &  1.228 & 6 & 0.83$\pm$0.18 & 10.50$\pm$4.85 & 1.17$\pm$0.41 & 0.33$\pm$0.82 & 1.33$\pm$1.03 & 0.17$\pm$0.41 \\ \hline
        \multicolumn{1}{|l|}{g8}  & 0.015 &  0.010 & 2 & 1.00$\pm$0.00 & 3.50$\pm$0.71 & 1.00$\pm$0.00 & 0$\pm$0 & 0$\pm$0 & 0$\pm$0 \\ \hline
        \multicolumn{1}{|l|}{g9}  & 0.007 &  0.011 & 2 & 1.00$\pm$0.00 & 4.00$\pm$1.41 & 1.00$\pm$0.00 & 0$\pm$0 & 0$\pm$0 & 0$\pm$0 \\ \hline
        \multicolumn{1}{|l|}{g10} & 0.006 &  0.004 & 2 & 1.00$\pm$0.00 & 2.50$\pm$0.71 & 0.50$\pm$0.71 & 0$\pm$0 & 0$\pm$0 & 0$\pm$0 \\ \hline
        \multicolumn{1}{|l|}{g12} & 0.009 &  0.002 & 1 & 1.00$\pm$0.00 & 3.00$\pm$0.00 & 1.00$\pm$0.00 & 0$\pm$0 & 0$\pm$0 & 0$\pm$0 \\ \hline
        \multicolumn{1}{|l|}{g13} & 0.002 &  0.006 & 2 & 1.00$\pm$0.00 & 5.50$\pm$0.71 & 1.00$\pm$0.00 & 0$\pm$0 & 0$\pm$0 & 0$\pm$0 \\ \hline
        \multicolumn{1}{|l|}{g14} & 0.019 &  0.156 & 5 & 1.00$\pm$0.00 & 5.20$\pm$1.10 & 1.00$\pm$0.00 & 0$\pm$0 & 0$\pm$0 & 0$\pm$0 \\ \hline
        \multicolumn{1}{|l|}{g15} & 0.013 &  0.005 & 3 & 0.92$\pm$0.14 & 2.33$\pm$0.58 & 0.33$\pm$0.58 & 0$\pm$0 & 0$\pm$0 & 0$\pm$0 \\ \hline
        \multicolumn{1}{|l|}{g19} & 0.002 &  0.011 & 2 & 0.86$\pm$0.19 & 5.00$\pm$1.41 & 1.00$\pm$0.00 & 0$\pm$0 & 1.00$\pm$1.41 & 0$\pm$0 \\ \hline
        \multicolumn{1}{|l|}{g20} & 0.022 &  0.014 & 3 & 1.00$\pm$0.00 & 2.67$\pm$0.58 & 0.67$\pm$0.58 & 0$\pm$0 & 0$\pm$0 & 0$\pm$0 \\ \hline
        \multicolumn{1}{|l|}{g21} & 0.001 &  0.003 & 3 & 0.89$\pm$0.18 & 3.00$\pm$1.00 & 0.67$\pm$0.58 & 0$\pm$0 & 0$\pm$0 & 0$\pm$0 \\ \hline
        \multicolumn{1}{|l|}{g22} & 0.001 &  0.001 & 2 & 1.00$\pm$0.00 & 2.50$\pm$0.71 & 0.50$\pm$0.71 & 0$\pm$0 & 0$\pm$0 & 0$\pm$0 \\ \hline
        \multicolumn{1}{|l|}{g23} & 0.052 &  0.317 & 6 & 1.00$\pm$0.00 & 2.67$\pm$1.03 & 0.33$\pm$0.52 & 0$\pm$0 & 0$\pm$0 & 0$\pm$0 \\ \hline
        \multicolumn{1}{|l|}{g24} & 0.002 &  0.001 & 2 & 1.00$\pm$0.00 & 2.00$\pm$0.00 & 0$\pm$0 & 0$\pm$0 & 0$\pm$0 & 0$\pm$0 \\ \hline
        \multicolumn{1}{|l|}{g25} & 0.041 &  0.065 & 5 & 1.00$\pm$0.00 & 3.20$\pm$1.10 & 0.20$\pm$0.45 & 0$\pm$0 & 0.80$\pm$1.10 & 0$\pm$0 \\

            \cline{1-10}

        \cline{2-10}
        & \multicolumn{9}{c|}{Threshold : 80\%}              \\ \cline{2-10} 
        & \multicolumn{2}{c|}{runtime (secs)} & \multicolumn{1}{c|}{\multirow{2}{*}{\#patt}}  & \multicolumn{1}{c|}{\multirow{2}{*}{frequency}}   & \multicolumn{1}{c|}{\multirow{2}{*}{\#tasks}}   & \multicolumn{1}{c|}{\multirow{2}{*}{\#sequences}}   & \multicolumn{1}{c|}{\multirow{2}{*}{\#choices}} & \multicolumn{1}{c|}{\multirow{2}{*}{\#parallels}} & \multicolumn{1}{c|}{\multirow{2}{*}{\#loops}}   \\ \cline{2-3} 
        & \multicolumn{1}{c|}{pre}  & \multicolumn{1}{c|}{alg}                      & \multicolumn{1}{c|}{}                             & \multicolumn{1}{c|}{}                         & \multicolumn{1}{c|}{}                             & \multicolumn{1}{c|}{}                         & \multicolumn{1}{c|}{}                           &  \multicolumn{1}{c|}{}            & \multicolumn{1}{c|}{}                         \\ \hline

        \multicolumn{1}{|l|}{g2}  & 0.005 &  0.004 & 2 & 1.00$\pm$0.00 & 3.00$\pm$1.41 & 0.50$\pm$0.71 & 0$\pm$0 & 0$\pm$0 & 0$\pm$0 \\ \hline
        \multicolumn{1}{|l|}{g3}  & 0.060 &  0.073 & 3 & 1.00$\pm$0.00 & 5.00$\pm$2.65 & 1.00$\pm$0.00 & 0$\pm$0 & 0.67$\pm$1.15 & 0$\pm$0 \\ \hline
        \multicolumn{1}{|l|}{g4}  & 0.010 &  0.010 & 4 & 1.00$\pm$0.00 & 3.25$\pm$1.89 & 0.50$\pm$0.58 & 0$\pm$0 & 0$\pm$0 & 0$\pm$0 \\ \hline
        \multicolumn{1}{|l|}{g5}  & 0.008 &  0.007 & 2 & 1.00$\pm$0.00 & 4.00$\pm$0.00 & 1.00$\pm$0.00 & 0$\pm$0 & 0$\pm$0 & 0$\pm$0 \\ \hline
        \multicolumn{1}{|l|}{g6}  & 0.007 &  0.008 & 2 & 1.00$\pm$0.00 & 4.00$\pm$2.83 & 0.50$\pm$0.71 & 0$\pm$0 & 0$\pm$0 & 0$\pm$0 \\ \hline
        \multicolumn{1}{|l|}{g7}  & 0.046 &  0.364 & 3 & 1.00$\pm$0.00 & 9.00$\pm$7.21 & 1.33$\pm$0.58 & 0$\pm$0 & 0.67$\pm$1.15 & 0$\pm$0 \\ \hline
        \multicolumn{1}{|l|}{g8}  & 0.015 &  0.009 & 2 & 1.00$\pm$0.00 & 3.50$\pm$0.71 & 1.00$\pm$0.00 & 0$\pm$0 & 0$\pm$0 & 0$\pm$0 \\ \hline
        \multicolumn{1}{|l|}{g9}  & 0.007 &  0.010 & 2 & 1.00$\pm$0.00 & 4.00$\pm$1.41 & 1.00$\pm$0.00 & 0$\pm$0 & 0$\pm$0 & 0$\pm$0 \\ \hline
        \multicolumn{1}{|l|}{g10} & 0.006 &  0.004 & 2 & 1.00$\pm$0.00 & 2.50$\pm$0.71 & 0.50$\pm$0.71 & 0$\pm$0 & 0$\pm$0 & 0$\pm$0 \\ \hline
        \multicolumn{1}{|l|}{g12} & 0.009 &  0.001 & 1 & 1.00$\pm$0.00 & 3.00$\pm$0.00 & 1.00$\pm$0.00 & 0$\pm$0 & 0$\pm$0 & 0$\pm$0 \\ \hline
        \multicolumn{1}{|l|}{g13} & 0.002 &  0.006 & 2 & 1.00$\pm$0.00 & 5.50$\pm$0.71 & 1.00$\pm$0.00 & 0$\pm$0 & 0$\pm$0 & 0$\pm$0 \\ \hline
        \multicolumn{1}{|l|}{g14} & 0.019 &  0.151 & 5 & 1.00$\pm$0.00 & 5.20$\pm$1.10 & 1.00$\pm$0.00 & 0$\pm$0 & 0$\pm$0 & 0$\pm$0 \\ \hline
        \multicolumn{1}{|l|}{g15} & 0.013 &  0.004 & 2 & 1.00$\pm$0.00 & 2.50$\pm$0.71 & 0.50$\pm$0.71 & 0$\pm$0 & 0$\pm$0 & 0$\pm$0 \\ \hline
        \multicolumn{1}{|l|}{g19} & 0.002 &  0.005 & 2 & 1.00$\pm$0.00 & 4.50$\pm$2.12 & 1.00$\pm$0.00 & 0$\pm$0 & 1.00$\pm$1.41 & 0$\pm$0 \\ \hline
        \multicolumn{1}{|l|}{g20} & 0.022 &  0.012 & 3 & 1.00$\pm$0.00 & 2.67$\pm$0.58 & 0.67$\pm$0.58 & 0$\pm$0 & 0$\pm$0 & 0$\pm$0 \\ \hline
        \multicolumn{1}{|l|}{g21} & 0.001 &  0.001 & 2 & 1.00$\pm$0.00 & 3.00$\pm$1.41 & 0.50$\pm$0.71 & 0$\pm$0 & 0$\pm$0 & 0$\pm$0 \\ \hline
        \multicolumn{1}{|l|}{g22} & 0.001 &  0.001 & 2 & 1.00$\pm$0.00 & 2.50$\pm$0.71 & 0.50$\pm$0.71 & 0$\pm$0 & 0$\pm$0 & 0$\pm$0 \\ \hline
        \multicolumn{1}{|l|}{g23} & 0.052 &  0.329 & 6 & 1.00$\pm$0.00 & 2.67$\pm$1.03 & 0.33$\pm$0.52 & 0$\pm$0 & 0$\pm$0 & 0$\pm$0 \\ \hline
        \multicolumn{1}{|l|}{g24} & 0.002 &  0.001 & 2 & 1.00$\pm$0.00 & 2.00$\pm$0.00 & 0$\pm$0 & 0$\pm$0 & 0$\pm$0 & 0$\pm$0 \\ \hline
        \multicolumn{1}{|l|}{g25} & 0.041 &  0.061 & 5 & 1.00$\pm$0.00 & 3.20$\pm$1.10 & 0.20$\pm$0.45 & 0$\pm$0 & 0.80$\pm$1.10 & 0$\pm$0 \\ \hline

        \end{tabular}
        \caption{Continuation of results in Table~\ref{tab:results-1} for thresholds of 60\% and 80\%.}
        \label{tab:results-2}
    }
\end{table}

Regarding the runtime of the algorithm, the preprocessing time is always under 60 ms, usually 20 ms.
This is the time to parse the 300 traces, and retrieve the executed arcs.
On the other hand, the runtime of the algorithm decreases when the threshold increases, as more patterns become infrequent and are pruned earlier, saving computational cost.
The global runtime ---preprocessing plus algorithm--- is under 500 milliseconds in all cases except the executions of \textit{g3} and \textit{g7}, both with thresholds of 40\% and 60\%.

\subsection{Frequent patterns for the BPI Challenges\label{subsec:bpic-logs}}

\begin{table}[b!]
    \centering
    {\footnotesize
    \def\arraystretch{.9}
    \subfloat[Statistics of the BPIC logs. The logs \textit{a} and \textit{o} from BPIC 2012 have been generated after a filtering in the log, maintaining the traces with tasks of the categories A and O, respectively.\label{tab:stats-logs}]{
        \begin{tabular}{cc|r|r|r|r|r|}
            \cline{3-7}
            \multicolumn{1}{l}{} & \multicolumn{1}{l|}{} & \multicolumn{1}{c|}{\multirow{2}{*}{\#traces}} & \multicolumn{1}{c|}{\multirow{2}{*}{\#events}} & \multicolumn{3}{c|}{events per trace}                     \\ \cline{5-7} 
            \multicolumn{1}{l}{} & \multicolumn{1}{l|}{} & \multicolumn{1}{c|}{}         & \multicolumn{1}{c|}{}         & \multicolumn{1}{c|}{min} & \multicolumn{1}{c|}{max} & \multicolumn{1}{c|}{$\bar{X}\pm\sigma$} \\ \hline
            \multicolumn{2}{|c|}{2011}                          & 1143 & 152577 & 3 & 1816 & 133.5$\pm$202.6 \\ \hline
            \multicolumn{1}{|c|}{\multirow{3}{*}{\rotatebox[origin=c]{90}{2012}}} & fin   & 13087 & 288374 & 5 & 177 & 22.0$\pm$19.9 \\ \cline{2-7} 
            \multicolumn{1}{|c|}{}                      & a     & 4085 & 21565 & 5 & 10 & 5.3$\pm$0.8 \\ \cline{2-7} 
            \multicolumn{1}{|c|}{}                      & o     & 4038 & 32384 & 5 & 32 & 8.0$\pm$2.8 \\ \hline
            \multicolumn{1}{|c|}{\multirow{3}{*}{\rotatebox[origin=c]{90}{2013}}} & inc   & 7554 & 80641 & 3 & 125 & 10.7$\pm$7.6 \\ \cline{2-7} 
            \multicolumn{1}{|c|}{}                      & clo   & 1487 & 9634 & 3 & 37 & 6.5$\pm$3.2 \\ \cline{2-7} 
            \multicolumn{1}{|c|}{}                      & op    & 819 & 3989 & 3 & 24 & 4.9$\pm$2.1 \\ \hline
            \multicolumn{1}{|c|}{\multirow{5}{*}{\rotatebox[origin=c]{90}{2015}}} & 1     & 1199 & 54615 & 4 & 103 & 45.6$\pm$17.0 \\ \cline{2-7} 
            \multicolumn{1}{|c|}{}                      & 2     & 832 & 46018 & 3 & 134 & 55.3$\pm$19.9 \\ \cline{2-7} 
            \multicolumn{1}{|c|}{}                      & 3     & 1409 & 62499 & 5 & 126 & 44.4$\pm$16.1 \\ \cline{2-7} 
            \multicolumn{1}{|c|}{}                      & 4     & 1053 & 49399 & 3 & 118 & 46.9$\pm$15.0 \\ \cline{2-7} 
            \multicolumn{1}{|c|}{}                      & 5     & 1156 & 61395 & 7 & 156 & 53.1$\pm$16.0 \\ \hline
        \end{tabular}
    }
    \hfill
    \subfloat[Number of tasks and arcs of the mined models, generated with two discovery algorithms: Heuristics Miner and Inductive Miner.\label{tab:stats-models}]{
        \begin{tabular}{cc|r|r|r|r|}
            \cline{3-6}
            \multicolumn{1}{l}{}                        & \multicolumn{1}{l|}{} & \multicolumn{2}{c|}{Heuristics Miner}                               & \multicolumn{2}{c|}{Inductive Miner}                               \\ \cline{3-6} 
            \multicolumn{1}{l}{}                        & \multicolumn{1}{l|}{} & \multicolumn{1}{c|}{\#tasks} & \multicolumn{1}{c|}{\#arcs} & \multicolumn{1}{c|}{\#tasks} & \multicolumn{1}{c|}{\#arcs} \\ \hline
            \multicolumn{2}{|c|}{2011}                          & 623 & 1480 & 626 & 390614 \\ \hline
            \multicolumn{1}{|c|}{\multirow{3}{*}{\rotatebox[origin=c]{90}{2012}}} & fin   & 38 & 112 & 38 & 1044 \\ \cline{2-6} 
            \multicolumn{1}{|c|}{}                      & a     & 12 & 14 & 12 & 19 \\ \cline{2-6} 
            \multicolumn{1}{|c|}{}                      & o     & 9 & 17 & 9 & 28 \\ \hline
            \multicolumn{1}{|c|}{\multirow{3}{*}{\rotatebox[origin=c]{90}{2013}}} & inc   & 15 & 101 & 15 & 171 \\ \cline{2-6} 
            \multicolumn{1}{|c|}{}                      & clo   & 9 & 34 & 9 & 28 \\ \cline{2-6} 
            \multicolumn{1}{|c|}{}                      & op    & 7 & 30 & 7 & 16 \\ \hline
            \multicolumn{1}{|c|}{\multirow{5}{*}{\rotatebox[origin=c]{90}{2015}}} & 1     & 400 & 719 & 400 & 153677 \\ \cline{2-6} 
            \multicolumn{1}{|c|}{}                      & 2     & 412 & 747 & 412 & 162797 \\ \cline{2-6} 
            \multicolumn{1}{|c|}{}                      & 3     & 383 & 697 & 385 & 142887 \\ \cline{2-6} 
            \multicolumn{1}{|c|}{}                      & 4     & 358 & 635 & 358 & 113266 \\ \cline{2-6} 
            \multicolumn{1}{|c|}{}                      & 5     & 391 & 733 & 391 & 147102 \\ \hline
        \end{tabular}
        }
    \caption{Statistics about the logs of the BPICs and the mined model.}
    \label{tab:stats-bpics}
    }
\end{table}

The objective of this subsection is twofold: on the one hand, to test WoMine on complex real logs from the Business Process Intelligence Challence (BPIC)~\cite{BPIC2011} and, on the other hand, to analyze the influence of the model in the retrieved patterns.

Table~\ref{tab:stats-logs} shows some statistics of the BPIC logs~\cite{BPIC2013-clo, BPIC2013-inc, BPIC2013-op, BPIC2011, BPIC2012, BPIC2015}.
These logs have been mined with two of the most popular discovery algorithms, the Heuristics Miner (HM)~\cite{weijters2006process} and the Inductive Miner (IM)~\cite{leemans2013discovering}.
Table~\ref{tab:stats-models} presents the characteristics of the mined models, which have been generated using ProM~\cite{van2005prom}.
As can be seen, the models mined by IM contain many more arcs than the HM models.
Also, models from years 2011 and 2015 are far more complex than models from other years.


A series of experiments have been run for these logs and models with different thresholds.
Table~\ref{tab:results-bpics-1} shows the results for thresholds of 20\%, 35\% and 50\%.
We do not show higher thresholds because, for such complex models, the execution of a path many times is very uncommon.
This would return a low number of patterns, with few structures, not allowing to study the differences between models.
The results demonstrate the ability of WoMine to extract patterns with loops, choices, parallels and sequences.

\begin{table}[H]
    \resizebox{\columnwidth}{!}{
    \setlength\tabcolsep{3.5pt}
        \begin{tabular}{cc|r|r|r|r|r|r|r|r|r|r|r|r|r|r|r|r|r|r|}
\cline{3-20}
                    &                   & \multicolumn{18}{c|}{\multirow{2}{*}{Threshold : 20\%}}   \\
                    &                   & \multicolumn{18}{c|}{}                                    \\ \cline{3-20} 
                    &                   & \multicolumn{9}{c|}{Heuristics Miner}                                                                                                                                                                                                                                                               & \multicolumn{9}{c|}{Inductive Miner}                                                                                                                                                                                                                                                               \\ \cline{3-20} 
                    &                   & \multicolumn{2}{c|}{runtime (secs)}                           & \multicolumn{1}{c|}{\multirow{2}{*}{\#patt}} & \multicolumn{1}{c|}{\multirow{2}{*}{frequency}} & \multicolumn{1}{c|}{\multirow{2}{*}{\#tasks}} & \multicolumn{1}{c|}{\multirow{2}{*}{\#sequences}} & \multicolumn{1}{c|}{\multirow{2}{*}{\#choices}} & \multicolumn{1}{c|}{\multirow{2}{*}{\#parallels}} & \multicolumn{1}{c|}{\multirow{2}{*}{\#loops}}   & \multicolumn{2}{c|}{runtime (secs)}                           & \multicolumn{1}{c|}{\multirow{2}{*}{\#patt}} & \multicolumn{1}{c|}{\multirow{2}{*}{frequency}} & \multicolumn{1}{c|}{\multirow{2}{*}{\#tasks}} & \multicolumn{1}{c|}{\multirow{2}{*}{\#sequences}} & \multicolumn{1}{c|}{\multirow{2}{*}{\#choices}} & \multicolumn{1}{c|}{\multirow{2}{*}{\#parallels}} & \multicolumn{1}{c|}{\multirow{2}{*}{\#loops}}   \\ \cline{3-4} \cline{12-13}
                    &                   & \multicolumn{1}{c|}{pre} & \multicolumn{1}{c|}{alg} & \multicolumn{1}{c|}{}                        &     &   &     &   &     &   & \multicolumn{1}{c|}{pre} & \multicolumn{1}{c|}{alg} & \multicolumn{1}{c|}{}                        &   &     &     &   &     &   \\ \hline

        \multicolumn{2}{|c|}{2011}                             & 5.847 & 289.127 & 20 & 0.25$\pm$0.08 & 5.65$\pm$4.30 & 0.70$\pm$0.66 & 0.80$\pm$1.15 & 0.25$\pm$0.44 & 0.40$\pm$0.50 & - & - & - & - & - & - & - & - & - \\ \hline
        \multicolumn{1}{|l|}{\multirow{3}{*}{\rotatebox[origin=c]{90}{2012}}} & fin      & 214.118 & 6.748 & 7 & 0.29$\pm$0.05 & 3.14$\pm$2.91 & 0.29$\pm$0.49 & 0.14$\pm$0.38 & 0$\pm$0 & 0.29$\pm$0.49 & 216.847 & 52.493 & 6 & 0.25$\pm$0.07 & 6.50$\pm$4.46 & 0.33$\pm$0.52 & 0.83$\pm$1.17 & 0.83$\pm$2.04 & 0.67$\pm$0.52 \\ \cline{2-20}
        \multicolumn{1}{|l|}{}                      & a        & 0.014 & 0.029 & 1 & 0.84$\pm$0.00 & 5.00$\pm$0.00 & 1.00$\pm$0.00 & 0$\pm$0 & 0$\pm$0 & 0$\pm$0 & 0.014 & 0.011 & 1 & 0.84$\pm$0.00 & 5.00$\pm$0.00 & 1.00$\pm$0.00 & 0$\pm$0 & 0$\pm$0 & 0$\pm$0 \\ \cline{2-20}
        \multicolumn{1}{|l|}{}                      & o        & 0.068 & 0.171 & 2 & 0.24$\pm$0.01 & 5.50$\pm$0.71 & 0$\pm$0 & 1.00$\pm$1.41 & 1.50$\pm$0.71 & 1.00$\pm$1.41 & 0.077 & 0.008 & 2 & 0.75$\pm$0.35 & 2.00$\pm$0.00 & 0$\pm$0 & 0$\pm$0 & 0$\pm$0 & 0$\pm$0 \\ \hline
        \multicolumn{1}{|l|}{\multirow{3}{*}{\rotatebox[origin=c]{90}{2013}}} & inc      & 26.141 & 44.621 & 6 & 0.25$\pm$0.06 & 5.33$\pm$2.25 & 0$\pm$0 & 1.17$\pm$0.98 & 0$\pm$0 & 0.67$\pm$0.82 & 24.650 & 44.393 & 6 & 0.25$\pm$0.06 & 5.33$\pm$2.25 & 0$\pm$0 & 1.17$\pm$0.98 & 0$\pm$0 & 0.67$\pm$0.82 \\ \cline{2-20}
        \multicolumn{1}{|l|}{}                      & clo      & 0.071 & 1.295 & 4 & 0.24$\pm$0.01 & 4.50$\pm$1.29 & 0$\pm$0 & 0.75$\pm$0.50 & 0$\pm$0 & 0.25$\pm$0.50 & 0.082 & 0.017 & 2 & 0.76$\pm$0.34 & 1.50$\pm$0.71 & 0$\pm$0 & 0$\pm$0 & 0$\pm$0 & 0.50$\pm$0.71 \\ \cline{2-20}
        \multicolumn{1}{|l|}{}                      & op       & 0.021 & 0.077 & 2 & 0.21$\pm$0.00 & 4.00$\pm$0.00 & 0.50$\pm$0.71 & 1.00$\pm$1.41 & 0$\pm$0 & 0$\pm$0 & 0.023 & 0.017 & 1 & 0.30$\pm$0.00 & 1.00$\pm$0.00 & 0$\pm$0 & 0$\pm$0 & 0$\pm$0 & 1.00$\pm$0.00 \\ \hline
        \multicolumn{1}{|l|}{\multirow{5}{*}{\rotatebox[origin=c]{90}{2015}}} & 1        & 2.200 & 0.928 & 14 & 0.32$\pm$0.11 & 2.57$\pm$0.76 & 0.21$\pm$0.43 & 0$\pm$0 & 0.21$\pm$0.43 & 0$\pm$0 & 108.069 & 145.077 & 19 & 0.24$\pm$0.05 & 4.79$\pm$2.74 & 0.32$\pm$0.48 & 1.05$\pm$1.18 & 0.11$\pm$0.32 & 0$\pm$0 \\ \cline{2-20}
        \multicolumn{1}{|l|}{}                      & 2        & 0.961 & 1.298 & 15 & 0.28$\pm$0.14 & 3.27$\pm$1.58 & 0.27$\pm$0.46 & 0.07$\pm$0.26 & 0.47$\pm$0.64 & 0$\pm$0 & 90.468 & 80.182 & 26 & 0.24$\pm$0.04 & 3.27$\pm$2.07 & 0.31$\pm$0.47 & 0.42$\pm$0.76 & 0$\pm$0 & 0$\pm$0 \\ \cline{2-20}
        \multicolumn{1}{|l|}{}                      & 3        & 3.133 & 1.640 & 14 & 0.31$\pm$0.11 & 2.50$\pm$0.94 & 0.07$\pm$0.27 & 0.07$\pm$0.27 & 0.14$\pm$0.36 & 0$\pm$0 & 105.733 & 345.460 & 30 & 0.23$\pm$0.04 & 4.87$\pm$2.47 & 0.37$\pm$0.49 & 1.30$\pm$1.29 & 0.10$\pm$0.40 & 0$\pm$0 \\ \cline{2-20}
        \multicolumn{1}{|l|}{}                      & 4        & 1.532 & 1.976 & 12 & 0.35$\pm$0.20 & 4.00$\pm$2.37 & 0.17$\pm$0.39 & 0.08$\pm$0.29 & 0.50$\pm$0.67 & 0$\pm$0 & 77.010 & 9.934 & 20 & 0.24$\pm$0.03 & 4.15$\pm$2.91 & 0.55$\pm$0.60 & 0.45$\pm$0.83 & 0$\pm$0 & 0$\pm$0 \\ \cline{2-20}
        \multicolumn{1}{|l|}{}                      & 5        & 2.008 & 2.583 & 9 & 0.27$\pm$0.07 & 4.56$\pm$1.81 & 0.44$\pm$0.53 & 0.22$\pm$0.44 & 0.56$\pm$0.53 & 0$\pm$0 & 131.711 & 11.093 & 21 & 0.24$\pm$0.03 & 3.86$\pm$2.39 & 0.43$\pm$0.51 & 0.48$\pm$0.75 & 0$\pm$0 & 0$\pm$0 \\ \hline

            \cline{1-20}
            \multicolumn{20}{l}{}\\[-5px]

\cline{3-20}
                    &                   & \multicolumn{18}{c|}{\multirow{2}{*}{Threshold : 35\%}}   \\
                    &                   & \multicolumn{18}{c|}{}                                    \\ \cline{3-20} 
                    &                   & \multicolumn{9}{c|}{Heuristics Miner}                                                                                                                                                                                                                                                               & \multicolumn{9}{c|}{Inductive Miner}                                                                                                                                                                                                                                                               \\ \cline{3-20} 
                    &                   & \multicolumn{2}{c|}{runtime (secs)}                           & \multicolumn{1}{c|}{\multirow{2}{*}{\#patt}} & \multicolumn{1}{c|}{\multirow{2}{*}{frequency}} & \multicolumn{1}{c|}{\multirow{2}{*}{\#tasks}} & \multicolumn{1}{c|}{\multirow{2}{*}{\#sequences}} & \multicolumn{1}{c|}{\multirow{2}{*}{\#choices}} & \multicolumn{1}{c|}{\multirow{2}{*}{\#parallels}} & \multicolumn{1}{c|}{\multirow{2}{*}{\#loops}}   & \multicolumn{2}{c|}{runtime (secs)}                           & \multicolumn{1}{c|}{\multirow{2}{*}{\#patt}} & \multicolumn{1}{c|}{\multirow{2}{*}{frequency}} & \multicolumn{1}{c|}{\multirow{2}{*}{\#tasks}} & \multicolumn{1}{c|}{\multirow{2}{*}{\#sequences}} & \multicolumn{1}{c|}{\multirow{2}{*}{\#choices}} & \multicolumn{1}{c|}{\multirow{2}{*}{\#parallels}} & \multicolumn{1}{c|}{\multirow{2}{*}{\#loops}}   \\ \cline{3-4} \cline{12-13}   
                    &                   & \multicolumn{1}{c|}{pre} & \multicolumn{1}{c|}{alg} & \multicolumn{1}{c|}{}                        &     &   &     &   &     &   & \multicolumn{1}{c|}{pre} & \multicolumn{1}{c|}{alg} & \multicolumn{1}{c|}{}                        &   &     &     &   &     &   \\ \hline

        \multicolumn{2}{|c|}{2011}                             & 5.847 & 4.602 & 14 & 0.44$\pm$0.07 & 2.64$\pm$1.45 & 0.36$\pm$0.50 & 0.14$\pm$0.36 & 0.07$\pm$0.27 & 0.36$\pm$0.50 & - & - & - & - & - & - & - & - & - \\ \hline
        \multicolumn{1}{|l|}{\multirow{3}{*}{\rotatebox[origin=c]{90}{2012}}} & fin      & 214.118 & 0.994 & 4 & 0.38$\pm$0.01 & 4.00$\pm$2.45 & 0.50$\pm$0.58 & 0.25$\pm$0.50 & 0.25$\pm$0.50 & 0$\pm$0 & 216.847 & 2.215 & 4 & 0.38$\pm$0.01 & 4.25$\pm$2.87 & 0.50$\pm$1.00 & 0.25$\pm$0.50 & 1.25$\pm$2.50 & 0$\pm$0 \\ \cline{2-20}
        \multicolumn{1}{|l|}{}                      & a        & 0.014 & 0.001 & 1 & 0.84$\pm$0.00 & 5.00$\pm$0.00 & 1.00$\pm$0.00 & 0$\pm$0 & 0$\pm$0 & 0$\pm$0 & 0.014 & 0.001 & 1 & 0.84$\pm$0.00 & 5.00$\pm$0.00 & 1.00$\pm$0.00 & 0$\pm$0 & 0$\pm$0 & 0$\pm$0 \\ \cline{2-20}
        \multicolumn{1}{|l|}{}                      & o        & 0.068 & 0.012 & 2 & 0.75$\pm$0.35 & 3.50$\pm$2.12 & 0.50$\pm$0.71 & 0$\pm$0 & 0$\pm$0 & 0$\pm$0 & 0.077 & 0.005 & 2 & 0.75$\pm$0.35 & 2.00$\pm$0.00 & 0$\pm$0 & 0$\pm$0 & 0$\pm$0 & 0$\pm$0 \\ \hline
        \multicolumn{1}{|l|}{\multirow{3}{*}{\rotatebox[origin=c]{90}{2013}}} & inc      & 26.141 & 1.384 & 2 & 0.43$\pm$0.09 & 4.00$\pm$1.41 & 0.50$\pm$0.71 & 0.50$\pm$0.71 & 0$\pm$0 & 1.00$\pm$1.41 & 24.650 & 1.495 & 2 & 0.43$\pm$0.09 & 4.00$\pm$1.41 & 0.50$\pm$0.71 & 0.50$\pm$0.71 & 0$\pm$0 & 1.00$\pm$1.41 \\ \cline{2-20}
        \multicolumn{1}{|l|}{}                      & clo      & 0.071 & 0.023 & 2 & 0.87$\pm$0.10 & 2.50$\pm$0.71 & 0.50$\pm$0.71 & 0$\pm$0 & 0$\pm$0 & 0$\pm$0 & 0.082 & 0.005 & 2 & 0.76$\pm$0.34 & 1.50$\pm$0.71 & 0$\pm$0 & 0$\pm$0 & 0$\pm$0 & 0.50$\pm$0.71 \\ \cline{2-20}
        \multicolumn{1}{|l|}{}                      & op       & 0.021 & 0.005 & 2 & 0.62$\pm$0.38 & 2.50$\pm$0.71 & 0.50$\pm$0.71 & 0$\pm$0 & 0$\pm$0 & 0$\pm$0 & 0.023 & 0.000 & 0 & 0$\pm$0 & 0$\pm$0 & 0$\pm$0 & 0$\pm$0 & 0$\pm$0 & 0$\pm$0 \\ \hline
        \multicolumn{1}{|l|}{\multirow{5}{*}{\rotatebox[origin=c]{90}{2015}}} & 1        & 2.200 & 0.452 & 7 & 0.48$\pm$0.10 & 2.43$\pm$0.79 & 0.14$\pm$0.38 & 0$\pm$0 & 0.14$\pm$0.38 & 0$\pm$0 & 108.069 & 1.649 & 7 & 0.49$\pm$0.10 & 2.57$\pm$1.13 & 0.14$\pm$0.38 & 0$\pm$0 & 0.14$\pm$0.38 & 0$\pm$0 \\ \cline{2-20}
        \multicolumn{1}{|l|}{}                      & 2        & 0.961 & 0.445 & 6 & 0.50$\pm$0.14 & 3.67$\pm$1.86 & 0.17$\pm$0.41 & 0$\pm$0 & 0.67$\pm$0.52 & 0$\pm$0 & 90.468 & 2.654 & 8 & 0.40$\pm$0.06 & 2.75$\pm$1.04 & 0.50$\pm$0.53 & 0.12$\pm$0.35 & 0$\pm$0 & 0$\pm$0 \\ \cline{2-20}
        \multicolumn{1}{|l|}{}                      & 3        & 3.133 & 0.519 & 6 & 0.43$\pm$0.11 & 2.67$\pm$0.82 & 0.33$\pm$0.52 & 0$\pm$0 & 0.17$\pm$0.41 & 0$\pm$0 & 105.733 & 2.079 & 11 & 0.47$\pm$0.11 & 2.27$\pm$0.65 & 0.09$\pm$0.30 & 0$\pm$0 & 0.09$\pm$0.30 & 0$\pm$0 \\ \cline{2-20}
        \multicolumn{1}{|l|}{}                      & 4        & 1.532 & 0.560 & 7 & 0.53$\pm$0.19 & 2.57$\pm$1.51 & 0$\pm$0 & 0$\pm$0 & 0.14$\pm$0.38 & 0$\pm$0 & 77.010 & 1.546 & 8 & 0.40$\pm$0.06 & 3.50$\pm$1.41 & 0.62$\pm$0.52 & 0$\pm$0 & 0$\pm$0 & 0$\pm$0 \\ \cline{2-20}
        \multicolumn{1}{|l|}{}                      & 5        & 2.008 & 0.921 & 6 & 0.49$\pm$0.17 & 4.00$\pm$1.67 & 0.33$\pm$0.52 & 0$\pm$0 & 0.67$\pm$0.52 & 0$\pm$0 & 131.711 & 2.588 & 8 & 0.38$\pm$0.03 & 3.00$\pm$1.41 & 0.38$\pm$0.52 & 0$\pm$0 & 0$\pm$0 & 0$\pm$0 \\ \hline

            \cline{1-20}
            \multicolumn{20}{l}{}\\[-5px]

\cline{3-20}
                    &                   & \multicolumn{18}{c|}{\multirow{2}{*}{Threshold : 50\%}}   \\
                    &                   & \multicolumn{18}{c|}{}                                    \\ \cline{3-20} 
                    &                   & \multicolumn{9}{c|}{Heuristics Miner}                                                                                                                                                                                                                                                               & \multicolumn{9}{c|}{Inductive Miner}                                                                                                                                                                                                                                                               \\ \cline{3-20} 
                    &                   & \multicolumn{2}{c|}{runtime (secs)}                           & \multicolumn{1}{c|}{\multirow{2}{*}{\#patt}} & \multicolumn{1}{c|}{\multirow{2}{*}{frequency}} & \multicolumn{1}{c|}{\multirow{2}{*}{\#tasks}} & \multicolumn{1}{c|}{\multirow{2}{*}{\#sequences}} & \multicolumn{1}{c|}{\multirow{2}{*}{\#choices}} & \multicolumn{1}{c|}{\multirow{2}{*}{\#parallels}} & \multicolumn{1}{c|}{\multirow{2}{*}{\#loops}}   & \multicolumn{2}{c|}{runtime (secs)}                           & \multicolumn{1}{c|}{\multirow{2}{*}{\#patt}} & \multicolumn{1}{c|}{\multirow{2}{*}{frequency}} & \multicolumn{1}{c|}{\multirow{2}{*}{\#tasks}} & \multicolumn{1}{c|}{\multirow{2}{*}{\#sequences}} & \multicolumn{1}{c|}{\multirow{2}{*}{\#choices}} & \multicolumn{1}{c|}{\multirow{2}{*}{\#parallels}} & \multicolumn{1}{c|}{\multirow{2}{*}{\#loops}}   \\ \cline{3-4} \cline{12-13}   
                    &                   & \multicolumn{1}{c|}{pre} & \multicolumn{1}{c|}{alg} & \multicolumn{1}{c|}{}                        &     &   &     &   &     &   & \multicolumn{1}{c|}{pre} & \multicolumn{1}{c|}{alg} & \multicolumn{1}{c|}{}                        &   &     &     &   &     &   \\ \hline

        \multicolumn{2}{|c|}{2011}                             & 5.847 & 0.914 & 9 & 0.53$\pm$0.02 & 2.44$\pm$1.01 & 0.33$\pm$0.50 & 0$\pm$0 & 0$\pm$0 & 0.22$\pm$0.44 & - & - & - & - & - & - & - & - & - \\ \hline
        \multicolumn{1}{|l|}{\multirow{3}{*}{\rotatebox[origin=c]{90}{2012}}} & fin      & 214.118 & 0.209 & 2 & 0.78$\pm$0.31 & 2.50$\pm$0.71 & 0.50$\pm$0.71 & 0$\pm$0 & 0$\pm$0 & 0$\pm$0 & 216.847 & 0.170 & 2 & 0.78$\pm$0.31 & 2.50$\pm$0.71 & 0.50$\pm$0.71 & 0$\pm$0 & 0$\pm$0 & 0$\pm$0 \\ \cline{2-20}
        \multicolumn{1}{|l|}{}                      & a        & 0.014 & 0.001 & 1 & 0.84$\pm$0.00 & 5.00$\pm$0.00 & 1.00$\pm$0.00 & 0$\pm$0 & 0$\pm$0 & 0$\pm$0 & 0.014 & 0.001 & 1 & 0.84$\pm$0.00 & 5.00$\pm$0.00 & 1.00$\pm$0.00 & 0$\pm$0 & 0$\pm$0 & 0$\pm$0 \\ \cline{2-20}
        \multicolumn{1}{|l|}{}                      & o        & 0.068 & 0.012 & 2 & 0.75$\pm$0.35 & 3.50$\pm$2.12 & 0.50$\pm$0.71 & 0$\pm$0 & 0$\pm$0 & 0$\pm$0 & 0.077 & 0.002 & 2 & 0.75$\pm$0.35 & 2.00$\pm$0.00 & 0$\pm$0 & 0$\pm$0 & 0$\pm$0 & 0$\pm$0 \\ \hline
        \multicolumn{1}{|l|}{\multirow{3}{*}{\rotatebox[origin=c]{90}{2013}}} & inc      & 26.141 & 0.780 & 4 & 0.67$\pm$0.15 & 2.75$\pm$0.50 & 0.25$\pm$0.50 & 0.50$\pm$0.58 & 0$\pm$0 & 0.25$\pm$0.50 & 24.650 & 0.755 & 4 & 0.67$\pm$0.15 & 2.75$\pm$0.50 & 0.25$\pm$0.50 & 0.50$\pm$0.58 & 0$\pm$0 & 0.25$\pm$0.50 \\ \cline{2-20}
        \multicolumn{1}{|l|}{}                      & clo       & 0.071 & 0.021 & 2 & 0.87$\pm$0.10 & 2.50$\pm$0.71 & 0.50$\pm$0.71 & 0$\pm$0 & 0$\pm$0 & 0$\pm$0 & 0.082 & 0.005 & 2 & 0.76$\pm$0.34 & 1.50$\pm$0.71 & 0$\pm$0 & 0$\pm$0 & 0$\pm$0 & 0.50$\pm$0.71 \\ \cline{2-20}
        \multicolumn{1}{|l|}{}                      & op       & 0.021 & 0.001 & 1 & 0.89$\pm$0.00 & 2.00$\pm$0.00 & 0$\pm$0 & 0$\pm$0 & 0$\pm$0 & 0$\pm$0 & 0.023 & 0.000 & 0 & 0$\pm$0 & 0$\pm$0 & 0$\pm$0 & 0$\pm$0 & 0$\pm$0 & 0$\pm$0 \\ \hline
        \multicolumn{1}{|l|}{\multirow{5}{*}{\rotatebox[origin=c]{90}{2015}}} & 1        & 2.200 & 0.241 & 3 & 0.63$\pm$0.03 & 2.67$\pm$0.58 & 0.33$\pm$0.58 & 0$\pm$0 & 0.33$\pm$0.58 & 0$\pm$0 & 108.069 & 1.970 & 4 & 0.61$\pm$0.06 & 2.75$\pm$1.50 & 0$\pm$0 & 0$\pm$0 & 0.25$\pm$0.50 & 0$\pm$0 \\ \cline{2-20}
        \multicolumn{1}{|l|}{}                      & 2        & 0.961 & 0.340 & 5 & 0.62$\pm$0.08 & 3.80$\pm$1.10 & 0$\pm$0 & 0$\pm$0 & 1.00$\pm$0.00 & 0$\pm$0 & 90.468 & 1.580 & 5 & 0.58$\pm$0.08 & 2.20$\pm$0.45 & 0.20$\pm$0.45 & 0$\pm$0 & 0$\pm$0 & 0$\pm$0 \\ \cline{2-20}
        \multicolumn{1}{|l|}{}                      & 3        & 3.133 & 0.255 & 3 & 0.68$\pm$0.05 & 2.67$\pm$0.58 & 0.33$\pm$0.58 & 0$\pm$0 & 0.33$\pm$0.58 & 0$\pm$0 & 105.733 & 1.353 & 6 & 0.59$\pm$0.10 & 2.33$\pm$0.82 & 0$\pm$0 & 0$\pm$0 & 0.17$\pm$0.41 & 0$\pm$0 \\ \cline{2-20}
        \multicolumn{1}{|l|}{}                      & 4        & 1.532 & 0.324 & 4 & 0.67$\pm$0.17 & 3.25$\pm$1.89 & 0$\pm$0 & 0$\pm$0 & 0.50$\pm$0.58 & 0$\pm$0 & 77.010 & 1.143 & 4 & 0.53$\pm$0.03 & 3.00$\pm$0.82 & 0.75$\pm$0.50 & 0$\pm$0 & 0$\pm$0 & 0$\pm$0 \\ \cline{2-20}
        \multicolumn{1}{|l|}{}                      & 5        & 2.008 & 0.555 & 4 & 0.62$\pm$0.11 & 3.50$\pm$1.00 & 0.50$\pm$0.58 & 0$\pm$0 & 0.50$\pm$0.58 & 0$\pm$0 & 131.711 & 2.089 & 5 & 0.53$\pm$0.03 & 2.80$\pm$1.10 & 0.40$\pm$0.55 & 0$\pm$0 & 0$\pm$0 & 0$\pm$0 \\ \hline

        \end{tabular}
    }
    \caption{Behavioral structure of the frequent patterns extracted with thresholds over 20\%, 35\% and 50\% from the process models of the BPICs. It shows the information for the results with two process models of each log (Heuristics and Inductive). The information contains the runtime, the number of patterns and the distribution (average and standard deviation) of the frequency, the number of tasks, sequences, choices, parallels and loops of each pattern. The missing results in the 2011 log with the IM's model are due to a non convergence of the algorithm, taking more than 5 hours to execute a few iterations.}
    \label{tab:results-bpics-1}
\end{table}


 Regarding the runtime, we can observe that most of the values are close to 5 seconds, with the exception of the more complex models (\textit{BPIC 2011}, \textit{BPIC 2012-fin}, \textit{BPIC 2013-inc}, IM of all \textit{BPIC 2015}) ranging from 2 to 7 minutes.
Most of this time is spent on the preprocessing, which can be shared between executions with different thresholds, reducing the total runtime in real applications.
Also, there is a significant difference in the algorithm runtime between models, specially for a threshold of 20\%.
These differences are due to the very different grades of complexity of the mined models.
For higher thresholds, the differences weaken, as most of the structures are pruned in the first analysis, and the expansion step of WoMine does not consider them.


Besides, we have compared the number of patterns discovered for the same threshold for the HM and the IM models.
With logs where the difference between the mined models is higher ---2011 and 2015---, the number of retrieved patterns with more complex models (IM) is significantly higher.
The algorithm builds more structures with these models and, consequently, extracts more patterns.
This difference is attenuated when the threshold increases, because the patterns not represented in the HM models are patterns with low frequency.
On the other hand, with simpler models ---2012 and 2013---, the differences are less notable and the number of patterns extracted are almost the same.

\begin{figure}[b!]
    \centering
    \subfloat[Frequent patterns from the IM model contained in the frequent patterns set from the HM model.]{
        \includegraphics[width=0.485\textwidth]{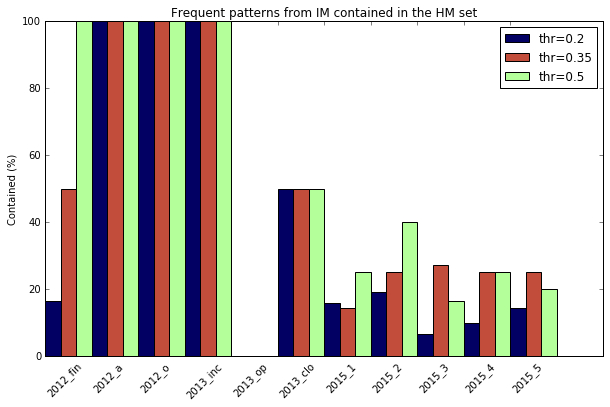}
        \label{fig:inductive-inside-heuristics}
    }
    \hfill
    \subfloat[Frequent patterns from the HM model contained in the frequent patterns set from the IM model.]{
        \includegraphics[width=0.485\textwidth]{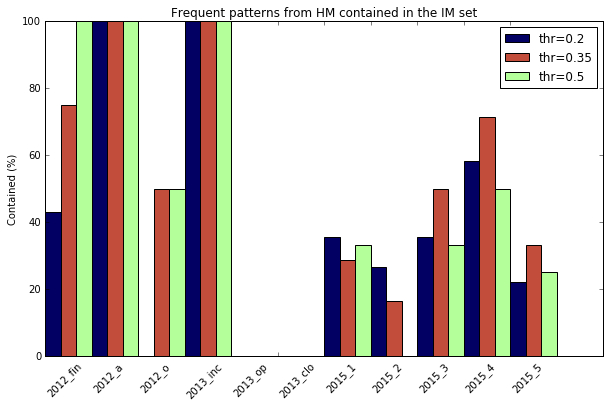}
        \label{fig:heuristics-inside-inductive}
    }
    \caption{Percentage of patterns obtained from the model mined by a discovery algorithm that are contained into patterns from the mined model of the other algorithm.}
    \label{fig:results-inside-results}
\end{figure}

In a more exhaustive comparison, we have analysed how many patterns extracted from one model are contained in the results of the other one (Fig.~\ref{fig:results-inside-results}).
With less complex logs ---2012 and 2013--- almost all of the patterns from the IM model are retrieved using the HM model.
In contrast, with complex models, the set of patterns from the IM model is more complete than the set of patterns from the HM model.
Usually, with a more complex model, more behaviour is examined and more patterns can be retrieved, with the penalty of a higher runtime.
But increasing the complexity of a model might hide structures in the search of WoMine, causing the lost of frequent behaviour.

\begin{figure}[t]
    \centering
    \subfloat[Frequent pattern (65.44\%) extracted from the 2011 log with the Heuristics model. The real name of the task is \textit{'aanname laboratoriumonderzoek'} (assumption laboratory).]{
        \includegraphics[width=0.24\textwidth]{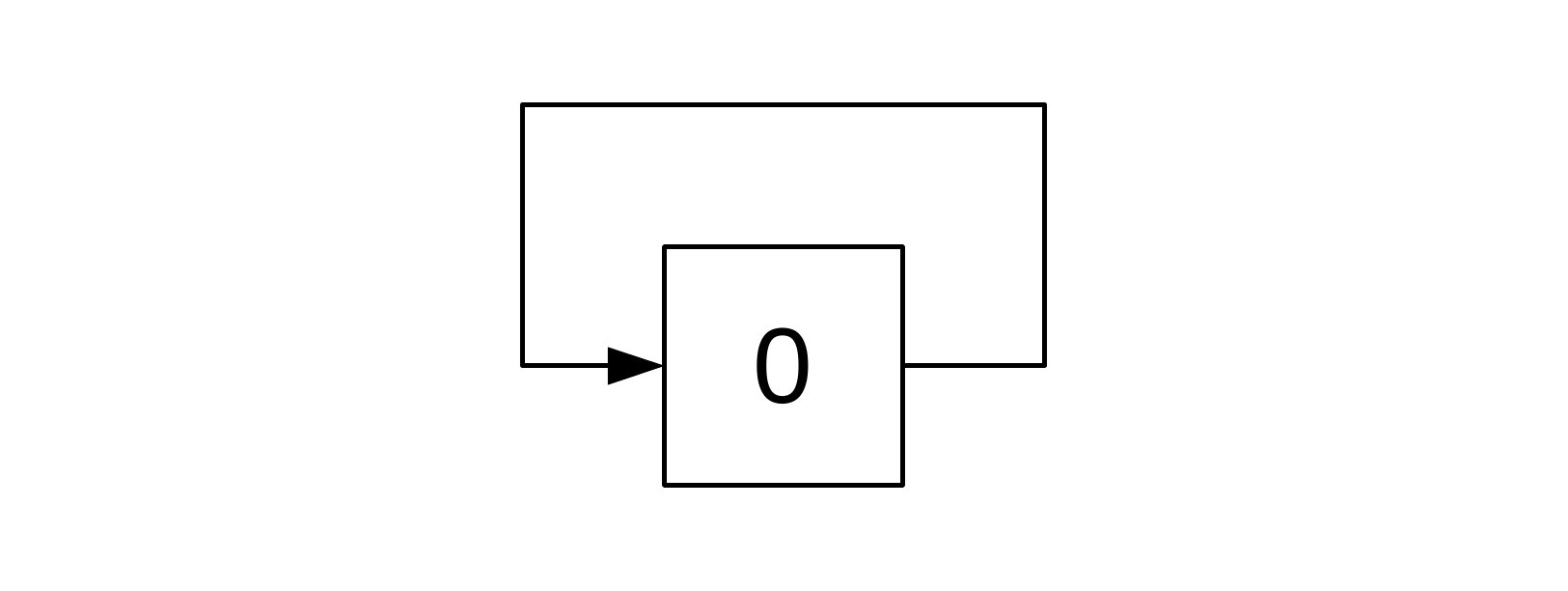}
        \label{fig:bpic-visual-results-1}
    }
    \hfill
    \subfloat[Frequent pattern (20.20\%) extracted from the 2011 log with the Heuristics model. The pattern is formed by two sequences joined by a choice (XOR-join). The real name of the tasks are: 1: \textit{'kalium potentiometrisch'} (potassium potentiometric); 2: \textit{'sgot - asat kinetisch'} (Glutamic-oxalacetic transaminase); 3: \textit{'sgpt - alat kinetisch'} (Glutamic-pyruvic transaminase); 4: \textit{'melkzuurdehydrogenase -ldh- kinetisch'} (Lactic acid dehydrogenase); 5: \textit{'bloedgroep abo en rhesusfactor'} (abo blood group and rhesus factor); 6: \textit{'rhesusfactor d - centrifugeermethode – e'} (Rhesus factor d - Centrifuge method); 7: \textit{'differentiele telling automatisch'} (differential count automatically); 8: \textit{'leukocyten tellen elektronisch'} (leukocyte count electronic); 9: \textit{'trombocyten tellen – elektronisch'} (platelet count electronic)]{
        \includegraphics[width=0.7\textwidth]{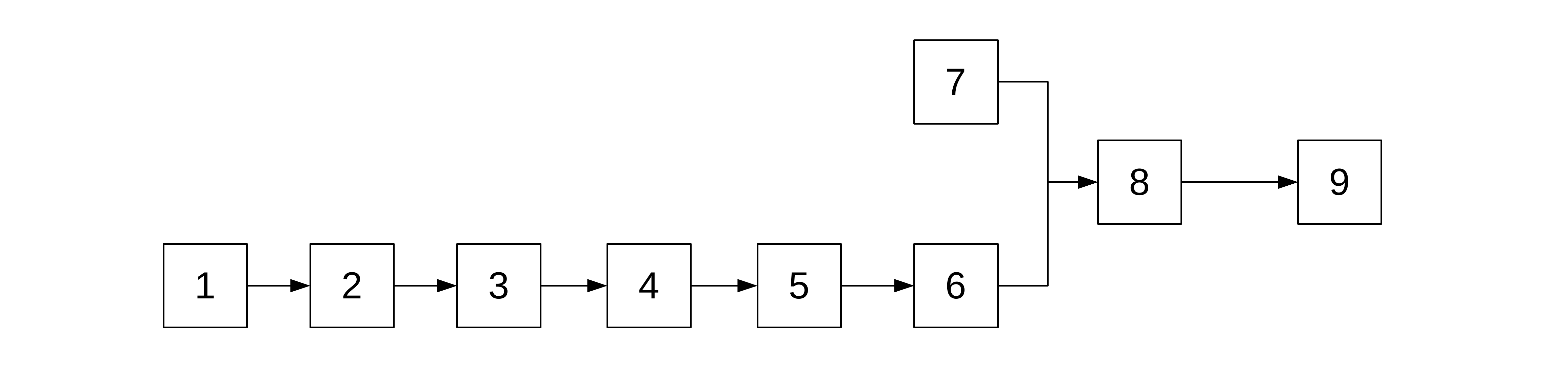}
        \label{fig:bpic-visual-results-2}
    }
    \caption{Two frequent patterns retrieved from the BPIC tests.}
    \label{fig:bpic-visual-results}
\end{figure}

Fig.~\ref{fig:bpic-visual-results} shows two examples of patterns extracted by WoMine from the HM model of the BPIC 2011 log, which corresponds to a Dutch Academic Hospital.
This model contains more than 623 tasks and almost 1,500 arcs.
Fig.~\ref{fig:bpic-visual-results-1} presents a pattern extracted from the model which appears in the 65\% of the traces.
This pattern is formed by a single task, with a loop to itself.
A frequent structure like this may warn the staff of the hospital about a possible error that is occurring in the process.
It might also be a correct behaviour, and its detection may help the process manager not to free the resources used after this task, because is very common to be executed more times.
Fig.~\ref{fig:bpic-visual-results-2} shows another pattern formed by two sequences, joined by a choice.
WoMine detects this pattern in the 20\% of the traces.
With this information, the process manager may try to optimize the subprocess, or schedule the resources to improve the execution of the process.

    \section{Conclusion and Future Work\label{sec:conclusions}}

We have presented WoMine, an algorithm designed to search frequent patterns in an already discovered process model.
The proposal, based on a novel a priori algorithm, is able to find patterns with the most common control structures, including loops.
We have compared WoMine with the state of the art approaches, showing that, although the other proposals fail for some of the models, WoMine always retrieves the correct frequent patterns.
Moreover, we have also tested WoMine with complex real logs from the BPICs.
Results show the importance of the frequent patterns to analyze and optimize the process model.
    \section*{Acknowledgments.}

This work was supported by the Spanish Ministry of Economy and Competitiveness (grant TIN2014-56633-C3-1-R co-funded by the European Regional Development Fund - FEDER program); the Galician Ministry of Education (projects EM2014/012, CN2012/151 and GRC2014/030); the Conseller\'ia de Cultura, Educaci\'on e Ordenaci\'on Universitaria (accreditation 2016-2019, ED431G/08); and the European Regional Development Fund (ERDF).


\section*{References}
    \begin{thebibliography}{31}
\expandafter\ifx\csname natexlab\endcsname\relax\def\natexlab#1{#1}\fi
\expandafter\ifx\csname url\endcsname\relax
  \def\url#1{\texttt{#1}}\fi
\expandafter\ifx\csname urlprefix\endcsname\relax\def\urlprefix{URL }\fi

\bibitem[{Agrawal et~al.(1993)Agrawal, Imieli{\'n}ski, and
  Swami}]{agrawal1993mining}
Agrawal, R., Imieli{\'n}ski, T., Swami, A., 1993. Mining association rules
  between sets of items in large databases. In: Acm sigmod record. Vol.~22.
  ACM, pp. 207--216.

\bibitem[{Agrawal and Srikant(1995)}]{agrawal1995mining}
Agrawal, R., Srikant, R., 1995. Mining sequential patterns. In: Data
  Engineering, 1995. Proceedings of the Eleventh International Conference on.
  IEEE, pp. 3--14.

\bibitem[{Bui et~al.(2012)Bui, Hadzic, and Potdar}]{bui2012framework}
Bui, D.~B., Hadzic, F., Potdar, V., 2012. A framework for application of
  tree-structured data mining to process log analysis. In: International
  Conference on Intelligent Data Engineering and Automated Learning. Springer,
  pp. 423--434.

\bibitem[{de~Medeiros(2006)}]{phdGenetic}
de~Medeiros, A., 2006. Genetic process mining. Ph.D. thesis, Technische
  Universiteit Eindhoven.

\bibitem[{De~San~Pedro et~al.(2015)De~San~Pedro, Carmona, and
  Cortadella}]{de2015log}
De~San~Pedro, J., Carmona, J., Cortadella, J., 2015. Log-based simplification
  of process models. In: International Conference on Business Process
  Management. Springer, pp. 457--474.

\bibitem[{Desel and Reisig(1998)}]{desel1998place}
Desel, J., Reisig, W., 1998. Place/transition petri nets. In: Lectures on Petri
  Nets I: Basic Models. Springer, pp. 122--173.

\bibitem[{Fahland and Van Der~Aalst(2011)}]{fahland2011simplifying}
Fahland, D., Van Der~Aalst, W.~M., 2011. Simplifying mined process models: An
  approach based on unfoldings. In: International Conference on Business
  Process Management. Springer, pp. 362--378.

\bibitem[{Greco et~al.(2006{\natexlab{a}})Greco, Guzzo, Manco, Pontieri, and
  Sacc{\`a}}]{greco2006mining}
Greco, G., Guzzo, A., Manco, G., Pontieri, L., Sacc{\`a}, D.,
  2006{\natexlab{a}}. Mining constrained graphs: The case of workflow systems.
  In: Constraint-Based Mining and Inductive Databases. Springer, pp. 155--171.

\bibitem[{Greco et~al.(2004)Greco, Guzzo, Pontieri, and
  Sacca}]{greco2004mining}
Greco, G., Guzzo, A., Pontieri, L., Sacca, D., 2004. Mining expressive process
  models by clustering workflow traces. In: Pacific-Asia Conference on
  Knowledge Discovery and Data Mining. Springer, pp. 52--62.

\bibitem[{Greco et~al.(2006{\natexlab{b}})Greco, Guzzo, Pontieri, and
  Sacca}]{greco2006discovering}
Greco, G., Guzzo, A., Pontieri, L., Sacca, D., 2006{\natexlab{b}}. Discovering
  expressive process models by clustering log traces. IEEE Transactions on
  Knowledge and Data Engineering 18~(8), 1010--1027.

\bibitem[{G{\"u}nther and Rozinat(2012)}]{gunther2012disco}
G{\"u}nther, C.~W., Rozinat, A., 2012. Disco: Discover your processes. BPM
  (Demos) 940, 40--44.

\bibitem[{Han et~al.(2007)Han, Cheng, Xin, and Yan}]{han2007frequent}
Han, J., Cheng, H., Xin, D., Yan, X., 2007. Frequent pattern mining: current
  status and future directions. Data Mining and Knowledge Discovery 15~(1),
  55--86.

\bibitem[{Han et~al.(2000)Han, Pei, Mortazavi-Asl, Chen, Dayal, and
  Hsu}]{han2000freespan}
Han, J., Pei, J., Mortazavi-Asl, B., Chen, Q., Dayal, U., Hsu, M.-C., 2000.
  Freespan: frequent pattern-projected sequential pattern mining. In:
  Proceedings of the sixth ACM SIGKDD international conference on Knowledge
  discovery and data mining. ACM, pp. 355--359.

\bibitem[{Han et~al.(2001)Han, Pei, Mortazavi-Asl, Pinto, Chen, Dayal, and
  Hsu}]{han2001prefixspan}
Han, J., Pei, J., Mortazavi-Asl, B., Pinto, H., Chen, Q., Dayal, U., Hsu, M.,
  2001. Prefixspan: Mining sequential patterns efficiently by prefix-projected
  pattern growth. In: proceedings of the 17th international conference on data
  engineering. pp. 215--224.

\bibitem[{Leemans and van~der Aalst(2014)}]{leemans2014discovery}
Leemans, M., van~der Aalst, W.~M., 2014. Discovery of frequent episodes in
  event logs. In: International Symposium on Data-Driven Process Discovery and
  Analysis. Springer, pp. 1--31.

\bibitem[{Leemans et~al.(2013)Leemans, Fahland, and van~der
  Aalst}]{leemans2013discovering}
Leemans, S.~J., Fahland, D., van~der Aalst, W.~M., 2013. Discovering
  block-structured process models from event logs-a constructive approach. In:
  International Conference on Applications and Theory of Petri Nets and
  Concurrency. Springer, pp. 311--329.

\bibitem[{Mannila et~al.(1997)Mannila, Toivonen, and
  Verkamo}]{mannila1997discovery}
Mannila, H., Toivonen, H., Verkamo, A.~I., 1997. Discovery of frequent episodes
  in event sequences. Data mining and knowledge discovery 1~(3), 259--289.

\bibitem[{Song et~al.(2008)Song, G{\"u}nther, and Van~der
  Aalst}]{song2008trace}
Song, M., G{\"u}nther, C.~W., Van~der Aalst, W.~M., 2008. Trace clustering in
  process mining. In: International Conference on Business Process Management.
  Springer, pp. 109--120.

\bibitem[{Steeman(2013{\natexlab{a}})}]{BPIC2013-clo}
Steeman, W., 2013{\natexlab{a}}. Bpi challenge 2013, closed problems.
\newline\urlprefix\url{https://doi.org/10.4121/uuid:c2c3b154-ab26-4b31-a0e8-8f2350ddac11}

\bibitem[{Steeman(2013{\natexlab{b}})}]{BPIC2013-inc}
Steeman, W., 2013{\natexlab{b}}. Bpi challenge 2013, incidents.
\newline\urlprefix\url{https://doi.org/10.4121/uuid:500573e6-accc-4b0c-9576-aa5468b10cee}

\bibitem[{Steeman(2013{\natexlab{c}})}]{BPIC2013-op}
Steeman, W., 2013{\natexlab{c}}. Bpi challenge 2013, open problems.
\newline\urlprefix\url{https://doi.org/10.4121/uuid:3537c19d-6c64-4b1d-815d-915ab0e479da}

\bibitem[{Tax et~al.(2016)Tax, Sidorova, Haakma, and van~der
  Aalst}]{tax2016mining}
Tax, N., Sidorova, N., Haakma, R., van~der Aalst, W.~M., 2016. Mining local
  process models. Journal of Innovation in Digital Ecosystems.

\bibitem[{van~der Aalst(2011)}]{vanderAalst2011discovery}
van~der Aalst, W., 2011. Process Mining: Discovery, Conformance and Enhancement
  of Business Processes, 1st Edition. Springer.

\bibitem[{van Dongen(2011)}]{BPIC2011}
van Dongen, B., 2011. Real-life event logs - hospital log.
\newline\urlprefix\url{https://doi.org/10.4121/uuid:d9769f3d-0ab0-4fb8-803b-0d1120ffcf54}

\bibitem[{van Dongen(2012)}]{BPIC2012}
van Dongen, B., 2012. Bpi challenge 2012.
\newline\urlprefix\url{https://doi.org/10.4121/uuid:3926db30-f712-4394-aebc-75976070e91f}

\bibitem[{van Dongen(2015)}]{BPIC2015}
van Dongen, B., 2015. Bpi challenge 2015.
\newline\urlprefix\url{https://doi.org/10.4121/uuid:31a308ef-c844-48da-948c-305d167a0ec1}

\bibitem[{Van~Dongen et~al.(2005)Van~Dongen, de~Medeiros, Verbeek, Weijters,
  and Van Der~Aalst}]{van2005prom}
Van~Dongen, B.~F., de~Medeiros, A. K.~A., Verbeek, H., Weijters, A., Van
  Der~Aalst, W.~M., 2005. The prom framework: A new era in process mining tool
  support. In: International Conference on Application and Theory of Petri
  Nets. Springer, pp. 444--454.

\bibitem[{V{\'a}zquez-Barreiros et~al.(2014)V{\'a}zquez-Barreiros, Lama,
  Mucientes, and Vidal}]{barreiros2014softlearn}
V{\'a}zquez-Barreiros, B., Lama, M., Mucientes, M., Vidal, J.~C., 2014.
  Softlearn: A process mining platform for the discovery of learning paths. In:
  2014 IEEE 14th International Conference on Advanced Learning Technologies.
  IEEE, pp. 373--375.

\bibitem[{V{\'a}zquez-Barreiros et~al.(2015)V{\'a}zquez-Barreiros, Mucientes,
  and Lama}]{vazquez2015prodigen}
V{\'a}zquez-Barreiros, B., Mucientes, M., Lama, M., 2015. Prodigen: Mining
  complete, precise and minimal structure process models with a genetic
  algorithm. Information Sciences 294, 315--333.

\bibitem[{Weijters et~al.(2006)Weijters, van Der~Aalst, and
  De~Medeiros}]{weijters2006process}
Weijters, A., van Der~Aalst, W.~M., De~Medeiros, A.~A., 2006. Process mining
  with the heuristics miner-algorithm. Technische Universiteit Eindhoven, Tech.
  Rep. WP 166, 1--34.

\bibitem[{Zaki(2001)}]{zaki2001spade}
Zaki, M.~J., 2001. Spade: An efficient algorithm for mining frequent sequences.
  Machine learning 42~(1-2), 31--60.

\end{thebibliography}
\end{document}